%% file: conference_101719.tex
\documentclass[conference]{IEEEtran}
\IEEEoverridecommandlockouts

\usepackage{graphicx}
\usepackage{textcomp}
\usepackage{xcolor}

\usepackage{url}

\usepackage{etoolbox}
\usepackage{xspace}
\usepackage[nolist]{acronym}
\usepackage{gnuplot-lua-tikz}
\usepackage{algorithm}
\usepackage{algpseudocode}
\usepackage{subcaption}
\usepackage[disable]{todonotes}
\usepackage{commath}
\def\BibTeX{{\rm B\kern-.05em{\sc i\kern-.025em b}\kern-.08em
    T\kern-.1667em\lower.7ex\hbox{E}\kern-.125emX}}

\newcommand{\ie}{i.\,e.,\xspace}
\newcommand{\eg}{e.\,g.,\xspace}
\newcommand{\etal}{et~al.\xspace}

\newbool{AppendixDetailedResult}
\newbool{AppendixDeviations}

\booltrue{AppendixDeviations} 
\booltrue{AppendixDetailedResult} 

\begin{document}

%\title{Active Few-Shot Learning for Vertex Classification without an Initially Labeled Dataset
%}

\title{Active Few-Shot Learning for Vertex Classification Starting from an Unlabeled Dataset}

%\author{\IEEEauthorblockN{Anonymized Authors} \\
\author{\IEEEauthorblockN{Felix Burr, Marcel Hoffmann, Ansgar Scherp} %
\IEEEauthorblockA{\textit{Ulm University}, Ulm, Germany \\
\{firstname.lastname\}@uni-ulm.de}
}

\begin{acronym}
\acro{GNN}{Graph Neural Network}
\acro{GCN}{Graph Convolutional Network}
\acro{GPN}{Graph Prototypical Network}
\acro{GAT}{Graph Attention Network}
\end{acronym}
\maketitle

\begin{abstract}
Despite the ample availability of graph data, obtaining vertex labels is a tedious and expensive task.
Therefore, it is desirable to learn from a few labeled vertices only. Existing few-shot learners assume a class oracle, which provides labeled vertices for a desired class.
However, such an oracle is not available in a real-world setting, \ie when drawing a vertex for labeling it is unknown to which class the vertex belongs. Few-shot learners are often combined with prototypical networks, while classical semi-supervised vertex classification uses discriminative models, \eg \acp{GCN}. In this paper, we train our models by iteratively prompting a human annotator with vertices to annotate. We perform three experiments where we continually relax our assumptions. First, we assume a class oracle, \ie the human annotator is provided with an equal number of vertices to label for each class.
We denote this as ``Balanced Sampling''. In the subsequent experiment, ``Unbalanced Sampling,'' we replace the class oracle with $k$-medoids clustering and draw vertices to label from the clusters. In the last experiment, the ``Unknown Number of Classes,'' we no longer assumed we knew the number and distribution of classes. Our results show that prototypical models outperform discriminative models in all experiments when fewer than $20$ samples per class are available. While dropping the assumption of the class oracle for the ``Unbalanced Sampling'' experiment reduces the performance of the \ac{GCN} by $9\%$, the prototypical network loses only $1\%$ on average. For the ``Unknown Number of Classes'' experiment, the average performance for both models decreased further by $1\%$.

Source code: \url{https://github.com/Ximsa/2023-felix-ma}
\end{abstract}

\section{Introduction}
\label{sec:introduction}

In many fields, data is organized as networks or graphs, where vertices are connected by links. 
For example, in citation networks, a link exists between two vertices if one paper cites another. 
In social networks, people represent vertices, and links are formed based on relationships or shared interests. 
While collecting such graphs is often inexpensive, obtaining labels for vertices from human annotators is tedious and expensive. 
Given an unlabeled graph and a fixed annotation budget, our goal is to select vertices for labeling in a way that maximizes the performance of \mbox{Graph Neural Networks (GNNs)} trained on these labels.

%background for few-shot learning
To best use a given budget, one must select the most informative vertices for annotation and use GNNs that learn efficiently from a small number of labeled samples.
Few-shot learning addresses this issue by enabling models to generalize from a small set of labeled instances~\cite{fei2006oneshot, koch2015siamese, liu2021location}.
For $N$-way $K$-shot few-shot learning, $N$ classes each with $K$ shots are chosen for training, \ie each class is represented by an equal number of $K$ samples during training.
Therefore, traditional few-shot learners rely on a class oracle, \ie it is guaranteed that a specific amount of annotated samples of each class are available.
However, in real-world scenarios, such oracles do not exist as it is unknown to which class a newly annotated vertex belongs.
Furthermore, existing few-shot learners such as~\cite{fei2006oneshot, koch2015siamese, liu2021location, finn2017modelagnosticmetalearningfastadaptation} require a large, initial set of labeled samples from other classes to be able to learn a new class from a few samples.
These labeled classes are not available in our setting.
We start without an initial labeled set of vertices, \ie we start annotating from scratch.

\begin{figure*}
\begin{subfigure}[t]{.32\textwidth}
  \centering
  \captionsetup{margin=0.05cm}
  \includegraphics[width=0.9\linewidth]{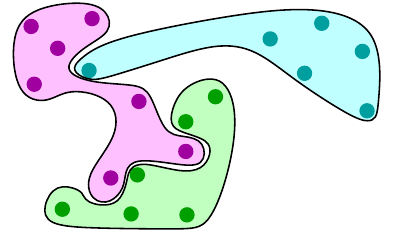}
  \caption{In the ``Balanced Sampling'' setting using the ground truth partition, we sample equally from each of the $k$ classes.}
\end{subfigure}
\hfill
\begin{subfigure}[t]{.32\textwidth}
  \centering
  \captionsetup{margin=0.05cm}
  \includegraphics[width=0.9\linewidth]{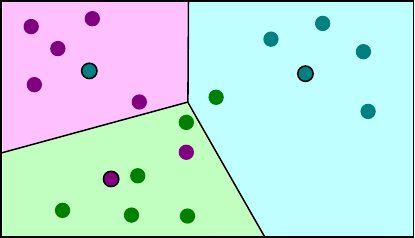}
  \caption{In the ``Unbalanced Sampling'' setting, we cluster the embedding space with \mbox{$k$-medoids} clustering and $k=|C|$.}
\end{subfigure}
\hfill
\begin{subfigure}[t]{.32\textwidth}
  \centering
  \captionsetup{margin=0.05cm}
  \includegraphics[width=0.9\linewidth]{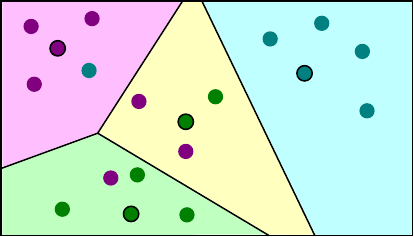}
  \caption{In the ``Unknown Number of Classes'' setting, we cluster the embeddings space with \mbox{$k$-medoids} clustering, but $k$ is an estimated number of classes.}
\end{subfigure}
\caption{The embedding space of each class partition for all experiments, where each color represents a (potential) class.}
    \label{fig:procedure}
\end{figure*}

To overcome the need for a class oracle in few-shot learning, we replace it with $k$-medoids clustering of the embedding space to assign pseudo-labels to each vertex.
These pseudo-labels are used to divide our vertices into disjoint sets.
From each set, we aim to select vertices that increase the model performance the most, which are then annotated by an artificial human-in-the-loop annotator.
We investigate four strategies to choose vertices based on random, entropy, PageRank, and medoid clustering.
In summary, we combine a few-shot learning model with an active learning strategy to select a small set of vertices for labeling and maximizing their potential information gain. 
By including label propagation, we exploit the homophily of our datasets.
%This increases the effectiveness of the active learning methods on homophilic graphs by artificially increasing the number of vertices used for training. 
Existing active learners usually ignore that many graphs are homophilic~\cite{bilgic2010active, garcia2018fewshot}.
We show that exploiting this property increases the effectiveness of graph active learners by artificially increasing the number of vertices used for training.
Typical GNNs are discriminative models, where the model directly learns a decision boundary between classes.
In literature, prototypes have been used for label-efficient classification in few-shot learning scenarios~\cite{snell2017prototypical}.
We calculate prototypes based on Euclidean distances between the embeddings of the \mbox{\ac{GNN}} to represent the prototypical models and compare both paradigms in our experiments.
As can be seen in Figure~\ref{fig:procedure}, we perform three experiments where we relax some assumptions in each setting.
In the ``Balanced Sampling'' experiment, we assume to have a class oracle like in traditional few-shot learning.
In the ``Unbalanced Sampling'' experiment, we drop this assumption and use $k$-medoids clustering for pseudo-label assignment.
In the ``Unknown Number of Classes'' experiment, we drop the assumption of knowing the number of classes and use an estimation instead.
Our results demonstrate that prototypical models outperform discriminative models across all experiments when fewer than $20$ samples per class are available.
While the average performance of discriminative models decreases by $9\%$ when we drop the assumption of a class oracle, the prototypical models only lose $1\%$.
Our main contributions are:
\begin{itemize}
    \item We combine active learning with label propagation and perform experiments on graph prototypical and discriminative networks.
    \item We overcome the limitation of using a class oracle in few-shot tasks and show that estimating the number of classes does not decrease performance.
    \item We analyze extending our budget with artificial labels obtained from label propagation and investigate the performance of discriminative and prototypical networks.
    \item We perform experiments on five benchmark datasets and three settings to verify our findings.
\end{itemize}

\section{Related Work}

We discuss graph neural networks, few-shot learning, and active learning, along with the cold start problem.
\label{sec:relatedwork}

\paragraph{Graph Neural Networks}
\label{sec:graphneuralnetworks}
Kipf~\etal~\cite{kipf2017semisupervised} introduced the \acf{GCN}, which performs message passing based on low-pass filters for vertex classification. %on citation and knowledge graphs, .
\ac{GAT} has been proposed by Veličković~\etal~\cite{veličković2018graph}, which computes the representations of each vertex in the graph by attending to its neighbors based on a self-attention mechanism.
The primary distinction between \ac{GAT} and \ac{GCN} is \ac{GAT}'s capability to assign importance to the edges, while \ac{GCN} treats all neighbors equally.
Hamilton~\etal~\cite{hamilton2018inductive} propose GraphSAGE.
GraphSAGE generalizes \ac{GCN} to arbitrary aggregation functions and separating a vertex weights from its neighborhood.
It leverages vertex feature information to generate vertex embeddings by sampling features from a vertex's neighborhood, keeping the computational cost limited for each batch.

\paragraph{Few-Shot Learning}
\label{sec:few-shotlearning} 
In few-shot learning, the model has to make predictions using only a few training examples~\cite{fei2006oneshot}.
Few-shot learning can be described as N-way-K-shot-classification, where $N$ denotes the number of classes to classify and $K$ is the number of labeled samples available from each class.
A special case is zero-shot learning, where the goal is to classify into classes where no training data is available~\cite{chang2008importance}.
Koch~\etal~\cite{koch2015siamese} use a metric learning approach for one-shot learning with deep siamese networks.
The deep siamese networks learn a distance metric in which items in the same class are closer together while items in different classes are further apart.
Snell~\etal~\cite{snell2017prototypical} introduce prototypical networks for the problem of few-shot classification using meta-learning.
Each class is represented by a weighted mean of its samples in the embedding space.
They randomly sample classes for each training episode.
These classes are separated into disjunct support and query sets.
From the support set, the prototype is computed using the average of the outputs from the prototypical network.
Finally, the loss is evaluated by comparing the distance of the embedding of the query set to the respective prototype.
Liu~\etal~\cite{liu2021location} introduce Relative and Absolute Location Embedding (RALE), based on hub vertices.
Hub vertices are structurally important vertices in the graph.
Those hubs are used to capture dependencies between distant vertices in a few-shot task.
Specifically, the dependencies are the path from a support vertex to a query vertex.
This is called the ``relative location`` of the query within the task.
Additionally, they use these hub vertices to align the ``relative location`` across few-shot tasks.
The hubs serve as a set of global references and assign each vertex an ``absolute location.''
The assignment is based on the path from each hub to a vertex using random walks.
Ding~\etal~\cite{ding2020graph} introduce \acfp{GPN}, which is based on the work of Snell~\etal~\cite{snell2017prototypical}.
They construct a pool of semi-supervised vertex classification tasks to mimic the real test environment.
Their \ac{GPN} performs meta-learning on an attributed network and derives a model to handle the target classification task.
They need fully labeled classes for their training procedure.
Sun~\etal~\cite{SUN2022medimagefewshot} proposed a method for medical image segmentation using few-shot learning. 
They developed a global correlation module to capture the correlation between a support and query image.
They impose a discriminative embedding constraint, which improves performance on unseen classes.
Pan~\etal~\cite{pan2024learning} propose a cascade framework for few-shot six degrees-of-freedom pose estimation that requires only RGB images.
They reduced the false positive rate, by initializing top-K pose candidates based on similarity and refined poses in stages.
%https://openreview.net/forum?id=p3m_WpN0rEX %they use a self-supervised vision transformer,

\paragraph{Active Learning}
\label{sec:activelearning}
In Active learning the user (human-in-the-loop) is involved to provide labeled data points~\cite{Yoo_2019_CVPR}.
In the context of vertex classification, labeled vertices can be provided to guide the training process~\cite{bilgic2010active}.
There are multiple vertex selection heuristics, including uncertainty and diversity sampling~\cite{Mosqueira-Rey2023}.
With uncertainty sampling, the vertices are chosen based on an uncertainty measure, computed on the model logits, \eg entropy, least confidence, or best versus second best~\cite{settles.tr09}.
% With entropy sampling, the vertex with the highest amount of entropy is chosen.
% Least confidence uses the maximum logit of a data point.
% The vertex with the lowest maximum logit is selected.
% Best versus second best considers the two highest logits, choosing the vertex with the lowest difference between those logits.
Diversity sampling involves selecting instances that are dissimilar from each other.
This maximizes the diversity and coverage of the labeled data.
Another line of work relies on measuring the similarity between samples.
Hsu~\etal~\cite{hsu2019unsupervised} propose a two-stage method in which they first use $k$-means to obtain pseudo-labels. 
In the second stage, they train a model under the supervised few-shot setting with those pseudo-labels.
Qin~\etal~\cite{qin2020diversity} aim to increase the diversity of generated few-shot tasks.
They strengthen the distribution shift between the support and query set using data augmentation.
Köksal~\etal~\cite{köksal2022meal} combine active learning with few-shot learning by Multiprompt finetuning and prediction Ensembling with Active Learning (MEAL) on text.
MEAL tries to eliminate model instability by using ensembles and combines uncertainty with diversity sampling.
They sample multiple training sets using $k$-means and choose the set with the highest uncertainty score.
This is used to fine-tune pre-trained models.

%\paragraph{Active Learning on Graphs} 
Bilgic~\etal~\cite{bilgic2010active} developed Alfnet, which uses the network structure to select samples for labeling. 
They use a disagreement score, for instance, selection, which uses the output of three classifiers, a linear regression classifier, a semi-supervised classifier, and a semi-supervised classifier with principal component analysis.
The more diverse the predictions of those classifiers, the greater the disagreement about a vertex label.
Wu~\etal~\cite{wu2021active} investigate active learning for vertex classification on citation graphs.
They propose an instance-selection method that uses vertex feature propagation followed by $k$-medoids clustering of the vertices.
They use the $L_2$ norm of the difference between the propagated vertex features.
Those vertex features resemble the output of a simplified \ac{GCN}~\cite{kipf2017semisupervised} by dropping all activation functions and layer-related parameters in the original structure.
They select $K$ instances, which is the labeling budget, where each cluster center is chosen for labeling.
Ren~\etal~\cite{ren2022dissimilar} use three types of dissimilarity measures for active instance selection,
%They introduce three types of dissimilarity measures: 
feature dissimilarity, structural dissimilarity, and embedding dissimilarity.
The feature dissimilarity uses the cosine distance between feature vectors as a measure of dissimilarity.
The structural dissimilarity is based on the neighborhood of a given vertex.
Two vertices are similar if they have a similar edge structure.
The embedding dissimilarity uses on the embeddings of feature vectors.
Sogli~\etal~\cite{Solgi2022} have a similar procedure as Ding~\etal~\cite{ding2020graph} based in meta-learning.
They combine few-shot with active learning by randomly choosing $|N|-1$ classes in a $N$-way $N$-shot setting. 
The $|N|$-th class is chosen by an active learning method, which selects the final class based on the model's uncertainty. 
The training samples are chosen from the labeled vertices based on the likelihood of belonging to the chosen class.
Huang~\etal~\cite{huang2024AL} developed an active learning method for \ac{GNN} that maximized the diversity of aggregated features in the sampling step.
However, they assume to have access to an initially labeled set of vertices, which we do not.

%\paragraph{Cold Start Problem}
A common problem with active learning on completely unlabeled datasets is the cold start problem.\todo{Is there a citation that defines the cold start problem?} 
As the model does not have information yet, it is hard to make decisions on which data points to query for labeling.
Chen~\etal~\cite{chen2022making} discover that %in many instances, 
uncertainty sampling performs worse than random selection, when using active learning.
They propose a strategy using $k$-means clustering in combination with contrastive learning to improve the performance of early labeling prompts.
Their strategy includes selecting hard-to-contrast data points based on cosine similarity.
Jin~\etal~\cite{JIN202216} propose a method to perform cold start active learning on image datasets.
They extract features by mapping images into the feature space using contrastive self-supervised learning.
These feature vectors get clustered into $m$ clusters using BIRCH~\cite{zhang1996birch}.
The most representative vertices from each cluster are used to form the core set, which is given to a human-in-the-loop for labeling.

\section{Prototypical and Discriminative Models for Vertex Classification}
\label{sec:methods}
We formalize our problem and introduce the assumptions that have been made. 
We use a prototypical model to obtain embeddings from the vertex features.
The model is trained on (few) vertices from a sampler that are perfectly labeled by a human annotator.
Using the embeddings, we calculate a prototype representation for each class. 
The loss is computed using the distance from the vertex embeddings to the computed prototype.

Given a graph $G=(V, E)$ and a budget $B$.
Each vertex belongs to one class $c\in C$.
The set of vertices $V$ consists of disjoint subsets of train and test vertices.
The training vertices consist of unlabeled and labeled vertices $V_u$ and $V_l$.
The set of labeled vertices $V_l$ is initially empty.
It is updated by labeling vertices from $V_u$ in each iteration.
% By using label propagation on $V_l$, some vertices from $V_u$ are included in the propagated set $V_p$ using the respective pseudo-label from $V_l$.
The objective is to iteratively select the most informative vertices from $V_u \setminus V_l$ such that the performance on the test set is maximized.
The model can use, at most, the budget of $B$-many labeled vertices for training.

We assume that the human annotator perfectly labels given vertices.
Furthermore, we assume a transductive learning setting, \ie all vertices and edges on the graph are present during training.
The dataset is assumed to be homophilic, which is the foundation of being able to apply label propagation.
In homophilic graphs, similar vertices are likely to share edges.
%
% \paragraph{Label Propagation}
Based on the homophily assumption, we use label propagation, as described by Huang~\etal~\cite{huang2020combining}, to generate pseudo-labels.
Label propagation uses the edges $E$ to propagate the labels of given vertices over $n$ hops.
The logits of propagated vertices are computed by 
$
    \label{eq:lp}
    \mathbf{Y}^{\prime} = \alpha \cdot \mathbf{D}^{-1/2} \mathbf{A}
\mathbf{D}^{-1/2} \mathbf{Y} + (1 - \alpha) \mathbf{Y}
$,
where $\mathbf{D}^{-1/2} \mathbf{A}\mathbf{D}^{-1/2}$ is the normalized adjacency matrix.
Pseudo-labels are gained by selecting the label with the highest logit.
From pre-experiments, we chose hyperparameter $\alpha = 0.9$, as this balances the influence of the own label with the neighbors' appropriately.
We also found benefits in removing pseudo-labels, which are too uncertain.
Thus, we compute the normalized entropy of the logits.
We found the threshold of $0.2$ appropriate for keeping a pseudo label, as this ensures that a pseudo label is not included when the neighborhood is too heterogeneous.

\paragraph{Discriminative Learning}
For graph neural networks, we denote $h_v^{(k)}$ as the embeddings of vertex $v$ on layer $k$.
With the first layer ($k=0$) being the original features of vertices $v$.
As a representative for discriminative learning, we use \ac{GCN}~\cite{kipf2017semisupervised}.%, which is an extension of convolutional neural networks to graphs.
As their \ac{GCN} performs very well on homophilic datasets, we choose their graph convolutional operator as a backbone for our models.
To get the final class predictions, we introduce a softmax layer to provide logits.

\paragraph{Prototypical Learning}
We use \ac{GCN} to construct a graph prototypical network, which serves as a representative for prototypical learning.
Instead of applying softmax on the \ac{GCN} for label assignment, we use the output vertices' logits to calculate a prototype representation for each class.
The prototype $r_c$ of class $c$ is the weighted mean of the embeddings belonging to $V_{l,c}\subseteq V_{l}$, where the label is known.
The weight is the normalized PageRank~\cite{brin1998pagerank} score $\widetilde{s}$.
Formally, the class-based prototype is defined by
$
    \label{eq:proto}
    r_c = \sum_{v \in V_{l,c}}{h_v\cdot \widetilde{s}_v}
$.
For the prediction, each vertex gets the label assigned by its closest prototype, measured by Euclidean distance.
The logits are computed by
$
    \label{eq:proto_logits}
    p(c|v) = {e^{-d(h_{v}, r_c)}}/{\sum_{c'\in C}{e^{-d(h_{v}, r_{c'})}}}
$,
$h_v$ denotes the embedding of vertex $v$, and $d(a,b)$ is the Euclidean distance between two vectors.
The loss for the prototypical network is based on Xu~\etal~\cite{10.1145/3607144}.
It consists of three parts, a prototype distance-based loss and two regularization terms.
The loss function is based on the Euclidean distance between vertices and their respective prototypes, while the regularization terms are solely based on prototype-to-prototype distances.
The intra-class loss $L_p$~\cite{10.1145/3607144} minimizes the distances between the embeddings and prototypes and is defined by 
\begin{equation*}
    \label{eq:proto_loss}
    L_p = \frac{1}{|V_l|}\sum_{c\in C}{\sum_{v\in V_{l,c}}{-log(p(c|v))}}.
\end{equation*}
The prototype distance regularization is divided into the Euclidean regularization $L_e$~\cite{10.1145/3607144}, which maximizes Euclidean distances between prototypes, as defined by
\begin{equation*}
    \label{eq:eucl_loss}
    L_e = \frac{1}{|C|}\sum_{c\in C}{max_{c' \in C \setminus c}\left(e^{-d(r_c,r_{c'})}\right)},
\end{equation*}
and the cosine regularization~\cite{10.1145/3607144}, defined as
\begin{equation*}
    \label{eq:cos_loss}
    L_c = \frac{1}{|C|}\sum_{c\in C}{max_{c' \in C \setminus c}\left(\frac{r_c - r_m}{||r_c - r_m||}\cdot\frac{r_{c'} - r_m}{||r_{c'} - r_m||}\right)} + 1,
\end{equation*}
with $r_m = \frac{1}{|C|}\sum_{c\in C}{r_c}$ being the mean of all prototypes. 
The total loss $L = L_r + \lambda(L_e + L_c)$ is the sum of the intra-class loss and between-prototype regularization, weighted by the hyperparameter $\lambda$.

\section{Experimental Apparatus}
\label{sec:experimentalapparatus}

We introduce the datasets and describe in detail the active learning procedure and evaluation.

\subsection{Datasets}
\label{sec:datasets}

Table~\ref{table:datasets} summarizes the dataset statistics, including the number of vertices, edges, features, classes, and homophily ratio~\cite{lim2021large}.
Homophily is the tendency of edges in a graph to connect vertices with the same class label.
We include the Planetoid~\cite{planetoid} datasets: Cora, CiteSeer, and PubMed, as well as the large Reddit2 and ogb-arXiv datasets.

\begin{table}
\centering
%\resizebox{\columnwidth}{!}{
\begin{tabular}{l|rrrrr}
 Dataset & $|V|$ & $|E|$ & $|F|$ & $|C|$ & Homophily\\
 \hline 
 Cora & $2,708$ & $5,429$ & $1,433$ & $7$ & $0.63$\\
 CiteSeer & $3,327$ & $4,732$ & $3,703$ & $6$ & $0.77$\\
 PubMed & $19,717$ & $44,338$ & $500$ & $3$ & $0.66$\\
 Reddit2 & $232,965$ & $23,213,838$ & $602$ & $41$ & $0.69$\\
 ogb-arXiv & $169,343$ & $1,166,243$ & $128$ & $40$ & $0.44$\\
\end{tabular}%}
\caption{Overview of our Datasets. $|V|$ denotes the amount of vertices in the graph, $|E|$ the amount of edges, $|F|$ the feature vector length, and $|C|$ the number of classes. Homophily~\cite{lim2021large} measures how many connected vertices share the same class\ifbool{AppendixDeviations}{ (details on this measure are in Appendix~\ref{eq:homophily})}.}
\label{table:datasets}
\end{table}

Cora~\cite{cora} is a citation graph with Bag of Words vertex features. Its publications are classified into seven computer science classes.
%: Case Based, Genetic Algorithms, Neural Networks, Probabilistic Methods, Reinforcement Learning, Rule Learning, and Theory.
%
CiteSeer~\cite{citeseer} is a citation graph consisting of scientific publications with six classes.
Each publication has a Bag of Words feature vector. 
The classes cover sub-fields of computer science.
%, namely Agents, Artificial Intelligence, Databases, Information Retrieval, Machine Learning, and Human-Computer Interaction.
%
PubMed~\cite{pubmed} is a citation graph containing publications on diabetes.
The feature vector consists of 500 Term Frequency-Inverse Document Frequency embeddings.
The task is to classify vertices into three types of diabetes.
Reddit2~\cite{zeng2020graphsaint} contains posts made in September 2014.
Posts are connected if a user comments on both posts.
To create feature representations for the posts, GloVe~\cite{pennington2014glove} embeddings were used, which incorporate word co-occurrences as well as local context information for word embeddings.
The task is to classify posts that are labeled by the subreddit in which the posts were made.
The ogb-arXiv~\cite{hu2021opengraphbenchmarkdatasets} is a citation graph of computer science papers on arXiv.
The papers are classified into 40 categories within computer science, where the labels were chosen by the paper's authors and arXiv moderators.
The $128$-dimensional feature vector is obtained using Word2Vec on its title and abstract.

\subsection{Procedure}
\label{sec:procedure}

The experiments are organized into three steps, ``Balanced Sampling'', ``Unbalanced Sampling'', and ``Unknown Number of Classes''.
The three steps are illustrated in Figure~\ref{fig:procedure}.
Furthermore, we investigate the influence of label propagation on each of the three experiments.

%\paragraph{Active Learning Strategies}
In each experiment, we proceed as follows:
Given a dataset of unlabeled vertices and a choice of a GNN model, we start by assigning pseudo-labels to each vertex using $k$-medoids clustering, where $k=|C|$, the number of classes. 
This partitions the vertex set according to the pseudo-labeled classes.
Next, we sample vertices from each pseudo-labeled class using one of various strategies.
We use random, entropy, PageRank, and medoid sampling.
Random sampling selects vertices uniformly from $V_{u,c}$ to be annotated.
With entropy sampling, we use weighted sampling on $V_{u,c}$ with the Shannon entropy of the logits as bias.
Similarly, PageRank sampling uses the PageRank of the vertices as a bias for sampling.
Medoid sampling performs $k$-medoids clustering on the logits and selects the cluster centers. 
The model is then trained on the extended training set, and we recalculate our pseudo-label assignment\ifbool{AppendixDeviations}{, as seen in Appendix~\ref{sec:training}}.

For the ``Balanced Sampling'' experiment, we use a class oracle to provide vertices from a specific class.
We assume that the vertices are perfectly labeled and that exactly one sample per class is provided. % (``perfect clustering'').
We sample one vertex from each class with the active learning strategy.
We start with $1$-shot learning, \ie select one vertex per class, and continue this process by increasing the number of training samples by one vertex per class at each iteration.
We continue this process until we reach a $|C|$-way $5$-shot learning setting.
This causes the budget to differ per dataset, as we set our budget to $B=|C|\cdot5$ to perform $5$ sampling rounds.
The limit of $5$ sampling rounds is used to balance computation time with the number of results.
All experiments are repeated $10$ times with random seeds.

For the ``Unbalanced Sampling'' experiment, we drop the assumption of a class oracle, \ie we do not guarantee to provide an annotated vertex per class but make it dependent on the sampling strategy.
However, we still retain the assumption of knowing the number of classes.
Although we do not guarantee an exact $|C|$-way $5$-shot scenario, we employ a sampling strategy that takes a budget of $B=|C| \cdot 5$ vertices for the human-in-the-loop annotator into account.
We still assume that the drawn vertices are perfectly annotated by a human in the loop, but this time, the coverage of all classes is not guaranteed.

For the ``Unknown Number of Classes'' experiment, we drop the assumption that we know the number of classes in our dataset.
Instead, we perform $k$-means clustering for $k\in[2,100]$ on embeddings generated by Deep Graph Infomax~\cite{vel2018deep}, which is known to increase downstream performance over raw vertex features.  
%Deep Graph Infomax is chosen over the raw vertex features since it is a well-established graph embedding technique.
Using the elbow rule~\cite{Thorndike1953} on each clustering, we get a class estimation for each dataset. \ifbool{AppendixDeviations}{This is further described in Appendix~\ref{sec:classest}.}
By having a class estimation for $k$, instead of assuming to know the number of classes $|C|$, our sampling strategy also samples the estimated number of $k$ vertices for each iteration.
To keep the results comparable with the previous experiment, we fix our budget to $B=|C|\cdot 5$ sampled vertices.

Additionally, we investigate the effect of exploiting the homophily of our graphs.
We compare the results of our three experiments to using label propagation on the labels provided by the \mbox{human-in-the-loop} annotator.
This artificially extends our labeled set by pseudo-labels, \ie it generates more labeled vertices, but also introduces noisy labels.
We measure the performance of our models by the prediction accuracy on a separate test set.
Again, we calculate the average accuracy over $10$ repeats.
As baselines, we adopt the method of Wu~\etal~\cite{wu2021active} to our setting, where we use $k$-medoids clustering for the active learning instance selection to train a \ac{GCN}.
We choose full label propagation for each drawn sample in combination with our sampling strategies as our second baseline.
For the active learning strategies, we use random, entropy, PageRank, and medoid sampling.

\section{Results}

\newcommand{\scaleboxnumber}{1.0}

% stage 1
\begin{table*}
\caption{Results of the ``Balanced Sampling'' experiment, where we assumed a class oracle. Measurements are given as average accuracy scores after five sampling rounds. Best accuracies in a block are highlighted with underlines, while the best accuracy in a column is bold. ``+LP'' denotes additional label propagation.}
\centering
\scalebox{\scaleboxnumber}{
\begin{tabular}{l|ll|ll|ll|ll|ll}
               & Cora          & +LP                               & CiteSeer     & +LP                               & PubMed        & +LP                               & Reddit2      & +LP                               & arXiv  & +LP         \\
 \hline
 FeatProp     & $62.3_{3.6}$  &                                   & $51.8_{6.8}$ &                                   & $61.4_{6.2}$  &                                   & $38.1_{3.1}$ &                                   & $40.4_{5.4}$&             \\
 \hline 
 LP + Random   & $54.7_{4.0}$  & $54.9_{4.2}$                      & $35.6_{3.4}$ & $35.7_{3.8}$                      & $54.4_{9.3}$  & $54.3_{9.3}$                      & $88.7_{0.1}$ & $88.7_{0.1}$                      & $47.1_{0.2}$& $47.1_{0.2}$\\
 LP + Entropy  & $54.7_{4.0}$  & $54.9_{4.2}$                      & $35.6_{3.4}$ & $35.7_{3.8}$                      & $55.0_{8.8}$  & $55.2_{8.8}$                      & $88.7_{0.1}$ & $88.7_{0.1}$                      & $47.1_{0.2}$& $47.1_{0.2}$\\
 LP + PageRank & $56.8_{2.4}$  & $57.1_{2.4}$                      & $37.2_{2.7}$ & $\underline{37.4}_{2.7}$          & $60.3_{7.6}$  & $60.2_{7.5}$                      & $\underline{\mathbf{89.4}}_{0.1}$ & $89.4_{0.1}$ & $\underline{\mathbf{50.5}}_{0.5}$& $50.5_{0.5}$\\
 LP + Medoid   & $62.8_{3.3}$  & $\underline{63.8}_{3.2}$          & $37.3_{2.1}$ & $37.3_{2.7}$                      & $\underline{62.4}_{4.7}$  & $59.9_{5.5}$          & $84.5_{0.5}$ & $84.5_{0.5}$                      & $43.5_{1.0}$& $43.5_{1.0}$\\
 \hline
 GCN + Random  & $54.6_{10.3}$ & $61.0_{6.9}$                      & $51.9_{7.5}$ & $52.7_{3.9}$                      & $48.2_{16.4}$ & $51.7_{6.4}$                      & $25.8_{1.9}$ & $19.1_{5.9}$                      & $17.6_{2.6}$& $6.0_{3.0}$ \\
 GCN + Entropy & $56.4_{4.9}$  & $61.7_{6.1}$                      & $50.6_{8.0}$ & $53.0_{3.9}$                      & $49.4_{17.7}$ & $54.5_{7.1}$                      & $25.0_{2.1}$ & $18.7_{5.3}$                      & $16.1_{0.0}$& $7.4_{4.4}$ \\
 GCN + PageRank& $60.0_{6.8}$  & $65.1_{5.3}$                      & $52.0_{5.3}$ & $55.4_{5.0}$                      & $48.9_{11.3}$ & $61.0_{7.0}$                      & $\underline{26.1}_{3.8}$ & $20.2_{4.3}$          & $\underline{18.9}_{4.8}$& $10.6_{5.3}$\\
 GCN + Medoid  & $67.8_{3.2}$  & $\underline{\mathbf{71.8}}_{3.2}$ & $\underline{60.5}_{3.5}$ & $58.9_{4.8}$          & $\underline{62.0}_{9.7}$ & $60.5_{6.2}$           & $25.8_{2.0}$ & $20.0_{7.8}$                      & $17.6_{2.6}$& $14.8_{7.6}$\\
 \hline         
 GPN + Random  & $64.4_{5.0}$  & $65.6_{4.2}$                      & $52.6_{4.4}$ & $52.9_{4.8}$                      & $66.9_{8.5}$ & $67.5_{2.7}$                       & $48.0_{3.5}$ & $42.6_{2.4}$                      & $40.0_{3.1}$& $39.9_{1.4}$\\
 GPN + Entropy & $63.2_{6.1}$  & $64.9_{4.4}$                      & $47.2_{6.3}$ & $53.9_{5.6}$                      & $64.7_{5.9}$ & $64.8_{4.7}$                       & $\underline{48.8}_{1.1}$ & $42.6_{1.7}$          & $40.5_{2.1}$& $39.3_{3.3}$\\
 GPN + PageRank& $62.0_{6.3}$  & $66.3_{3.8}$                      & $52.8_{6.7}$ & $52.8_{3.2}$                      & $65.8_{4.3}$ & $62.9_{6.0}$                       & $46.6_{0.9}$ & $39.5_{0.9}$                      & $42.1_{2.0}$& $29.2_{2.4}$\\
 GPN + Medoid  & $65.6_{5.1}$  & $\underline{69.0}_{2.7}$          & $60.3_{4.6}$ & $\underline{\mathbf{60.8}}_{5.7}$ & $\underline{\mathbf{68.1}}_{5.1}$  & $65.4_{6.9}$ & $46.5_{3.2}$ & $20.2_{5.5}$                      & $41.0_{1.8}$& $\underline{46.9}_{1.5}$\\
\end{tabular}}

\label{table:results_1st}
%\end{table}

% stage 2
%\begin{table}
\vspace{0.3cm}
\caption{Results of the ``Unbalanced Sampling'' experiment, where we dropped the assumption of a class oracle. Measurements are given as average accuracy scores after five sampling rounds. Best accuracies in a block are highlighted with underlines, while the best accuracy in a column is bold. ``+LP'' denotes additional label propagation.}
\centering
\scalebox{\scaleboxnumber}{
\begin{tabular}{l|ll|ll|ll|ll|ll}
               & Cora         & +LP                                & CiteSeer     & +LP                               & PubMed        & +LP                                & Reddit2      & +LP                               & arXiv  & +LP        \\
               \hline
 FeatProp      & $62.3_{3.6}$ &                                    & $51.8_{6.8}$ &                                   & $61.4_{6.2}$  &                                    & $38.1_{3.1}$ &                                   & $40.4_{5.4}$&             \\
 \hline
 LP + Random   & $\underline{52.4}_{6.2}$ & $52.4_{6.2}$           & $34.9_{3.8}$ & $34.9_{3.8}$                      & $\underline{58.6}_{3.2}$  & $57.6_{4.2}$                       & $\underline{\mathbf{83.9}}_{0.4}$ & $83.9_{0.4}$                      & $\underline{55.2}_{0.7}$& $55.2_{0.7}$\\
 LP + Entropy  & $52.4_{6.2}$ & $52.4_{6.2}$                       & $35.0_{3.8}$ & $35.0_{3.8}$                      & $58.0_{3.3}$  & $57.8_{4.6}$                       & $83.9_{0.4}$ & $83.9_{0.4}$                      & $55.2_{0.7}$& $55.2_{0.7}$\\
 LP + PageRank & $51.1_{4.3}$ & $51.1_{4.3}$                       & $\underline{36.1}_{4.3}$ & $36.1_{4.3}$          & $58.1_{7.5}$  & $58.5_{7.4}$                       & $83.9_{1.2}$ & $83.9_{1.2}$                      & $55.1_{0.7}$& $55.1_{0.7}$\\
 LP + Medoid   & $51.7_{4.8}$ & $51.7_{4.8}$                       & $34.1_{4.7}$ & $34.1_{4.7}$                      & $55.7_{9.6}$  & $55.5_{9.4}$                       & $83.3_{1.2}$ & $83.3_{1.2}$                      & $54.4_{0.8}$& $54.6_{0.7}$\\
 \hline
 GCN + Random  & $\underline{59.3}_{7.6}$ & $55.0_{8.1}$           & $39.2_{9.4}$ & $43.3_{7.8}$                      & $45.2_{7.9}$  & $45.5_{8.7}$                       & $20.3_{3.9}$ & $23.5_{4.1}$                      & $24.3_{6.7}$& $27.6_{0.9}$\\
 GCN + Entropy & $54.2_{7.4}$ & $54.4_{10.5}$                      & $43.8_{5.2}$ & $42.3_{8.5}$                      & $47.1_{9.1}$  & $49.6_{10.3}$                      & $19.6_{4.1}$ & $21.4_{4.2}$                      & $27.6_{1.0}$& $27.8_{0.5}$\\
 GCN + PageRank& $52.3_{8.5}$ & $44.9_{11.2}$                      & $43.7_{7.6}$ & $41.6_{8.8}$                      & $49.3_{9.5}$  & $47.3_{9.0}$                       & $20.1_{4.9}$ & $22.2_{3.8}$                      & $25.7_{7.8}$& $25.2_{5.2}$\\
 GCN + Medoid  & $52.1_{7.1}$ & $52.2_{12.4}$                      & $\underline{45.4}_{5.6}$ & $44.1_{7.2}$          & $41.5_{3.2}$  & $\underline{50.8}_{10.5}$          & $17.5_{3.9}$ & $\underline{23.9}_{4.7}$          & $25.5_{8.5}$& $\underline{28.1}_{0.1}$\\
 \hline
 GPN + Random  & $64.0_{4.5}$ & $64.3_{6.9}$                       & $48.8_{8.0}$ & $55.1_{2.6}$                      & $56.1_{12.7}$ & $67.0_{6.4}$                       & $39.2_{6.8}$ & $\underline{46.8}_{1.4}$          & $53.5_{1.2}$& $50.4_{2.0}$\\
 GPN + Entropy & $65.5_{4.5}$ & $65.1_{5.5}$                       & $51.4_{4.8}$ & $55.5_{2.6}$                      & $56.6_{8.5}$  & $63.2_{5.7}$                       & $39.6_{5.9}$ & $42.1_{1.8}$                      & $53.7_{2.1}$& $50.3_{1.5}$\\
 GPN + PageRank& $63.8_{3.9}$ & $64.8_{4.4}$                       & $53.1_{5.1}$ & $52.8_{6.6}$                      & $61.5_{8.1}$  & $65.8_{4.2}$                       & $41.3_{2.0}$ & $42.5_{4.8}$                      & $47.2_{3.8}$& $50.4_{1.3}$\\
 GPN + Medoid  & $65.6_{3.5}$ & $\underline{\mathbf{68.4}}_{4.0}$  & $55.5_{5.4}$ & $\underline{\mathbf{56.8}}_{3.8}$ & $63.4_{6.8}$  & $\underline{\mathbf{67.0}}_{7.7}$  & $42.2_{1.0}$ & $44.6_{4.3}$                      & $\underline{\mathbf{56.1}}_{0.9}$& $53.3_{1.6}$\\
\end{tabular}}

\label{table:results_2nd}
%\end{table}
\vspace{0.3cm}

% stage 3
%\begin{table}
\caption{Results of the ``Unknown Number of Classes'' experiment, where the amount of classes is estimated. Cora is estimated to have 11 classes, CiteSeer 13, and PubMed 7. The best accuracies in a block are highlighted with underlines, while the best accuracy in a column is bold. ``+LP'' denotes additional label propagation.}
\centering

\scalebox{\scaleboxnumber}{
\begin{tabular}{l|ll|ll|ll}
               & Cora         & +LP          & CiteSeer     & +LP         & PubMed & +LP\\
\hline
 FeatProp     & $62.3_{3.6}$ &              & $51.8_{6.8}$ &             & $61.4_{6.2}$&\\
 \hline
 LP + Random   & $53.5_{6.5}$ & $53.5_{6.5}$ & $34.4_{5.1}$ & $34.4_{5.1}$& $53.6_{6.7}$&$53.6_{6.7}$\\
 LP + Entropy  & $53.5_{6.5}$ & $53.5_{6.5}$ & $34.1_{4.8}$ & $34.1_{4.8}$& $53.3_{6.6}$&$53.3_{6.6}$\\
 LP + PageRank & $53.2_{3.3}$ & $53.2_{3.3}$ & $\underline{38.4}_{2.0}$ & $38.4_{2.0}$& $\underline{59.3}_{5.4}$&$59.3_{5.4}$\\
 LP + Medoid   & $\underline{56.0}_{5.7}$ & $56.0_{5.7}$ & $35.4_{4.0}$ & $35.4_{4.0}$& $55.4_{5.5}$&$55.4_{5.5}$\\
 \hline
 GCN + Random  & $49.7_{5.3}$ & $50.4_{9.8}$ & $44.8_{4.9}$ & $40.7_{7.6}$& $44.2_{6.6}$&$40.3_{7.9}$\\
 GCN + Entropy & $50.2_{7.9}$ & $49.4_{9.5}$ & $45.6_{6.9}$ & $44.2_{8.1}$& $44.1_{6.5}$&$38.4_{6.0}$\\
 GCN + PageRank& $54.5_{7.6}$ & $49.1_{8.5}$ & $42.0_{6.9}$ & $45.9_{6.0}$& $42.1_{5.9}$&$\underline{44.6}_{7.6}$\\
 GCN + Medoid  & $\underline{56.5}_{8.2}$ & $53.2_{7.9}$ & $47.4_{6.4}$ & $\underline{48.5}_{5.8}$& $41.0_{9.0}$&$42.7_{6.9}$\\
 \hline
 GPN + Random  & $\mathbf{\underline{67.2}}_{5.3}$ & $67.1_{4.3}$ & $54.0_{4.9}$ & $48.6_{5.9}$& $58.7_{8.0}$ &$62.4_{10.4}$ \\
 GPN + Entropy & $65.6_{5.0}$ & $65.7_{4.4}$ & $53.9_{4.5}$ & $49.6_{5.4}$& $56.8_{10.0}$&$60.3_{11.5}$\\
 GPN + PageRank& $65.1_{3.7}$ & $65.9_{2.4}$ & $50.2_{7.2}$ & $52.9_{5.3}$& $57.3_{6.5}$ &$\underline{\mathbf{64.8}}_{4.0}$\\
 GPN + Medoid  & $65.7_{3.0}$ & $65.1_{6.1}$ & $51.0_{7.6}$ & $\underline{\mathbf{56.9}}_{4.5}$& $53.7_{8.2}$ &$62.3_{8.5}$\\

\end{tabular}}

\label{table:results_3rd}
\end{table*}

In our ``Balanced Sampling'' experiment, we assumed an oracle to perfectly separate our vertices by their classes.
The results are presented in Table~\ref{table:results_1st}, where we show the average test accuracy after a budget of $B=|C|\cdot5$.
Label propagation outperforms the discriminative and prototypical model on the large datasets, ogb-arXiv and Reddit2.
The discriminative model performs well on the smaller Planetoid datasets.
In comparison to the prototypical model, it is outperformed by $5\%$.
On the large datasets, the discriminative networks are outperformed by our prototypical networks.
Note that FeatProp~\cite{wu2021active} assumes to know the number of classes and does not use the class oracle.

%\paragraph{K-Medoids Clustering}
In our ``Unbalanced Sampling'' experiment, we no longer assume the existence of an oracle but use $k$-medoids clustering on the vertex embeddings instead.
The results can be found in Table~\ref{table:results_2nd}.
The accuracy became lower on all smaller datasets, with Reddit2 being an exception, where $k$-medoids improves the accuracy.
The discriminative model lost $9\%$ accuracy on the Planetoid datasets, while the prototypical model lost only $1\%$ accuracy on average.

%\paragraph{Class estimation}
In our ``Unknown Number of Classes'' experiment, we additionally no longer assume to know the number of classes in a dataset, 
and\ifbool{AppendixDeviations}{, as seen in Appendix~\ref{sec:classest},} use an estimation instead.
The results can be found in Table~\ref{table:results_3rd}.
The accuracies stayed mostly within the error of the second experiment.

%\paragraph{Perfect clustering with label propagation}
The observations above hold true when adding label propagation to each method.
The label propagation baseline does not profit from the additional label propagation, while the discriminative network increases accuracy on the larger datasets but loses accuracy on the smaller datasets.
The prototypical network benefits the most from the additional label propagation but is also losing accuracy on smaller datasets.

%\paragraph{Deviations}
Label propagation is the most stable method, followed by the prototypical network. 
The discriminative network generally has the most deviation.
\ifbool{AppendixDeviations}{This is further analyzed in Appendix~\ref{sec:deviations}, where we see a plot of the second experiment using the Cora dataset, illustrating standard deviations using whisker bars.}{}
We observe that deviations decrease with a higher budget on the prototypical networks and with label propagation.
However, deviations increase when using discriminative networks.

%\paragraph{Active learning strategies}
When looking at the active learning strategies, the medoid strategy performs the best overall, while entropy and PageRank perform similarly to the random sampling strategy.
This holds true for all experiments. 
However, when comparing the first and second experiments, $k$-medoid sampling lost the most accuracy at an average of $8\%$ for Planetoid, while the other samplers stayed below $5\%$. 
\ifbool{AppendixDetailedResult}{
Plots for the different experiments, as well as the $k$-medoids clustering strategy from the Cora dataset, can be seen in Appendix~\ref{sec:plots}~and~\ref{sec:clustering_effectiveness}.
}
\ifbool{AppendixDeviations}{
Plots for the $k$-medoids clustering strategy on the Cora dataset can be seen in Appendix~\ref{sec:clustering_effectiveness}.
Average accuracies across all datasets can be found in Appendix~\ref{sec:averages}.}

\section{Discussion}
\label{sec:discussion}

In contrast to existing few-shot learners, we do not assume to have a pre-labeled dataset.
In this setting, we observe that prototypical models outperform discriminative models.
This suggests that prototypical networks are more effective in scenarios with a lack of labeled examples.
For this reason, label propagation on homophilic datasets improves the model by artificially adding labeled samples.

%\paragraph{Using pseudo-labels reduces accuracy}
Generally, the cluster-based sampling results in a more balanced class distribution compared to the original dataset.
%Appendix~\ref{sec:clustering_effectiveness} shows the sampled class imbalances of Cora and ogb-arXiv, when using $k$-medoids.
Since the cluster algorithm still did not manage to sample an equal amount of vertices per class, the results of the second experiment, where we used $k$-medoids clustering, are lower than those where the class oracle was used in the first experiment.
An exception is the ogb-arXiv dataset, where the results were generally better in the second experiment.
%\todo{Discuss alternatives to ill trained: Homophily (LP is strong), GCN needs more data / bad HP}
We assume that this is due to suboptimal hyperparameters, since the accuracy scores of ogb-arXiv are also worse than label propagation. 

%\paragraph{K-medoids clustering works well on embeddings from prototypical models}
When using label propagation or the discriminative model, we effectively perform $k$-medoids clustering on logits, which is sub-optimal, as these do not represent a metric space.
The embeddings from prototypical networks are designed to represent Euclidean space.
This enables $k$-medoids to efficiently exploit the embeddings\ifbool{AppendixDeviations}{, as further shown in Appendix~\ref{sec:clustering_effectiveness}}.
We observe that the clustering of the embeddings from the prototypical network reflected the true classes more than the embeddings from label propagation or the discriminative model.

%\paragraph{Medoid sampling performed the best}
When comparing the different active learning strategies, medoid performed best on average.
It managed to sample good representatives for each class, boosting model performance.
Unlike PageRank, which is a measure of graph centrality, $k$-medoids is a measure of embedding centrality.
This also causes the medoid sampler to be robust against outliers.
As such, PageRank does not take features from vertices into account but only operates on the edges.
The $k$-medoids clustering is applied to the model embeddings, effectively considering processed features and edges from the graph.
The entropy active learning strategy performed worse than the random strategy in some cases. 
This is because of our early training setting.
As the model is not trained yet, but randomly initialized, the entropy of the logits is also mostly random.
With entropy, we also sample vertices, which are less important to learn during the early stages. 
They usually do not represent their class as a whole but can represent special cases and outliers in their class.
This stresses the challenges for few-shot learning when an initially labeled dataset is not available and training has to be performed from a cold start.

%\paragraph{Additional label propagation performs good on prototypical networks}
The additional label propagation helps, especially for early iterations\ifbool{AppendixDetailedResult}{, as shown in the plots in Appendix~\ref{sec:plots}}{}.
The label propagation artificially increases the training set at the cost of potentially wrongly labeled samples.
When the amount of training samples gets larger, we observe that the influence of the propagated labels diminishes. 
In some instances, they disturbed the training during later stages, decreasing accuracy.
The discriminative model did not profit as much from the additional label propagation compared to the prototypical model.
This shows that our discriminative model is more sensitive to wrong labels than the prototypical model, which is more likely to overfit.
Still, at some point, the gain from the increased training set may be outweighed by the influence of wrong labels in the training set.

%\paragraph{Influence of Weak Annotator}
As it is uncommon for a human annotator to annotate perfectly, we investigate the effect of wrongly labeled vertices on the overall model accuracy. 
For this setting, we choose the Cora and Pubmed datasets and assume the number of classes to be known.
We use the $k$-medoids sampling strategy for pseudo-label assignment.
We use the prototypical model with the medoid active learning strategy and label propagation, as it provides the best accuracy scores for both datasets.
The human annotator assigns a wrong label to a given vertex with probability $\epsilon$.
The amount of wrong labels is tested using $\epsilon\in[0,50]\%$, in increments of $10\%$.
For the Cora dataset, the model performs well until human error provides a wrong label with a percentage of more than $10\%$.
After $5$ sampling rounds, we only lost $4\%$ on accuracy.
With $50\%$ of the labels wrong, the model was still able to learn from the annotations with an accuracy score of $40\%$, compared to random, which would be at $14\%$.
On the PubMed dataset, the model was more sensitive to wrong labels.
Only $10\%$ wrong labels caused the accuracy to drop by almost $9\%$ after $5$ sampling iterations.
The model was barely able to improve the initial accuracy score with more than $40\%$ of wrong labels.
Still, the model performed better than random with $50\%$ of wrong labels, managing an accuracy score of $49\%$, compared to $33\%$.
\ifbool{AppendixDeviations}{The detailed plots for this experiment are in Appendix~\ref{sec:imperfect}.}

\paragraph{Limitations and Future Work}
\label{sec:threattovalidity}
We assume human annotations provide perfect labels.
In real-world scenarios, such perfect human annotations do not exist.
Thus, we conduct an additional experiment involving an imperfect human annotator to assess the influence of label noise on the model's robustness.
We discovered that our model is robust against label noise when less than $20\%$ of the labels are wrong. 
Another limitation of our experiments is the fixation of hyperparameters, given our constrained setting. 
Sub-optimal hyperparameters can strongly impact model performance. 
To address this, using model soup~\cite{wortsman2022model} could alleviate the issue.
With this technique, identical models get trained with different hyperparameters.
After each training episode, the models are combined by averaging their weights.
We rely on the homophily assumption for the label propagation.
Real-world data can also be heterophile.
There is a need to investigate vertex selection strategies for few-shot learning on heterophilic graphs~\cite{chien2021adaptiveuniversalgeneralizedpagerank,maurya2021improvinggraphneuralnetworks,10.5555/3495724.3496377}.
Furthermore, our experiments are limited to the transductive setting.
This is usually not a problem since the data can always be collected automatically before the training without annotation.

While future work may extend this study by optimizing the annotation process, such as leveraging alternative GNN architectures and baselines, our primary objective is to address the challenge of annotating a graph entirely from scratch—starting with no labeled vertices. 
To this end, we not only explore in this work various sampling strategies but also establish a strong baseline by comparing our approach to FeatProp~\cite{wu2021active}, a well-regarded method for active learning on graphs. This deliberate focus allows us to systematically evaluate the impact of different sampling strategies in a setting where no prior labels exist.

\section{Conclusion}
\label{sec:conclusion}
We explore active few-shot learning for vertex classification without an initially labeled dataset.
Our experiments show that the models using prototypes are more robust and accurate than the discriminative variants. 
The prototypical models suffer less when replacing the class oracle with $k$-medoids clustering for pseudo-labels.
Furthermore, label propagation on homophilic datasets improves the overall performance by artificially expanding the training set.
However, this effect diminishes when more labeled samples are available.
The best active learning strategy was the medoid sampler.
\bibliographystyle{IEEEtran}
\bibliography{reference.bib}

\ifbool{AppendixDeviations}{
\clearpage
\appendix

\subsection{Training Procedure}
\label{sec:training}
The detailed training procedure of our models is described by Algorithm~\ref{alg:procedure}.

\begin{algorithm}
\caption{Pseudo-code of our procedure. We classify vertices in $V_u$ while iteratively moving vertices from $V_u$ to the initially empty $V_l$ using the sampler.}\label{alg:procedure}
\begin{algorithmic}
\Require Label budget $B$
\Require Set of unlabeled vertices $V_u$
\Require Model $M$
\State $V_l \gets \emptyset$ \Comment{Set of labeled vertices}
\While{$|V_l| \leq B$}
\State $V_l \gets V_l \cup \Call{SampleVertices}{M,V_u}$
\State $V_u \gets V_u \setminus V_l$
\State $V_p \gets \Call{PropagateLabels}{V_l, V_u}$ \Comment{Optional}
\State $V_t, V_v \gets \Call{SplitTrainValidation}{V_l \cup V_p}$
\State $M' \gets M$
\While{Model improved the last 4 iterations}
\State Train $M$ on $V_t$
\If{$M$ improved on $V_v$}
\State $M' \gets M$
\EndIf
\EndWhile
\State $M\gets M'$
\EndWhile
\end{algorithmic}
\end{algorithm}

\subsection{Homophily Measurement}
We calculate homophily using the class insensitive edge homophily ratio~\cite{lim2021large},
\begin{equation*}
    \label{eq:homophily}
    H_c = \frac{1}{C-1}\sum_{k=1}^{C}\text{max}\left(0, h_k - \frac{|C_k|}{|V|}\right),
\end{equation*}
 where $C$ denotes the number of classes, $|C_k|$ denotes the vertices of class $k$, and  $h_k$ denotes the edge homophily ratio of vertices of class $k$. 

\subsection{Result Averages}
\label{sec:averages}
Table~\ref{table:averages} shows the average accuracy per active learning strategy and model for each experiment.
For a better comparison of the experiments, Planetoid scores are given in brackets.
During all experiments, the entropy and random sampling strategies perform similarly.
The best strategy is the medoid sampling strategy, which consistently outperforms the PageRank sampling strategy.

However, when comparing the first and second experiments, $k$-medoid sampling lost the most accuracy at an average of 8\% for Planetoid, while the other samplers stayed below 5\%. 
\begin{table}
\caption{Averages for each experiment, Numbers in brackets are only for Planetoid datasets.}
\centering
\begin{tabular}{l|rrr}
  & Perfect clustering & K-medoids clustering& Class estimation\\
 \hline
 Total     &$50.7(56.3)$& $50.0(52.2)$ & $(51.4)$\\
 \hline
 LP        & $57.4(50.0)$ & $56.7(48.4)$ & $(48.5)$ \\
 GCN       & $41.5(56.9)$ & $38.3(47.9)$ & $(46.2)$ \\
 GPN       & $53.3(61.6)$ & $55.0(60.4)$ & $(59.5)$ \\
 \hline
 Random    & $49.7(54.4)$ & $49.9(51.9)$ & $(50.6)$ \\
 Entropy   & $49.5(54.2)$ & $50.0(52.2)$ & $(50.3)$ \\
 PageRank  & $50.9(56.3)$ & $49.5(51.8)$ & $(52.0)$ \\
 Medoid    & $52.8(60.8)$ & $50.4(52.5)$ & $(52.1)$ \\
\end{tabular}

\label{table:averages}
\end{table}

\subsection{Hyperparameters}
We choose the same hyperparameters for GCN, GPN, and LP based on literature~\cite{wu2021active,zeng2020graphsaint,huang2020combining,10.1145/3607144}.
Regular hyperparameter tuning is impossible in our setting since we do not have access to a validation set and the number of available labeled vertices per class stays very small (\eg only 5).
For FeatProp~\cite{wu2021active}, we used the same parameters as for our GCN.
The number of samples used for training in each iteration is one per class for the small datasets, \ie Cora, CiteSeer PubMed, and $10$ samples per class for the large datasets ogb-arXiv and Reddit2.
Our detailed hyperparameter values are in Table~\ref{table:hyperparams}.
\begin{table}
\caption{The final hyperparameters we used for our experiment}
\centering
\begin{tabular}{l|rrrrr}
 Dataset & Hidden size & Dropout & $\lambda$ & lr & Samples / class\\
 \hline
 CiteSeer   & $64$ & $0.5$ & $1$ & $0.005$ & $1$\\
 Cora       & $64$ & $0.5$ & $1$ & $0.005$ & $1$\\
 PubMed     & $64$ & $0.5$ & $1$ & $0.005$ & $1$\\
 Reddit2    & $64$ & $0.5$ & $1$ & $0.005$ & $10$\\
 ogb-arXiv  & $64$ & $0.5$ & $1$ & $0.005$ & $10$\\
\end{tabular}

\label{table:hyperparams}
\end{table}

\subsection{Class Estimation}
\label{sec:classest}
For our number of classes estimation, we compute embeddings with Deep Graph Infomax~\cite{vel2018deep}.
On these embeddings, we compute multiple $k$-means clusterings for $k\in[2,100]$ and determine the best $k$ by using the elbow method~\cite{Thorndike1953}.
This means each clustering is evaluated by distortion score, which is the sum of square distances from each point to the center to which it is assigned.
The best value of $k$ is determined by detecting the elbow of the distortion score series, which is the point of inflection on the curve. %Figure~\ref{fig:n_class_estimation} shows the curve for each dataset.
For Cora, we found $k=11$, CiteSeer is estimated to have $k=13$, and PubMed to have $k=7$ classes.
For the large datasets ogb-arXiv and Reddit2, the calculation of the distortion scores failed due to time constraints. The computation ran for 48 hours without finishing, and it is therefore excluded from the results.
%\FloatBarrier

\subsection{Imperfect Human Annotator}
In Figure~\ref{fig:imperfect}, we show the results of the imperfect human annotator experiment for the Cora and PubMed datasets.
The human annotator assigns an incorrect label to a vertex. 
The proportion of incorrect labels is evaluated for $\epsilon\in[0,50]\%$, increasing in increments of $10\%$.

The effects of label propagation for the imperfect human annotator experiment on the Cora dataset are shown in Figure~\ref{fig:imperfect_lp}.
We observe that label propagation can have a negative effect during later stages of training, especially with an imperfect human annotator.

\label{sec:imperfect}
\begin{figure}
\centering
\begin{subfigure}[t]{0.24\textwidth}
  \centering
  \captionsetup{margin=0.05cm}
  \resizebox{\textwidth}{!}{\includegraphics{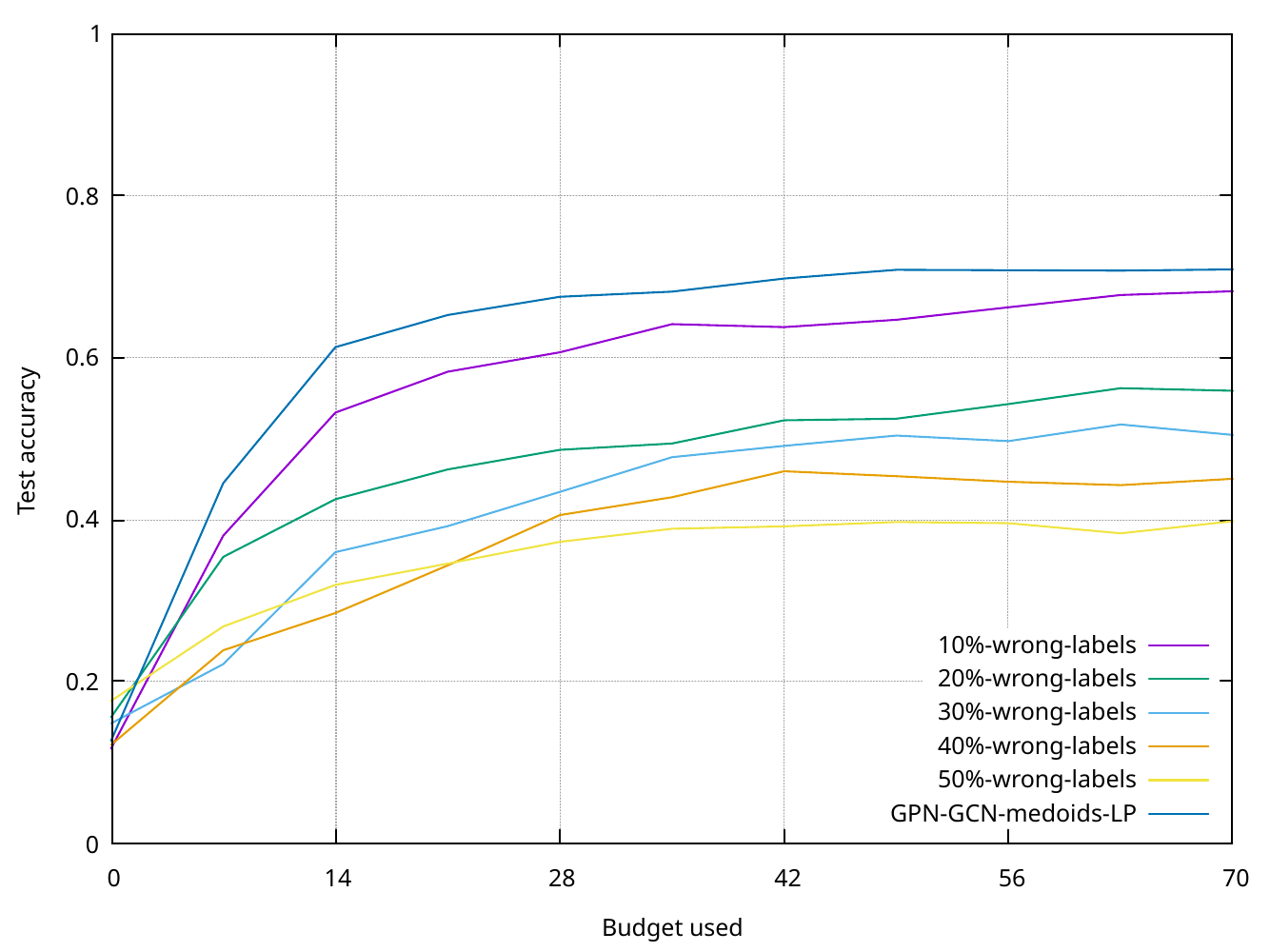}} %0.5625\textwidth
  \caption{Imperfect human annotator for Cora}
  \label{fig:imperfect_cora}
\end{subfigure}
\begin{subfigure}[t]{0.24\textwidth}
  \centering
  \captionsetup{margin=0.05cm}
  \resizebox{\textwidth}{!}{\includegraphics{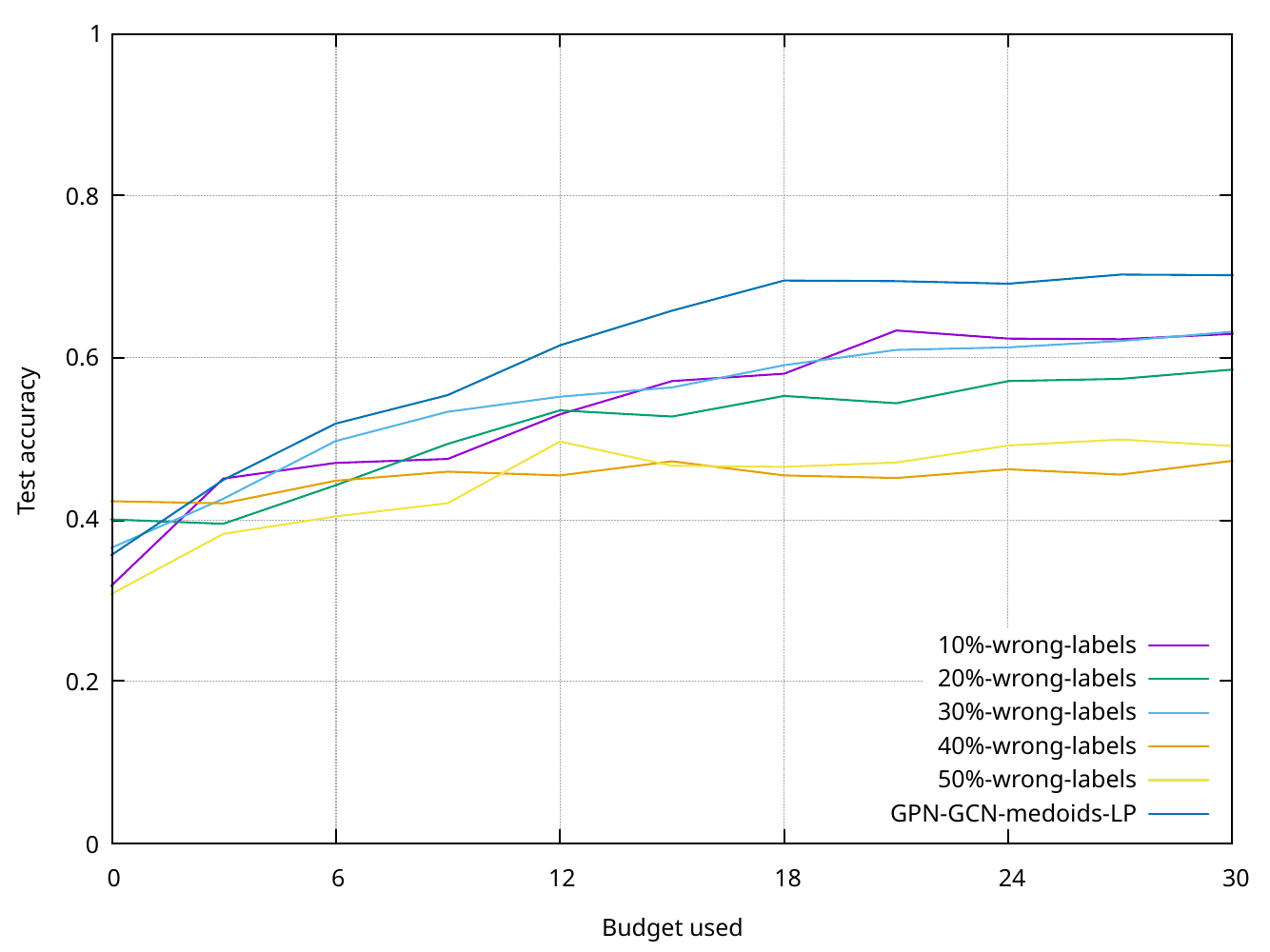}}
  \caption{Imperfect human annotator for PubMed}
  \label{fig:imperfect_pubmed}
\end{subfigure}
\caption{
Imperfect human Annotator experiment on the Cora and Pubmed dataset.
Using the prototypical model from the ``Unbalanced Sampling'' experiment with the medoid sub-sampling strategy and label propagation.}
    \label{fig:imperfect}
\end{figure}
\begin{figure}
\centering
\begin{subfigure}[t]{0.24\textwidth}
  \centering
  \captionsetup{margin=0.2cm}
  \includegraphics[width=\linewidth]{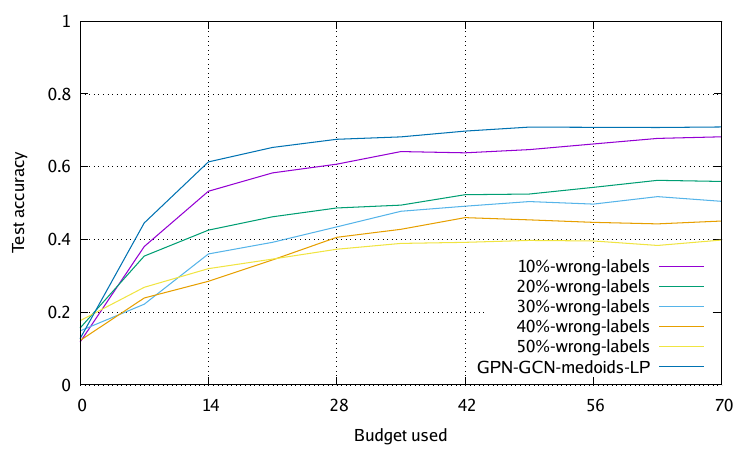}
  \caption{Imperfect human annotator, using label propagation.}
  \label{fig:imperfect_cora_2}
\end{subfigure}
\begin{subfigure}[t]{0.24\textwidth}
  \centering
  \captionsetup{margin=0.2cm}
  \includegraphics[width=\linewidth]{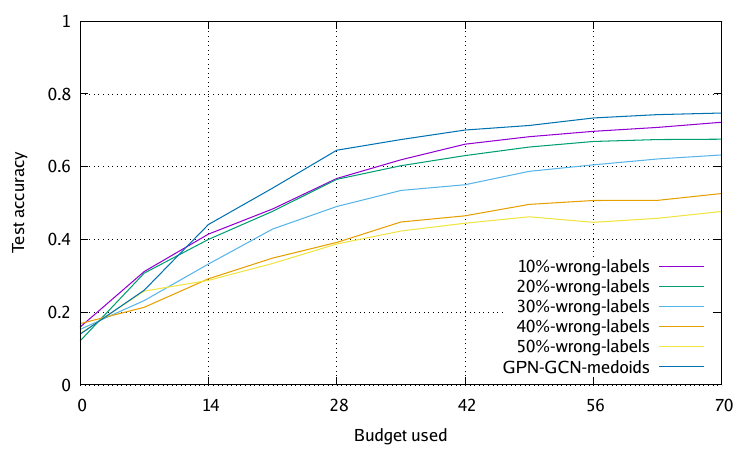}
  \caption{Imperfect human annotator, using no label propagation}
  \label{fig:imperfect_cora_nolp}
\end{subfigure}
\caption{
Imperfect human Annotator experiment on the Cora dataset, comparing the influence of the imperfect annotator on label propagation.
Using the prototypical model from the ``Unbalanced Sampling'' experiment with the medoid sub-sampling strategy and label propagation.}
    \label{fig:imperfect_lp}
\end{figure}

\ifbool{AppendixDeviations}{
\subsection{Standard Deviations}
\label{sec:deviations}
Figure~\ref{fig:single_runs} shows the standard deviations for the test accuracy on different models.
The models are trained on the Cora dataset and are combined with additional label propagation and the medoid sampling strategy.
Each of the three models starts with the same seed, consequent runs on the same model have different but predictable seeds.
}{}

\subsection{Clustering Effectiveness}
\label{sec:clustering_effectiveness}
Figure~\ref{fig:label_distribution_cora} %~and~\ref{fig:label_distribution} 
shows the effectiveness of $k$-medoids clustering, using the model embeddings during the ``Unbalanced Sampling'' experiment.
Replacing $k$-medoids clustering with a perfect clustering, as in the ``Balanced Sampling'' experiment, results in sampling the same number of vertices for each class.
The embeddings from the prototypical model are more suitable for $k$-medoids clustering than on the discriminative model.

%\FloatBarrier
\begin{figure}
\centering
  \begin{subfigure}{.22\textwidth}
  \captionsetup{margin=0.05cm}
  \centering
  \includegraphics[width=\textwidth]{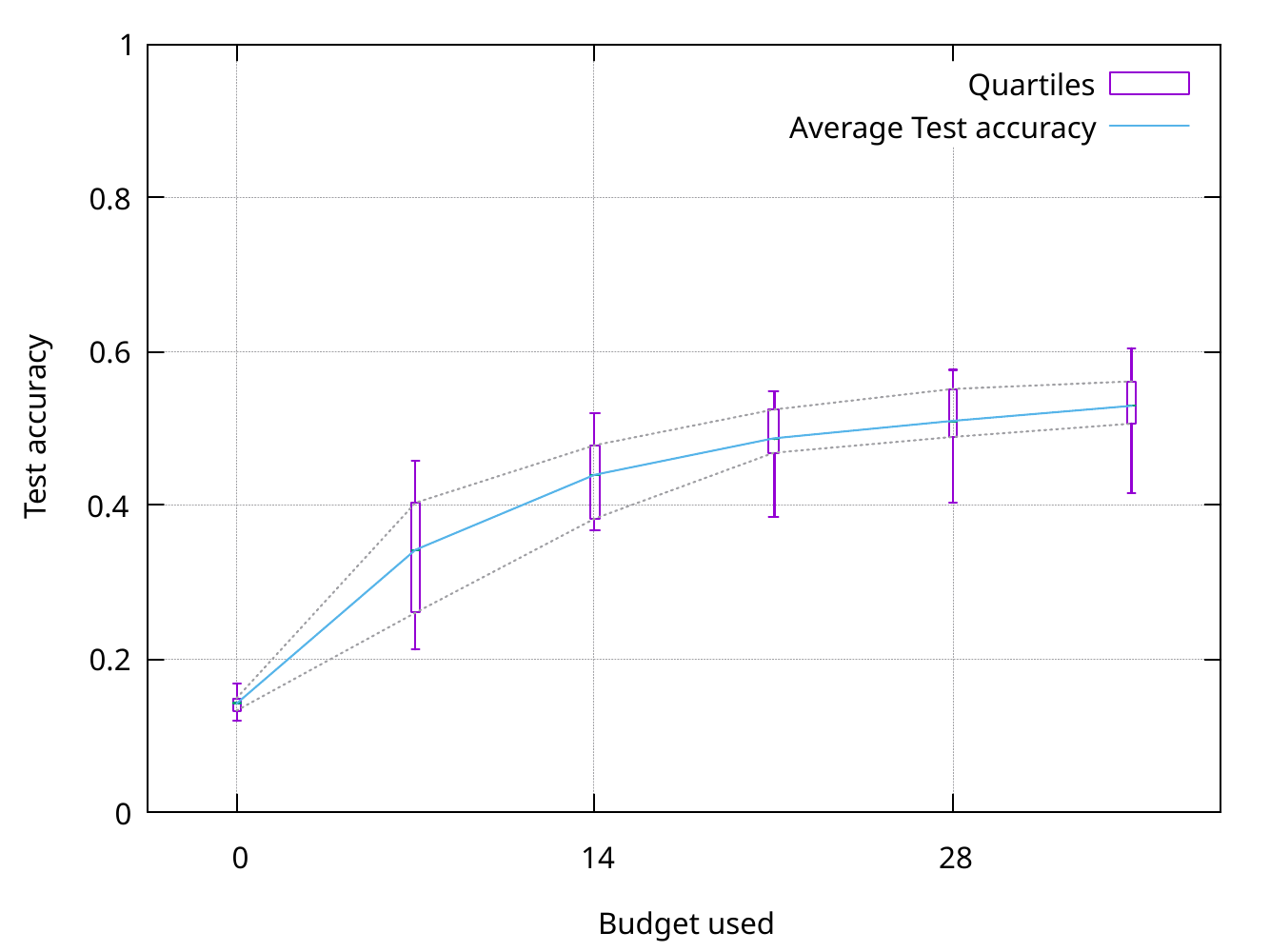}
  \caption{Deviations for label propagation.}
  \label{fig:single_runs_lp}
\end{subfigure}
\begin{subfigure}{.22\textwidth}
  \centering
  \captionsetup{margin=0.05cm}
  \includegraphics[width=\textwidth]{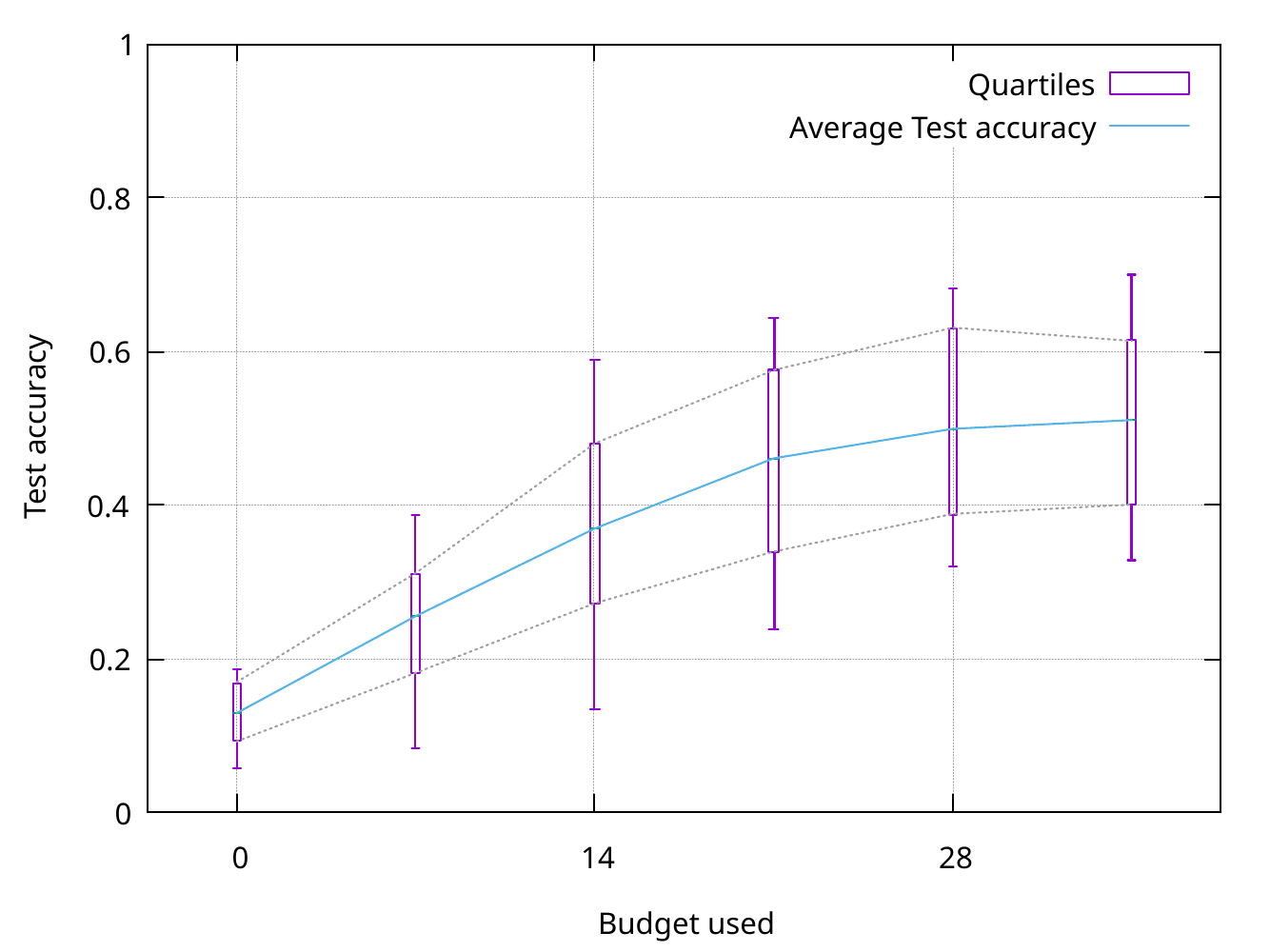}
  \caption{Deviations for discriminative networks.}
  \label{fig:single_runs_gcn}
\end{subfigure}
\begin{subfigure}{.22\textwidth}
  \centering
  \captionsetup{margin=0.05cm}
  \includegraphics[width=\textwidth]{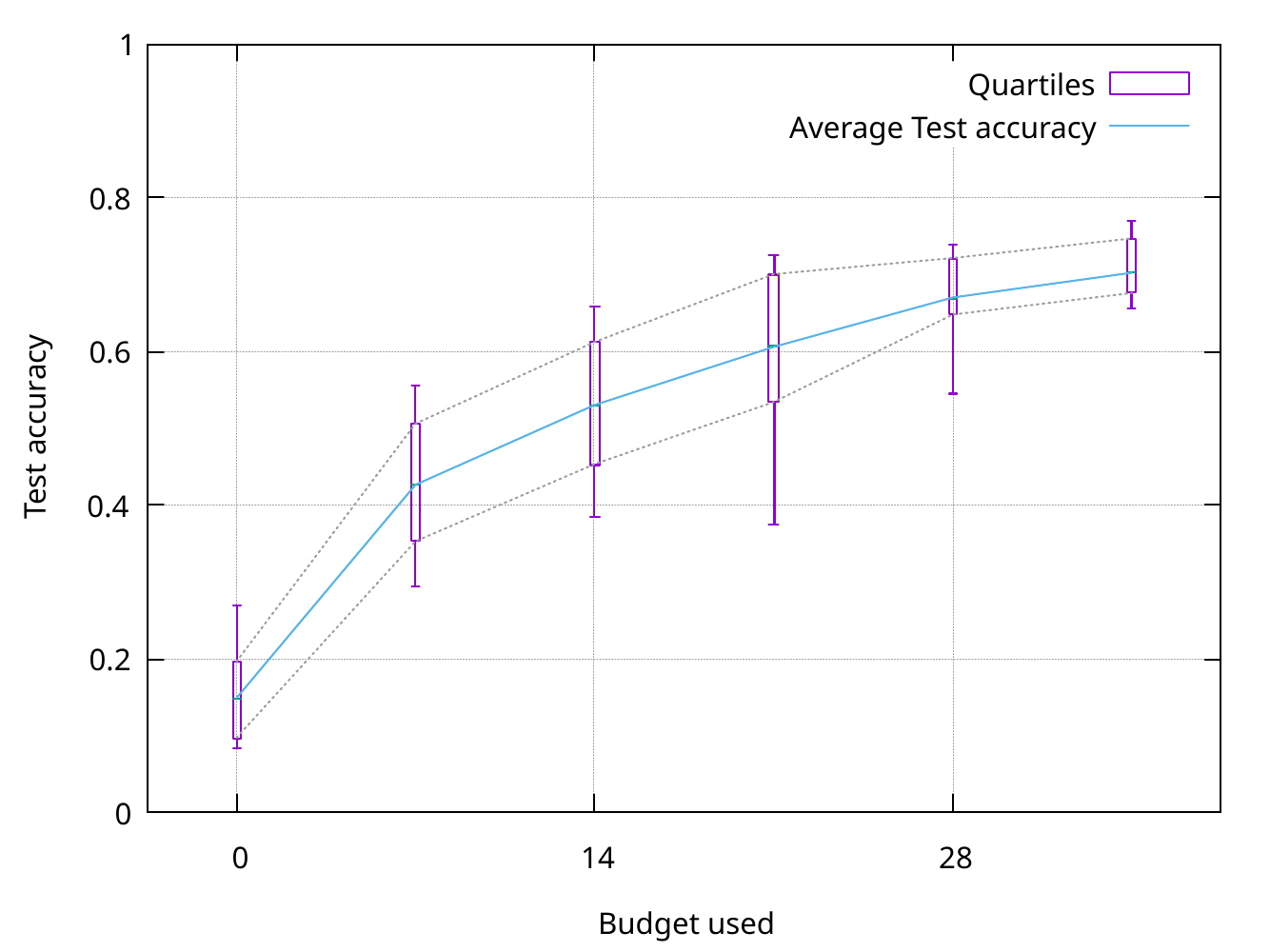}
  \caption{Deviations for prototypical networks.}
  \label{fig:single_runs_gpn}
\end{subfigure}
\caption{Overview of test accuracy deviations on Cora with 10 runs. Using the medoid sampling strategy together with label propagation.}
    \label{fig:single_runs}
\end{figure}

\begin{figure}
\centering
\begin{subfigure}{.22\textwidth}
  \centering
  \captionsetup{margin=0.05cm}
  \includegraphics[width=\textwidth]{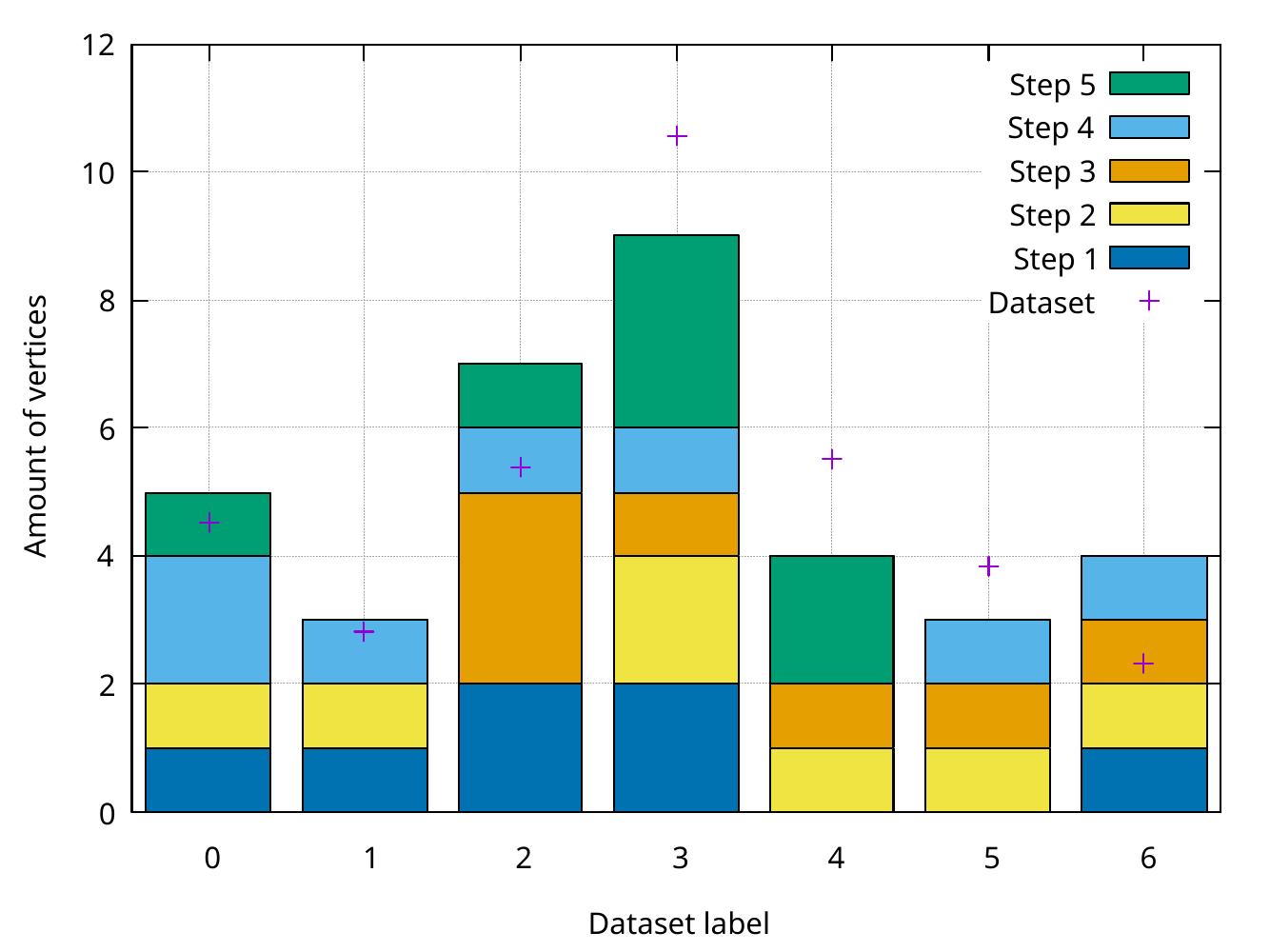}
  \caption{Sampling stats for the prototypical model}
  \label{fig:label_distribution_cora_gpn}
\end{subfigure}
\begin{subfigure}{.22\textwidth}
  \centering
  \captionsetup{margin=0.05cm}
  \includegraphics[width=\textwidth]{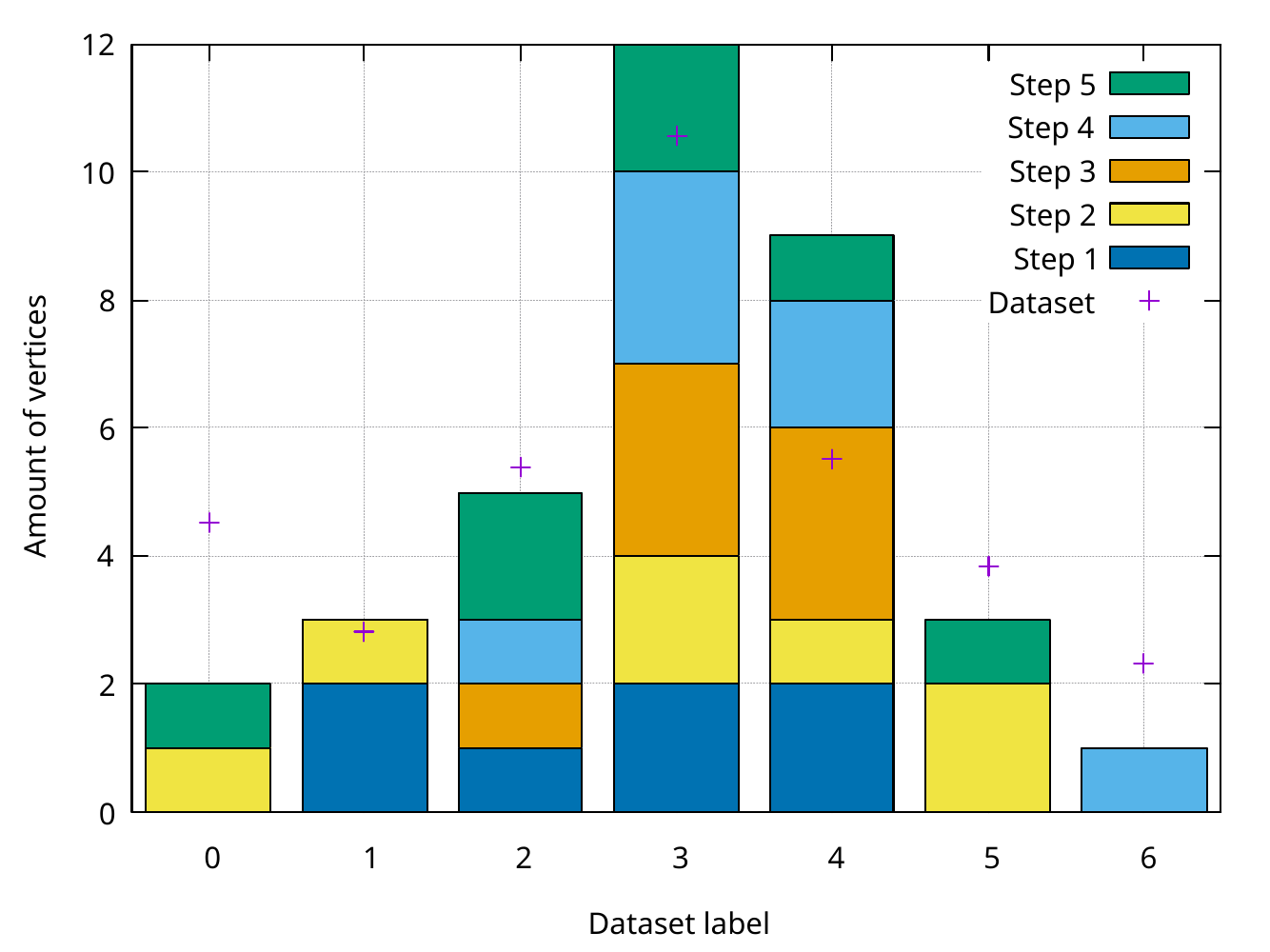}
  \caption{Sampling stats for the discriminative model}
  \label{fig:label_distribution_cora_pcn}
\end{subfigure}
\begin{subfigure}{.22\textwidth}
  \centering
  \captionsetup{margin=0.05cm}
  \includegraphics[width=\textwidth]{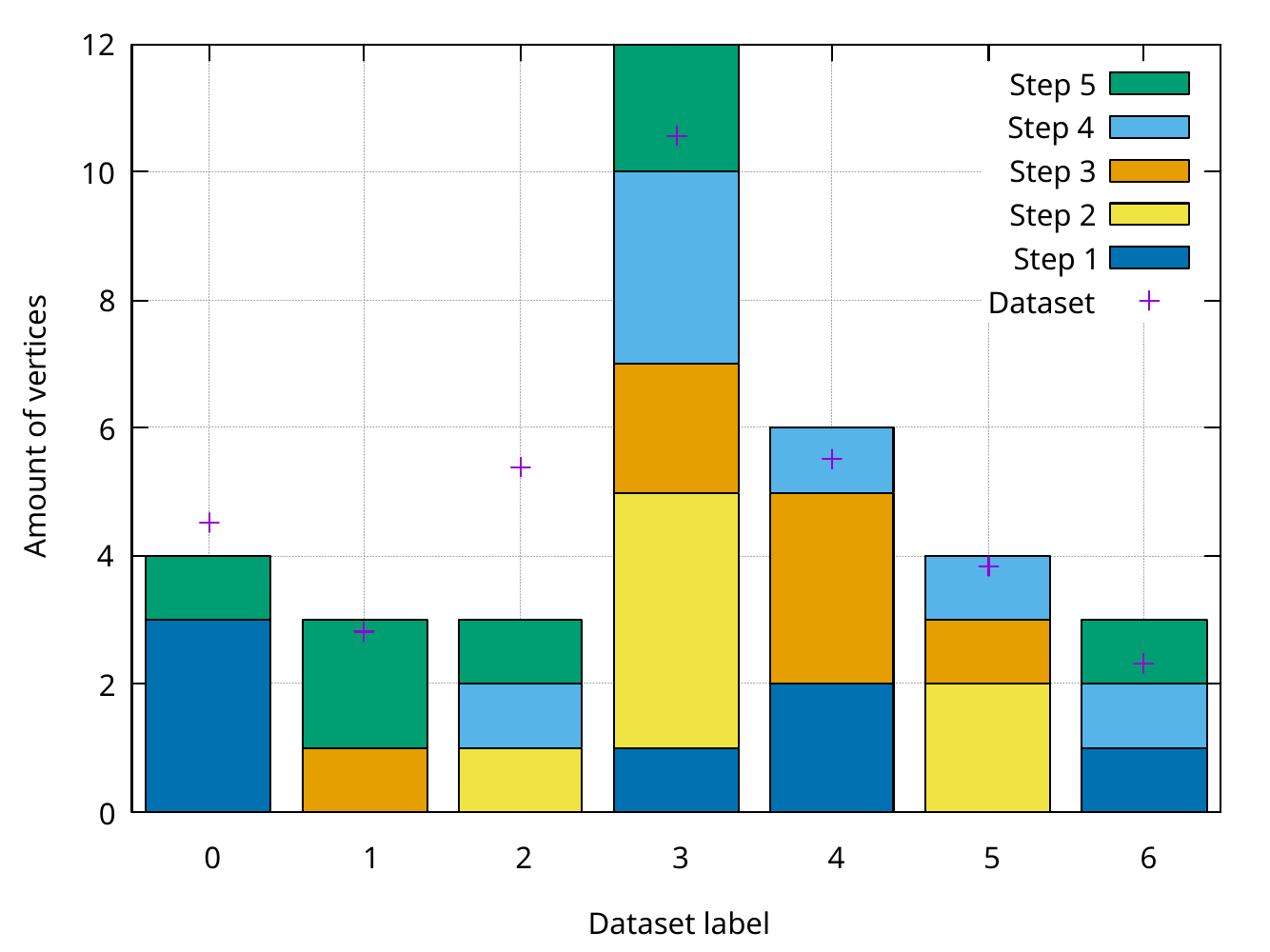}
  \caption{Sampling stats for label propagation}
  \label{fig:label_distribution_cora_lp}
\end{subfigure}
\caption{Overview of $k$-medoids sampling effectiveness on different models.
Using the Cora dataset with the medoid sub-sampling strategy and label propagation.}
\label{fig:label_distribution_cora}
\end{figure}

}

% ----------------------------------------------

\ifbool{AppendixDetailedResult}{
    \subsection{Detailed Results for the Cora Dataset}
    \label{sec:plots}
    Figure~\ref{fig:cora_stage1}, ~\ref{fig:cora_stage2}~and~\ref{fig:cora_stage3} show the plots for the different experiments on the Cora dataset.

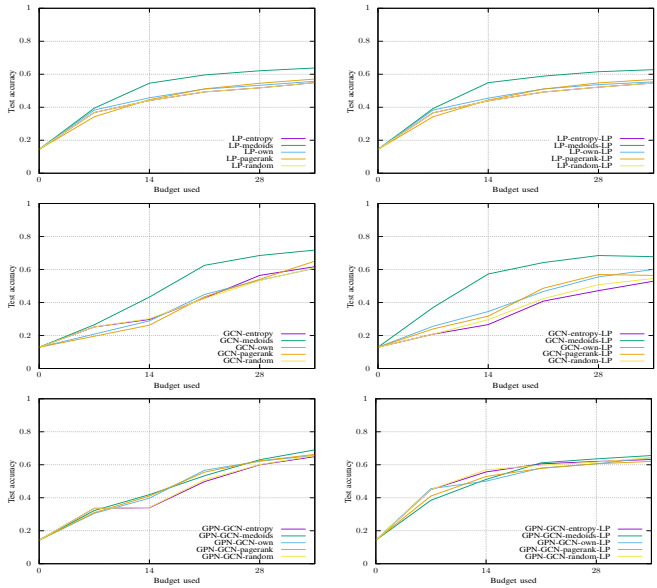
\begin{figure}[t]
    \begin{subfigure}[b]{.24\textwidth}
        \centering
        \resizebox{\textwidth}{0.11\textheight}{\input{plots/stage1/cora_lp}}
    \end{subfigure}
    \begin{subfigure}[b]{.24\textwidth}
        \centering
        \resizebox{\textwidth}{0.11\textheight}{\input{plots/stage1/cora_lp_lp}}
    \end{subfigure}
    \begin{subfigure}[b]{.24\textwidth}
        \centering
        \resizebox{\textwidth}{0.11\textheight}{\input{plots/stage1/cora_gcn}}
    \end{subfigure}
    \begin{subfigure}[b]{.24\textwidth}
        \centering
        \resizebox{\textwidth}{0.11\textheight}{\input{plots/stage1/cora_gcn_lp}}
    \end{subfigure}
    \begin{subfigure}[b]{.24\textwidth}
        \centering
        \resizebox{\textwidth}{0.11\textheight}{\input{plots/stage1/cora_gpn-gcn}}
    \end{subfigure}
    \begin{subfigure}[b]{.24\textwidth}
        \centering
        \resizebox{\textwidth}{0.11\textheight}{\input{plots/stage1/cora_gpn-gcn_lp}}
    \end{subfigure}
\caption{Test accuracy on Cora, assuming perfect clustering}
\label{fig:cora_stage1}
\end{figure}
\begin{figure}
    \begin{subfigure}[b]{.24\textwidth}
        \centering
        \resizebox{\textwidth}{0.11\textheight}{\input{plots/stage2/cora_lp}}
    \end{subfigure}
    \begin{subfigure}[b]{.24\textwidth}
        \centering
        \resizebox{\textwidth}{0.11\textheight}{\input{plots/stage2/cora_lp_lp}}
    \end{subfigure}
    \begin{subfigure}[b]{.24\textwidth}
        \centering
        \resizebox{\textwidth}{0.11\textheight}{\input{plots/stage2/cora_gcn}}
    \end{subfigure}
    \begin{subfigure}[b]{.24\textwidth}
        \centering
        \resizebox{\textwidth}{0.11\textheight}{\input{plots/stage2/cora_gcn_lp}}
    \end{subfigure}
    \begin{subfigure}[b]{.24\textwidth}
        \centering
        \resizebox{\textwidth}{0.11\textheight}{\input{plots/stage2/cora_gpn-gcn}}
    \end{subfigure}
    \begin{subfigure}[b]{.24\textwidth}
        \centering
        \resizebox{\textwidth}{0.11\textheight}{\input{plots/stage2/cora_gpn-gcn_lp}}
    \end{subfigure}
\caption{Test accuracy on Cora, using $k$-medoids clustering}
\label{fig:cora_stage2}
\end{figure}
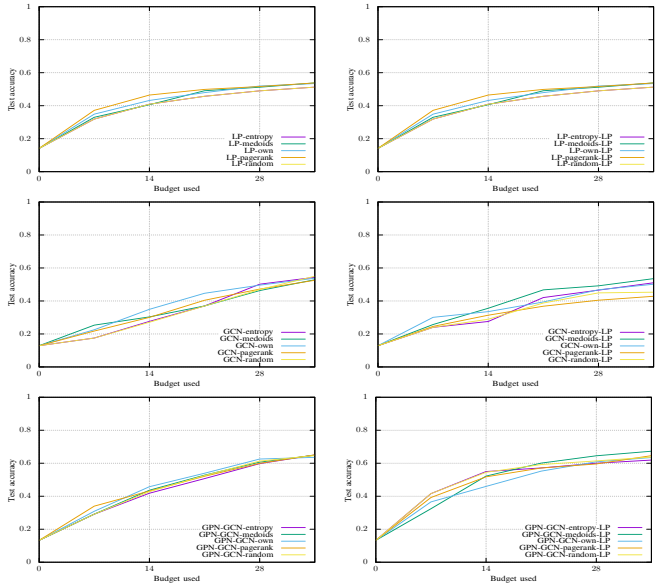
\begin{figure}
    \begin{subfigure}[b]{.24\textwidth}
        \centering
        \resizebox{\textwidth}{0.11\textheight}{\input{plots/stage3/cora_lp}}
    \end{subfigure}
    \begin{subfigure}[b]{.24\textwidth}
        \centering
        \resizebox{\textwidth}{0.11\textheight}{\input{plots/stage3/cora_lp_lp}}
    \end{subfigure}
    \begin{subfigure}[b]{.24\textwidth}
        \centering
        \resizebox{\textwidth}{0.11\textheight}{\input{plots/stage3/cora_gcn}}
    \end{subfigure}
    \begin{subfigure}[b]{.24\textwidth}
        \centering
        \resizebox{\textwidth}{0.11\textheight}{\input{plots/stage3/cora_gcn_lp}}
    \end{subfigure}
    \begin{subfigure}[b]{.24\textwidth}
        \centering
        \resizebox{\textwidth}{0.11\textheight}{\input{plots/stage3/cora_gpn}}
    \end{subfigure}
    \begin{subfigure}[b]{.24\textwidth}
        \centering
        \resizebox{\textwidth}{0.11\textheight}{\input{plots/stage3/cora_gpn_lp}}
    \end{subfigure}
\caption{Test accuracy on Cora, using $k$-medoids clustering and estimating classes}
\label{fig:cora_stage3}
\end{figure}
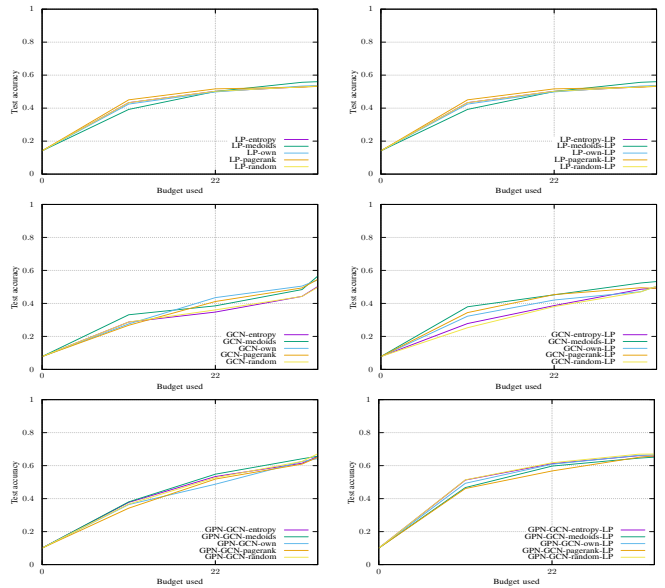
}{}

\end{document}

%% file: plots/stage1/cora_lp.tex
\begin{tikzpicture}[gnuplot]
%% generated with GNUPLOT 5.4p3 (Lua 5.4; terminal rev. Jun 2020, script rev. 115)
%% Sun 21 Jan 2024 04:26:38 PM CET
\path (0.000,0.000) rectangle (12.500,8.750);
\gpcolor{color=gp lt color axes}
\gpsetlinetype{gp lt axes}
\gpsetdashtype{gp dt axes}
\gpsetlinewidth{0.50}
\draw[gp path] (1.320,0.985)--(11.947,0.985);
\gpcolor{color=gp lt color border}
\gpsetlinetype{gp lt border}
\gpsetdashtype{gp dt solid}
\gpsetlinewidth{1.00}
\draw[gp path] (1.320,0.985)--(1.500,0.985);
\draw[gp path] (11.947,0.985)--(11.767,0.985);
\node[gp node right] at (1.136,0.985) {$0$};
\gpcolor{color=gp lt color axes}
\gpsetlinetype{gp lt axes}
\gpsetdashtype{gp dt axes}
\gpsetlinewidth{0.50}
\draw[gp path] (1.320,2.476)--(8.455,2.476);
\draw[gp path] (11.763,2.476)--(11.947,2.476);
\gpcolor{color=gp lt color border}
\gpsetlinetype{gp lt border}
\gpsetdashtype{gp dt solid}
\gpsetlinewidth{1.00}
\draw[gp path] (1.320,2.476)--(1.500,2.476);
\draw[gp path] (11.947,2.476)--(11.767,2.476);
\node[gp node right] at (1.136,2.476) {$0.2$};
\gpcolor{color=gp lt color axes}
\gpsetlinetype{gp lt axes}
\gpsetdashtype{gp dt axes}
\gpsetlinewidth{0.50}
\draw[gp path] (1.320,3.967)--(11.947,3.967);
\gpcolor{color=gp lt color border}
\gpsetlinetype{gp lt border}
\gpsetdashtype{gp dt solid}
\gpsetlinewidth{1.00}
\draw[gp path] (1.320,3.967)--(1.500,3.967);
\draw[gp path] (11.947,3.967)--(11.767,3.967);
\node[gp node right] at (1.136,3.967) {$0.4$};
\gpcolor{color=gp lt color axes}
\gpsetlinetype{gp lt axes}
\gpsetdashtype{gp dt axes}
\gpsetlinewidth{0.50}
\draw[gp path] (1.320,5.459)--(11.947,5.459);
\gpcolor{color=gp lt color border}
\gpsetlinetype{gp lt border}
\gpsetdashtype{gp dt solid}
\gpsetlinewidth{1.00}
\draw[gp path] (1.320,5.459)--(1.500,5.459);
\draw[gp path] (11.947,5.459)--(11.767,5.459);
\node[gp node right] at (1.136,5.459) {$0.6$};
\gpcolor{color=gp lt color axes}
\gpsetlinetype{gp lt axes}
\gpsetdashtype{gp dt axes}
\gpsetlinewidth{0.50}
\draw[gp path] (1.320,6.950)--(11.947,6.950);
\gpcolor{color=gp lt color border}
\gpsetlinetype{gp lt border}
\gpsetdashtype{gp dt solid}
\gpsetlinewidth{1.00}
\draw[gp path] (1.320,6.950)--(1.500,6.950);
\draw[gp path] (11.947,6.950)--(11.767,6.950);
\node[gp node right] at (1.136,6.950) {$0.8$};
\gpcolor{color=gp lt color axes}
\gpsetlinetype{gp lt axes}
\gpsetdashtype{gp dt axes}
\gpsetlinewidth{0.50}
\draw[gp path] (1.320,8.441)--(11.947,8.441);
\gpcolor{color=gp lt color border}
\gpsetlinetype{gp lt border}
\gpsetdashtype{gp dt solid}
\gpsetlinewidth{1.00}
\draw[gp path] (1.320,8.441)--(1.500,8.441);
\draw[gp path] (11.947,8.441)--(11.767,8.441);
\node[gp node right] at (1.136,8.441) {$1$};
\gpcolor{color=gp lt color axes}
\gpsetlinetype{gp lt axes}
\gpsetdashtype{gp dt axes}
\gpsetlinewidth{0.50}
\draw[gp path] (1.320,0.985)--(1.320,8.441);
\gpcolor{color=gp lt color border}
\gpsetlinetype{gp lt border}
\gpsetdashtype{gp dt solid}
\gpsetlinewidth{1.00}
\draw[gp path] (1.320,0.985)--(1.320,1.165);
\draw[gp path] (1.320,8.441)--(1.320,8.261);
\node[gp node center] at (1.320,0.677) {$0$};
\gpcolor{color=gp lt color axes}
\gpsetlinetype{gp lt axes}
\gpsetdashtype{gp dt axes}
\gpsetlinewidth{0.50}
\draw[gp path] (5.571,0.985)--(5.571,8.441);
\gpcolor{color=gp lt color border}
\gpsetlinetype{gp lt border}
\gpsetdashtype{gp dt solid}
\gpsetlinewidth{1.00}
\draw[gp path] (5.571,0.985)--(5.571,1.165);
\draw[gp path] (5.571,8.441)--(5.571,8.261);
\node[gp node center] at (5.571,0.677) {$14$};
\gpcolor{color=gp lt color axes}
\gpsetlinetype{gp lt axes}
\gpsetdashtype{gp dt axes}
\gpsetlinewidth{0.50}
\draw[gp path] (9.822,0.985)--(9.822,1.165);
\draw[gp path] (9.822,2.705)--(9.822,8.441);
\gpcolor{color=gp lt color border}
\gpsetlinetype{gp lt border}
\gpsetdashtype{gp dt solid}
\gpsetlinewidth{1.00}
\draw[gp path] (9.822,0.985)--(9.822,1.165);
\draw[gp path] (9.822,8.441)--(9.822,8.261);
\node[gp node center] at (9.822,0.677) {$28$};
\draw[gp path] (1.320,8.441)--(1.320,0.985)--(11.947,0.985)--(11.947,8.441)--cycle;
\node[gp node center,rotate=-270] at (0.292,4.713) {Test accuracy};
\node[gp node center] at (6.633,0.215) {Budget used};
\node[gp node right] at (10.479,2.551) {LP-entropy};
\gpcolor{rgb color={0.580,0.000,0.827}}
\draw[gp path] (10.663,2.551)--(11.579,2.551);
\draw[gp path] (1.320,2.063)--(3.445,3.729)--(5.571,4.270)--(7.696,4.660)--(9.822,4.844)%
  --(11.947,5.079);
\gpcolor{color=gp lt color border}
\node[gp node right] at (10.479,2.243) {LP-medoids};
\gpcolor{rgb color={0.000,0.620,0.451}}
\draw[gp path] (10.663,2.243)--(11.579,2.243);
\draw[gp path] (1.320,2.063)--(3.445,3.926)--(5.571,5.053)--(7.696,5.427)--(9.822,5.616)%
  --(11.947,5.742);
\gpcolor{color=gp lt color border}
\node[gp node right] at (10.479,1.935) {LP-own};
\gpcolor{rgb color={0.337,0.706,0.914}}
\draw[gp path] (10.663,1.935)--(11.579,1.935);
\draw[gp path] (1.320,2.063)--(3.445,3.841)--(5.571,4.395)--(7.696,4.788)--(9.822,4.960)%
  --(11.947,5.139);
\gpcolor{color=gp lt color border}
\node[gp node right] at (10.479,1.627) {LP-pagerank};
\gpcolor{rgb color={0.902,0.624,0.000}}
\draw[gp path] (10.663,1.627)--(11.579,1.627);
\draw[gp path] (1.320,2.063)--(3.445,3.537)--(5.571,4.310)--(7.696,4.799)--(9.822,5.052)%
  --(11.947,5.241);
\gpcolor{color=gp lt color border}
\node[gp node right] at (10.479,1.319) {LP-random};
\gpcolor{rgb color={0.941,0.894,0.259}}
\draw[gp path] (10.663,1.319)--(11.579,1.319);
\draw[gp path] (1.320,2.063)--(3.445,3.729)--(5.571,4.270)--(7.696,4.660)--(9.822,4.844)%
  --(11.947,5.079);
\gpcolor{color=gp lt color border}
\draw[gp path] (1.320,8.441)--(1.320,0.985)--(11.947,0.985)--(11.947,8.441)--cycle;
%% coordinates of the plot area
\gpdefrectangularnode{gp plot 1}{\pgfpoint{1.320cm}{0.985cm}}{\pgfpoint{11.947cm}{8.441cm}}
\end{tikzpicture}
%% gnuplot variables

%% file: plots/stage1/cora_lp_lp.tex
\begin{tikzpicture}[gnuplot]
%% generated with GNUPLOT 5.4p3 (Lua 5.4; terminal rev. Jun 2020, script rev. 115)
%% Sun 21 Jan 2024 04:27:13 PM CET
\path (0.000,0.000) rectangle (12.500,8.750);
\gpcolor{color=gp lt color axes}
\gpsetlinetype{gp lt axes}
\gpsetdashtype{gp dt axes}
\gpsetlinewidth{0.50}
\draw[gp path] (1.320,0.985)--(11.947,0.985);
\gpcolor{color=gp lt color border}
\gpsetlinetype{gp lt border}
\gpsetdashtype{gp dt solid}
\gpsetlinewidth{1.00}
\draw[gp path] (1.320,0.985)--(1.500,0.985);
\draw[gp path] (11.947,0.985)--(11.767,0.985);
\node[gp node right] at (1.136,0.985) {$0$};
\gpcolor{color=gp lt color axes}
\gpsetlinetype{gp lt axes}
\gpsetdashtype{gp dt axes}
\gpsetlinewidth{0.50}
\draw[gp path] (1.320,2.476)--(7.903,2.476);
\draw[gp path] (11.763,2.476)--(11.947,2.476);
\gpcolor{color=gp lt color border}
\gpsetlinetype{gp lt border}
\gpsetdashtype{gp dt solid}
\gpsetlinewidth{1.00}
\draw[gp path] (1.320,2.476)--(1.500,2.476);
\draw[gp path] (11.947,2.476)--(11.767,2.476);
\node[gp node right] at (1.136,2.476) {$0.2$};
\gpcolor{color=gp lt color axes}
\gpsetlinetype{gp lt axes}
\gpsetdashtype{gp dt axes}
\gpsetlinewidth{0.50}
\draw[gp path] (1.320,3.967)--(11.947,3.967);
\gpcolor{color=gp lt color border}
\gpsetlinetype{gp lt border}
\gpsetdashtype{gp dt solid}
\gpsetlinewidth{1.00}
\draw[gp path] (1.320,3.967)--(1.500,3.967);
\draw[gp path] (11.947,3.967)--(11.767,3.967);
\node[gp node right] at (1.136,3.967) {$0.4$};
\gpcolor{color=gp lt color axes}
\gpsetlinetype{gp lt axes}
\gpsetdashtype{gp dt axes}
\gpsetlinewidth{0.50}
\draw[gp path] (1.320,5.459)--(11.947,5.459);
\gpcolor{color=gp lt color border}
\gpsetlinetype{gp lt border}
\gpsetdashtype{gp dt solid}
\gpsetlinewidth{1.00}
\draw[gp path] (1.320,5.459)--(1.500,5.459);
\draw[gp path] (11.947,5.459)--(11.767,5.459);
\node[gp node right] at (1.136,5.459) {$0.6$};
\gpcolor{color=gp lt color axes}
\gpsetlinetype{gp lt axes}
\gpsetdashtype{gp dt axes}
\gpsetlinewidth{0.50}
\draw[gp path] (1.320,6.950)--(11.947,6.950);
\gpcolor{color=gp lt color border}
\gpsetlinetype{gp lt border}
\gpsetdashtype{gp dt solid}
\gpsetlinewidth{1.00}
\draw[gp path] (1.320,6.950)--(1.500,6.950);
\draw[gp path] (11.947,6.950)--(11.767,6.950);
\node[gp node right] at (1.136,6.950) {$0.8$};
\gpcolor{color=gp lt color axes}
\gpsetlinetype{gp lt axes}
\gpsetdashtype{gp dt axes}
\gpsetlinewidth{0.50}
\draw[gp path] (1.320,8.441)--(11.947,8.441);
\gpcolor{color=gp lt color border}
\gpsetlinetype{gp lt border}
\gpsetdashtype{gp dt solid}
\gpsetlinewidth{1.00}
\draw[gp path] (1.320,8.441)--(1.500,8.441);
\draw[gp path] (11.947,8.441)--(11.767,8.441);
\node[gp node right] at (1.136,8.441) {$1$};
\gpcolor{color=gp lt color axes}
\gpsetlinetype{gp lt axes}
\gpsetdashtype{gp dt axes}
\gpsetlinewidth{0.50}
\draw[gp path] (1.320,0.985)--(1.320,8.441);
\gpcolor{color=gp lt color border}
\gpsetlinetype{gp lt border}
\gpsetdashtype{gp dt solid}
\gpsetlinewidth{1.00}
\draw[gp path] (1.320,0.985)--(1.320,1.165);
\draw[gp path] (1.320,8.441)--(1.320,8.261);
\node[gp node center] at (1.320,0.677) {$0$};
\gpcolor{color=gp lt color axes}
\gpsetlinetype{gp lt axes}
\gpsetdashtype{gp dt axes}
\gpsetlinewidth{0.50}
\draw[gp path] (5.571,0.985)--(5.571,8.441);
\gpcolor{color=gp lt color border}
\gpsetlinetype{gp lt border}
\gpsetdashtype{gp dt solid}
\gpsetlinewidth{1.00}
\draw[gp path] (5.571,0.985)--(5.571,1.165);
\draw[gp path] (5.571,8.441)--(5.571,8.261);
\node[gp node center] at (5.571,0.677) {$14$};
\gpcolor{color=gp lt color axes}
\gpsetlinetype{gp lt axes}
\gpsetdashtype{gp dt axes}
\gpsetlinewidth{0.50}
\draw[gp path] (9.822,0.985)--(9.822,1.165);
\draw[gp path] (9.822,2.705)--(9.822,8.441);
\gpcolor{color=gp lt color border}
\gpsetlinetype{gp lt border}
\gpsetdashtype{gp dt solid}
\gpsetlinewidth{1.00}
\draw[gp path] (9.822,0.985)--(9.822,1.165);
\draw[gp path] (9.822,8.441)--(9.822,8.261);
\node[gp node center] at (9.822,0.677) {$28$};
\draw[gp path] (1.320,8.441)--(1.320,0.985)--(11.947,0.985)--(11.947,8.441)--cycle;
\node[gp node center,rotate=-270] at (0.292,4.713) {Test accuracy};
\node[gp node center] at (6.633,0.215) {Budget used};
\node[gp node right] at (10.479,2.551) {LP-entropy-LP};
\gpcolor{rgb color={0.580,0.000,0.827}}
\draw[gp path] (10.663,2.551)--(11.579,2.551);
\draw[gp path] (1.320,2.063)--(3.445,3.716)--(5.571,4.257)--(7.696,4.654)--(9.822,4.875)%
  --(11.947,5.062);
\gpcolor{color=gp lt color border}
\node[gp node right] at (10.479,2.243) {LP-medoids-LP};
\gpcolor{rgb color={0.000,0.620,0.451}}
\draw[gp path] (10.663,2.243)--(11.579,2.243);
\draw[gp path] (1.320,2.063)--(3.445,3.910)--(5.571,5.077)--(7.696,5.375)--(9.822,5.572)%
  --(11.947,5.666);
\gpcolor{color=gp lt color border}
\node[gp node right] at (10.479,1.935) {LP-own-LP};
\gpcolor{rgb color={0.337,0.706,0.914}}
\draw[gp path] (10.663,1.935)--(11.579,1.935);
\draw[gp path] (1.320,2.063)--(3.445,3.840)--(5.571,4.378)--(7.696,4.792)--(9.822,4.989)%
  --(11.947,5.115);
\gpcolor{color=gp lt color border}
\node[gp node right] at (10.479,1.627) {LP-pagerank-LP};
\gpcolor{rgb color={0.902,0.624,0.000}}
\draw[gp path] (10.663,1.627)--(11.579,1.627);
\draw[gp path] (1.320,2.063)--(3.445,3.538)--(5.571,4.292)--(7.696,4.789)--(9.822,5.069)%
  --(11.947,5.223);
\gpcolor{color=gp lt color border}
\node[gp node right] at (10.479,1.319) {LP-random-LP};
\gpcolor{rgb color={0.941,0.894,0.259}}
\draw[gp path] (10.663,1.319)--(11.579,1.319);
\draw[gp path] (1.320,2.063)--(3.445,3.716)--(5.571,4.257)--(7.696,4.654)--(9.822,4.875)%
  --(11.947,5.062);
\gpcolor{color=gp lt color border}
\draw[gp path] (1.320,8.441)--(1.320,0.985)--(11.947,0.985)--(11.947,8.441)--cycle;
%% coordinates of the plot area
\gpdefrectangularnode{gp plot 1}{\pgfpoint{1.320cm}{0.985cm}}{\pgfpoint{11.947cm}{8.441cm}}
\end{tikzpicture}
%% gnuplot variables

%% file: plots/stage1/cora_gcn.tex
\begin{tikzpicture}[gnuplot]
%% generated with GNUPLOT 5.4p3 (Lua 5.4; terminal rev. Jun 2020, script rev. 115)
%% Sun 21 Jan 2024 04:28:08 PM CET
\path (0.000,0.000) rectangle (12.500,8.750);
\gpcolor{color=gp lt color axes}
\gpsetlinetype{gp lt axes}
\gpsetdashtype{gp dt axes}
\gpsetlinewidth{0.50}
\draw[gp path] (1.320,0.985)--(11.947,0.985);
\gpcolor{color=gp lt color border}
\gpsetlinetype{gp lt border}
\gpsetdashtype{gp dt solid}
\gpsetlinewidth{1.00}
\draw[gp path] (1.320,0.985)--(1.500,0.985);
\draw[gp path] (11.947,0.985)--(11.767,0.985);
\node[gp node right] at (1.136,0.985) {$0$};
\gpcolor{color=gp lt color axes}
\gpsetlinetype{gp lt axes}
\gpsetdashtype{gp dt axes}
\gpsetlinewidth{0.50}
\draw[gp path] (1.320,2.476)--(8.271,2.476);
\draw[gp path] (11.763,2.476)--(11.947,2.476);
\gpcolor{color=gp lt color border}
\gpsetlinetype{gp lt border}
\gpsetdashtype{gp dt solid}
\gpsetlinewidth{1.00}
\draw[gp path] (1.320,2.476)--(1.500,2.476);
\draw[gp path] (11.947,2.476)--(11.767,2.476);
\node[gp node right] at (1.136,2.476) {$0.2$};
\gpcolor{color=gp lt color axes}
\gpsetlinetype{gp lt axes}
\gpsetdashtype{gp dt axes}
\gpsetlinewidth{0.50}
\draw[gp path] (1.320,3.967)--(11.947,3.967);
\gpcolor{color=gp lt color border}
\gpsetlinetype{gp lt border}
\gpsetdashtype{gp dt solid}
\gpsetlinewidth{1.00}
\draw[gp path] (1.320,3.967)--(1.500,3.967);
\draw[gp path] (11.947,3.967)--(11.767,3.967);
\node[gp node right] at (1.136,3.967) {$0.4$};
\gpcolor{color=gp lt color axes}
\gpsetlinetype{gp lt axes}
\gpsetdashtype{gp dt axes}
\gpsetlinewidth{0.50}
\draw[gp path] (1.320,5.459)--(11.947,5.459);
\gpcolor{color=gp lt color border}
\gpsetlinetype{gp lt border}
\gpsetdashtype{gp dt solid}
\gpsetlinewidth{1.00}
\draw[gp path] (1.320,5.459)--(1.500,5.459);
\draw[gp path] (11.947,5.459)--(11.767,5.459);
\node[gp node right] at (1.136,5.459) {$0.6$};
\gpcolor{color=gp lt color axes}
\gpsetlinetype{gp lt axes}
\gpsetdashtype{gp dt axes}
\gpsetlinewidth{0.50}
\draw[gp path] (1.320,6.950)--(11.947,6.950);
\gpcolor{color=gp lt color border}
\gpsetlinetype{gp lt border}
\gpsetdashtype{gp dt solid}
\gpsetlinewidth{1.00}
\draw[gp path] (1.320,6.950)--(1.500,6.950);
\draw[gp path] (11.947,6.950)--(11.767,6.950);
\node[gp node right] at (1.136,6.950) {$0.8$};
\gpcolor{color=gp lt color axes}
\gpsetlinetype{gp lt axes}
\gpsetdashtype{gp dt axes}
\gpsetlinewidth{0.50}
\draw[gp path] (1.320,8.441)--(11.947,8.441);
\gpcolor{color=gp lt color border}
\gpsetlinetype{gp lt border}
\gpsetdashtype{gp dt solid}
\gpsetlinewidth{1.00}
\draw[gp path] (1.320,8.441)--(1.500,8.441);
\draw[gp path] (11.947,8.441)--(11.767,8.441);
\node[gp node right] at (1.136,8.441) {$1$};
\gpcolor{color=gp lt color axes}
\gpsetlinetype{gp lt axes}
\gpsetdashtype{gp dt axes}
\gpsetlinewidth{0.50}
\draw[gp path] (1.320,0.985)--(1.320,8.441);
\gpcolor{color=gp lt color border}
\gpsetlinetype{gp lt border}
\gpsetdashtype{gp dt solid}
\gpsetlinewidth{1.00}
\draw[gp path] (1.320,0.985)--(1.320,1.165);
\draw[gp path] (1.320,8.441)--(1.320,8.261);
\node[gp node center] at (1.320,0.677) {$0$};
\gpcolor{color=gp lt color axes}
\gpsetlinetype{gp lt axes}
\gpsetdashtype{gp dt axes}
\gpsetlinewidth{0.50}
\draw[gp path] (5.571,0.985)--(5.571,8.441);
\gpcolor{color=gp lt color border}
\gpsetlinetype{gp lt border}
\gpsetdashtype{gp dt solid}
\gpsetlinewidth{1.00}
\draw[gp path] (5.571,0.985)--(5.571,1.165);
\draw[gp path] (5.571,8.441)--(5.571,8.261);
\node[gp node center] at (5.571,0.677) {$14$};
\gpcolor{color=gp lt color axes}
\gpsetlinetype{gp lt axes}
\gpsetdashtype{gp dt axes}
\gpsetlinewidth{0.50}
\draw[gp path] (9.822,0.985)--(9.822,1.165);
\draw[gp path] (9.822,2.705)--(9.822,8.441);
\gpcolor{color=gp lt color border}
\gpsetlinetype{gp lt border}
\gpsetdashtype{gp dt solid}
\gpsetlinewidth{1.00}
\draw[gp path] (9.822,0.985)--(9.822,1.165);
\draw[gp path] (9.822,8.441)--(9.822,8.261);
\node[gp node center] at (9.822,0.677) {$28$};
\draw[gp path] (1.320,8.441)--(1.320,0.985)--(11.947,0.985)--(11.947,8.441)--cycle;
\node[gp node center,rotate=-270] at (0.292,4.713) {Test accuracy};
\node[gp node center] at (6.633,0.215) {Budget used};
\node[gp node right] at (10.479,2.551) {GCN-entropy};
\gpcolor{rgb color={0.580,0.000,0.827}}
\draw[gp path] (10.663,2.551)--(11.579,2.551);
\draw[gp path] (1.320,1.946)--(3.445,2.862)--(5.571,3.200)--(7.696,4.166)--(9.822,5.192)%
  --(11.947,5.586);
\gpcolor{color=gp lt color border}
\node[gp node right] at (10.479,2.243) {GCN-medoids};
\gpcolor{rgb color={0.000,0.620,0.451}}
\draw[gp path] (10.663,2.243)--(11.579,2.243);
\draw[gp path] (1.320,1.946)--(3.445,2.961)--(5.571,4.213)--(7.696,5.648)--(9.822,6.093)%
  --(11.947,6.336);
\gpcolor{color=gp lt color border}
\node[gp node right] at (10.479,1.935) {GCN-own};
\gpcolor{rgb color={0.337,0.706,0.914}}
\draw[gp path] (10.663,1.935)--(11.579,1.935);
\draw[gp path] (1.320,1.946)--(3.445,2.538)--(5.571,3.137)--(7.696,4.332)--(9.822,4.975)%
  --(11.947,5.513);
\gpcolor{color=gp lt color border}
\node[gp node right] at (10.479,1.627) {GCN-pagerank};
\gpcolor{rgb color={0.902,0.624,0.000}}
\draw[gp path] (10.663,1.627)--(11.579,1.627);
\draw[gp path] (1.320,1.946)--(3.445,2.442)--(5.571,2.942)--(7.696,4.231)--(9.822,5.015)%
  --(11.947,5.837);
\gpcolor{color=gp lt color border}
\node[gp node right] at (10.479,1.319) {GCN-random};
\gpcolor{rgb color={0.941,0.894,0.259}}
\draw[gp path] (10.663,1.319)--(11.579,1.319);
\draw[gp path] (1.320,1.946)--(3.445,2.862)--(5.571,3.238)--(7.696,4.150)--(9.822,4.955)%
  --(11.947,5.533);
\gpcolor{color=gp lt color border}
\draw[gp path] (1.320,8.441)--(1.320,0.985)--(11.947,0.985)--(11.947,8.441)--cycle;
%% coordinates of the plot area
\gpdefrectangularnode{gp plot 1}{\pgfpoint{1.320cm}{0.985cm}}{\pgfpoint{11.947cm}{8.441cm}}
\end{tikzpicture}
%% gnuplot variables

%% file: plots/stage1/cora_gcn_lp.tex
\begin{tikzpicture}[gnuplot]
%% generated with GNUPLOT 5.4p3 (Lua 5.4; terminal rev. Jun 2020, script rev. 115)
%% Sun 21 Jan 2024 04:28:34 PM CET
\path (0.000,0.000) rectangle (12.500,8.750);
\gpcolor{color=gp lt color axes}
\gpsetlinetype{gp lt axes}
\gpsetdashtype{gp dt axes}
\gpsetlinewidth{0.50}
\draw[gp path] (1.320,0.985)--(11.947,0.985);
\gpcolor{color=gp lt color border}
\gpsetlinetype{gp lt border}
\gpsetdashtype{gp dt solid}
\gpsetlinewidth{1.00}
\draw[gp path] (1.320,0.985)--(1.500,0.985);
\draw[gp path] (11.947,0.985)--(11.767,0.985);
\node[gp node right] at (1.136,0.985) {$0$};
\gpcolor{color=gp lt color axes}
\gpsetlinetype{gp lt axes}
\gpsetdashtype{gp dt axes}
\gpsetlinewidth{0.50}
\draw[gp path] (1.320,2.476)--(7.719,2.476);
\draw[gp path] (11.763,2.476)--(11.947,2.476);
\gpcolor{color=gp lt color border}
\gpsetlinetype{gp lt border}
\gpsetdashtype{gp dt solid}
\gpsetlinewidth{1.00}
\draw[gp path] (1.320,2.476)--(1.500,2.476);
\draw[gp path] (11.947,2.476)--(11.767,2.476);
\node[gp node right] at (1.136,2.476) {$0.2$};
\gpcolor{color=gp lt color axes}
\gpsetlinetype{gp lt axes}
\gpsetdashtype{gp dt axes}
\gpsetlinewidth{0.50}
\draw[gp path] (1.320,3.967)--(11.947,3.967);
\gpcolor{color=gp lt color border}
\gpsetlinetype{gp lt border}
\gpsetdashtype{gp dt solid}
\gpsetlinewidth{1.00}
\draw[gp path] (1.320,3.967)--(1.500,3.967);
\draw[gp path] (11.947,3.967)--(11.767,3.967);
\node[gp node right] at (1.136,3.967) {$0.4$};
\gpcolor{color=gp lt color axes}
\gpsetlinetype{gp lt axes}
\gpsetdashtype{gp dt axes}
\gpsetlinewidth{0.50}
\draw[gp path] (1.320,5.459)--(11.947,5.459);
\gpcolor{color=gp lt color border}
\gpsetlinetype{gp lt border}
\gpsetdashtype{gp dt solid}
\gpsetlinewidth{1.00}
\draw[gp path] (1.320,5.459)--(1.500,5.459);
\draw[gp path] (11.947,5.459)--(11.767,5.459);
\node[gp node right] at (1.136,5.459) {$0.6$};
\gpcolor{color=gp lt color axes}
\gpsetlinetype{gp lt axes}
\gpsetdashtype{gp dt axes}
\gpsetlinewidth{0.50}
\draw[gp path] (1.320,6.950)--(11.947,6.950);
\gpcolor{color=gp lt color border}
\gpsetlinetype{gp lt border}
\gpsetdashtype{gp dt solid}
\gpsetlinewidth{1.00}
\draw[gp path] (1.320,6.950)--(1.500,6.950);
\draw[gp path] (11.947,6.950)--(11.767,6.950);
\node[gp node right] at (1.136,6.950) {$0.8$};
\gpcolor{color=gp lt color axes}
\gpsetlinetype{gp lt axes}
\gpsetdashtype{gp dt axes}
\gpsetlinewidth{0.50}
\draw[gp path] (1.320,8.441)--(11.947,8.441);
\gpcolor{color=gp lt color border}
\gpsetlinetype{gp lt border}
\gpsetdashtype{gp dt solid}
\gpsetlinewidth{1.00}
\draw[gp path] (1.320,8.441)--(1.500,8.441);
\draw[gp path] (11.947,8.441)--(11.767,8.441);
\node[gp node right] at (1.136,8.441) {$1$};
\gpcolor{color=gp lt color axes}
\gpsetlinetype{gp lt axes}
\gpsetdashtype{gp dt axes}
\gpsetlinewidth{0.50}
\draw[gp path] (1.320,0.985)--(1.320,8.441);
\gpcolor{color=gp lt color border}
\gpsetlinetype{gp lt border}
\gpsetdashtype{gp dt solid}
\gpsetlinewidth{1.00}
\draw[gp path] (1.320,0.985)--(1.320,1.165);
\draw[gp path] (1.320,8.441)--(1.320,8.261);
\node[gp node center] at (1.320,0.677) {$0$};
\gpcolor{color=gp lt color axes}
\gpsetlinetype{gp lt axes}
\gpsetdashtype{gp dt axes}
\gpsetlinewidth{0.50}
\draw[gp path] (5.571,0.985)--(5.571,8.441);
\gpcolor{color=gp lt color border}
\gpsetlinetype{gp lt border}
\gpsetdashtype{gp dt solid}
\gpsetlinewidth{1.00}
\draw[gp path] (5.571,0.985)--(5.571,1.165);
\draw[gp path] (5.571,8.441)--(5.571,8.261);
\node[gp node center] at (5.571,0.677) {$14$};
\gpcolor{color=gp lt color axes}
\gpsetlinetype{gp lt axes}
\gpsetdashtype{gp dt axes}
\gpsetlinewidth{0.50}
\draw[gp path] (9.822,0.985)--(9.822,1.165);
\draw[gp path] (9.822,2.705)--(9.822,8.441);
\gpcolor{color=gp lt color border}
\gpsetlinetype{gp lt border}
\gpsetdashtype{gp dt solid}
\gpsetlinewidth{1.00}
\draw[gp path] (9.822,0.985)--(9.822,1.165);
\draw[gp path] (9.822,8.441)--(9.822,8.261);
\node[gp node center] at (9.822,0.677) {$28$};
\draw[gp path] (1.320,8.441)--(1.320,0.985)--(11.947,0.985)--(11.947,8.441)--cycle;
\node[gp node center,rotate=-270] at (0.292,4.713) {Test accuracy};
\node[gp node center] at (6.633,0.215) {Budget used};
\node[gp node right] at (10.479,2.551) {GCN-entropy-LP};
\gpcolor{rgb color={0.580,0.000,0.827}}
\draw[gp path] (10.663,2.551)--(11.579,2.551);
\draw[gp path] (1.320,1.946)--(3.445,2.536)--(5.571,2.969)--(7.696,4.029)--(9.822,4.503)%
  --(11.947,4.931);
\gpcolor{color=gp lt color border}
\node[gp node right] at (10.479,2.243) {GCN-medoids-LP};
\gpcolor{rgb color={0.000,0.620,0.451}}
\draw[gp path] (10.663,2.243)--(11.579,2.243);
\draw[gp path] (1.320,1.946)--(3.445,3.742)--(5.571,5.256)--(7.696,5.773)--(9.822,6.091)%
  --(11.947,6.039);
\gpcolor{color=gp lt color border}
\node[gp node right] at (10.479,1.935) {GCN-own-LP};
\gpcolor{rgb color={0.337,0.706,0.914}}
\draw[gp path] (10.663,1.935)--(11.579,1.935);
\draw[gp path] (1.320,1.946)--(3.445,2.894)--(5.571,3.558)--(7.696,4.466)--(9.822,5.127)%
  --(11.947,5.462);
\gpcolor{color=gp lt color border}
\node[gp node right] at (10.479,1.627) {GCN-pagerank-LP};
\gpcolor{rgb color={0.902,0.624,0.000}}
\draw[gp path] (10.663,1.627)--(11.579,1.627);
\draw[gp path] (1.320,1.946)--(3.445,2.772)--(5.571,3.351)--(7.696,4.613)--(9.822,5.232)%
  --(11.947,5.192);
\gpcolor{color=gp lt color border}
\node[gp node right] at (10.479,1.319) {GCN-random-LP};
\gpcolor{rgb color={0.941,0.894,0.259}}
\draw[gp path] (10.663,1.319)--(11.579,1.319);
\draw[gp path] (1.320,1.946)--(3.445,2.536)--(5.571,3.194)--(7.696,4.144)--(9.822,4.772)%
  --(11.947,5.055);
\gpcolor{color=gp lt color border}
\draw[gp path] (1.320,8.441)--(1.320,0.985)--(11.947,0.985)--(11.947,8.441)--cycle;
%% coordinates of the plot area
\gpdefrectangularnode{gp plot 1}{\pgfpoint{1.320cm}{0.985cm}}{\pgfpoint{11.947cm}{8.441cm}}
\end{tikzpicture}
%% gnuplot variables

%% file: plots/stage1/cora_gpn-gcn.tex
\begin{tikzpicture}[gnuplot]
%% generated with GNUPLOT 5.4p3 (Lua 5.4; terminal rev. Jun 2020, script rev. 115)
%% Sun 21 Jan 2024 04:32:30 PM CET
\path (0.000,0.000) rectangle (12.500,8.750);
\gpcolor{color=gp lt color axes}
\gpsetlinetype{gp lt axes}
\gpsetdashtype{gp dt axes}
\gpsetlinewidth{0.50}
\draw[gp path] (1.320,0.985)--(11.947,0.985);
\gpcolor{color=gp lt color border}
\gpsetlinetype{gp lt border}
\gpsetdashtype{gp dt solid}
\gpsetlinewidth{1.00}
\draw[gp path] (1.320,0.985)--(1.500,0.985);
\draw[gp path] (11.947,0.985)--(11.767,0.985);
\node[gp node right] at (1.136,0.985) {$0$};
\gpcolor{color=gp lt color axes}
\gpsetlinetype{gp lt axes}
\gpsetdashtype{gp dt axes}
\gpsetlinewidth{0.50}
\draw[gp path] (1.320,2.476)--(7.535,2.476);
\draw[gp path] (11.763,2.476)--(11.947,2.476);
\gpcolor{color=gp lt color border}
\gpsetlinetype{gp lt border}
\gpsetdashtype{gp dt solid}
\gpsetlinewidth{1.00}
\draw[gp path] (1.320,2.476)--(1.500,2.476);
\draw[gp path] (11.947,2.476)--(11.767,2.476);
\node[gp node right] at (1.136,2.476) {$0.2$};
\gpcolor{color=gp lt color axes}
\gpsetlinetype{gp lt axes}
\gpsetdashtype{gp dt axes}
\gpsetlinewidth{0.50}
\draw[gp path] (1.320,3.967)--(11.947,3.967);
\gpcolor{color=gp lt color border}
\gpsetlinetype{gp lt border}
\gpsetdashtype{gp dt solid}
\gpsetlinewidth{1.00}
\draw[gp path] (1.320,3.967)--(1.500,3.967);
\draw[gp path] (11.947,3.967)--(11.767,3.967);
\node[gp node right] at (1.136,3.967) {$0.4$};
\gpcolor{color=gp lt color axes}
\gpsetlinetype{gp lt axes}
\gpsetdashtype{gp dt axes}
\gpsetlinewidth{0.50}
\draw[gp path] (1.320,5.459)--(11.947,5.459);
\gpcolor{color=gp lt color border}
\gpsetlinetype{gp lt border}
\gpsetdashtype{gp dt solid}
\gpsetlinewidth{1.00}
\draw[gp path] (1.320,5.459)--(1.500,5.459);
\draw[gp path] (11.947,5.459)--(11.767,5.459);
\node[gp node right] at (1.136,5.459) {$0.6$};
\gpcolor{color=gp lt color axes}
\gpsetlinetype{gp lt axes}
\gpsetdashtype{gp dt axes}
\gpsetlinewidth{0.50}
\draw[gp path] (1.320,6.950)--(11.947,6.950);
\gpcolor{color=gp lt color border}
\gpsetlinetype{gp lt border}
\gpsetdashtype{gp dt solid}
\gpsetlinewidth{1.00}
\draw[gp path] (1.320,6.950)--(1.500,6.950);
\draw[gp path] (11.947,6.950)--(11.767,6.950);
\node[gp node right] at (1.136,6.950) {$0.8$};
\gpcolor{color=gp lt color axes}
\gpsetlinetype{gp lt axes}
\gpsetdashtype{gp dt axes}
\gpsetlinewidth{0.50}
\draw[gp path] (1.320,8.441)--(11.947,8.441);
\gpcolor{color=gp lt color border}
\gpsetlinetype{gp lt border}
\gpsetdashtype{gp dt solid}
\gpsetlinewidth{1.00}
\draw[gp path] (1.320,8.441)--(1.500,8.441);
\draw[gp path] (11.947,8.441)--(11.767,8.441);
\node[gp node right] at (1.136,8.441) {$1$};
\gpcolor{color=gp lt color axes}
\gpsetlinetype{gp lt axes}
\gpsetdashtype{gp dt axes}
\gpsetlinewidth{0.50}
\draw[gp path] (1.320,0.985)--(1.320,8.441);
\gpcolor{color=gp lt color border}
\gpsetlinetype{gp lt border}
\gpsetdashtype{gp dt solid}
\gpsetlinewidth{1.00}
\draw[gp path] (1.320,0.985)--(1.320,1.165);
\draw[gp path] (1.320,8.441)--(1.320,8.261);
\node[gp node center] at (1.320,0.677) {$0$};
\gpcolor{color=gp lt color axes}
\gpsetlinetype{gp lt axes}
\gpsetdashtype{gp dt axes}
\gpsetlinewidth{0.50}
\draw[gp path] (5.571,0.985)--(5.571,8.441);
\gpcolor{color=gp lt color border}
\gpsetlinetype{gp lt border}
\gpsetdashtype{gp dt solid}
\gpsetlinewidth{1.00}
\draw[gp path] (5.571,0.985)--(5.571,1.165);
\draw[gp path] (5.571,8.441)--(5.571,8.261);
\node[gp node center] at (5.571,0.677) {$14$};
\gpcolor{color=gp lt color axes}
\gpsetlinetype{gp lt axes}
\gpsetdashtype{gp dt axes}
\gpsetlinewidth{0.50}
\draw[gp path] (9.822,0.985)--(9.822,1.165);
\draw[gp path] (9.822,2.705)--(9.822,8.441);
\gpcolor{color=gp lt color border}
\gpsetlinetype{gp lt border}
\gpsetdashtype{gp dt solid}
\gpsetlinewidth{1.00}
\draw[gp path] (9.822,0.985)--(9.822,1.165);
\draw[gp path] (9.822,8.441)--(9.822,8.261);
\node[gp node center] at (9.822,0.677) {$28$};
\draw[gp path] (1.320,8.441)--(1.320,0.985)--(11.947,0.985)--(11.947,8.441)--cycle;
\node[gp node center,rotate=-270] at (0.292,4.713) {Test accuracy};
\node[gp node center] at (6.633,0.215) {Budget used};
\node[gp node right] at (10.479,2.551) {GPN-GCN-entropy};
\gpcolor{rgb color={0.580,0.000,0.827}}
\draw[gp path] (10.663,2.551)--(11.579,2.551);
\draw[gp path] (1.320,2.050)--(3.445,3.490)--(5.571,3.516)--(7.696,4.697)--(9.822,5.457)%
  --(11.947,5.827);
\gpcolor{color=gp lt color border}
\node[gp node right] at (10.479,2.243) {GPN-GCN-medoids};
\gpcolor{rgb color={0.000,0.620,0.451}}
\draw[gp path] (10.663,2.243)--(11.579,2.243);
\draw[gp path] (1.320,2.050)--(3.445,3.390)--(5.571,4.108)--(7.696,4.963)--(9.822,5.686)%
  --(11.947,6.132);
\gpcolor{color=gp lt color border}
\node[gp node right] at (10.479,1.935) {GPN-GCN-own};
\gpcolor{rgb color={0.337,0.706,0.914}}
\draw[gp path] (10.663,1.935)--(11.579,1.935);
\draw[gp path] (1.320,2.050)--(3.445,3.263)--(5.571,3.948)--(7.696,5.210)--(9.822,5.618)%
  --(11.947,5.883);
\gpcolor{color=gp lt color border}
\node[gp node right] at (10.479,1.627) {GPN-GCN-pagerank};
\gpcolor{rgb color={0.902,0.624,0.000}}
\draw[gp path] (10.663,1.627)--(11.579,1.627);
\draw[gp path] (1.320,2.050)--(3.445,3.295)--(5.571,4.045)--(7.696,5.125)--(9.822,5.636)%
  --(11.947,5.930);
\gpcolor{color=gp lt color border}
\node[gp node right] at (10.479,1.319) {GPN-GCN-random};
\gpcolor{rgb color={0.941,0.894,0.259}}
\draw[gp path] (10.663,1.319)--(11.579,1.319);
\draw[gp path] (1.320,2.050)--(3.445,3.490)--(5.571,3.531)--(7.696,4.772)--(9.822,5.469)%
  --(11.947,5.874);
\gpcolor{color=gp lt color border}
\draw[gp path] (1.320,8.441)--(1.320,0.985)--(11.947,0.985)--(11.947,8.441)--cycle;
%% coordinates of the plot area
\gpdefrectangularnode{gp plot 1}{\pgfpoint{1.320cm}{0.985cm}}{\pgfpoint{11.947cm}{8.441cm}}
\end{tikzpicture}
%% gnuplot variables

%% file: plots/stage1/cora_gpn-gcn_lp.tex
\begin{tikzpicture}[gnuplot]
%% generated with GNUPLOT 5.4p3 (Lua 5.4; terminal rev. Jun 2020, script rev. 115)
%% Sun 21 Jan 2024 04:32:12 PM CET
\path (0.000,0.000) rectangle (12.500,8.750);
\gpcolor{color=gp lt color axes}
\gpsetlinetype{gp lt axes}
\gpsetdashtype{gp dt axes}
\gpsetlinewidth{0.50}
\draw[gp path] (1.320,0.985)--(11.947,0.985);
\gpcolor{color=gp lt color border}
\gpsetlinetype{gp lt border}
\gpsetdashtype{gp dt solid}
\gpsetlinewidth{1.00}
\draw[gp path] (1.320,0.985)--(1.500,0.985);
\draw[gp path] (11.947,0.985)--(11.767,0.985);
\node[gp node right] at (1.136,0.985) {$0$};
\gpcolor{color=gp lt color axes}
\gpsetlinetype{gp lt axes}
\gpsetdashtype{gp dt axes}
\gpsetlinewidth{0.50}
\draw[gp path] (1.320,2.476)--(6.983,2.476);
\draw[gp path] (11.763,2.476)--(11.947,2.476);
\gpcolor{color=gp lt color border}
\gpsetlinetype{gp lt border}
\gpsetdashtype{gp dt solid}
\gpsetlinewidth{1.00}
\draw[gp path] (1.320,2.476)--(1.500,2.476);
\draw[gp path] (11.947,2.476)--(11.767,2.476);
\node[gp node right] at (1.136,2.476) {$0.2$};
\gpcolor{color=gp lt color axes}
\gpsetlinetype{gp lt axes}
\gpsetdashtype{gp dt axes}
\gpsetlinewidth{0.50}
\draw[gp path] (1.320,3.967)--(11.947,3.967);
\gpcolor{color=gp lt color border}
\gpsetlinetype{gp lt border}
\gpsetdashtype{gp dt solid}
\gpsetlinewidth{1.00}
\draw[gp path] (1.320,3.967)--(1.500,3.967);
\draw[gp path] (11.947,3.967)--(11.767,3.967);
\node[gp node right] at (1.136,3.967) {$0.4$};
\gpcolor{color=gp lt color axes}
\gpsetlinetype{gp lt axes}
\gpsetdashtype{gp dt axes}
\gpsetlinewidth{0.50}
\draw[gp path] (1.320,5.459)--(11.947,5.459);
\gpcolor{color=gp lt color border}
\gpsetlinetype{gp lt border}
\gpsetdashtype{gp dt solid}
\gpsetlinewidth{1.00}
\draw[gp path] (1.320,5.459)--(1.500,5.459);
\draw[gp path] (11.947,5.459)--(11.767,5.459);
\node[gp node right] at (1.136,5.459) {$0.6$};
\gpcolor{color=gp lt color axes}
\gpsetlinetype{gp lt axes}
\gpsetdashtype{gp dt axes}
\gpsetlinewidth{0.50}
\draw[gp path] (1.320,6.950)--(11.947,6.950);
\gpcolor{color=gp lt color border}
\gpsetlinetype{gp lt border}
\gpsetdashtype{gp dt solid}
\gpsetlinewidth{1.00}
\draw[gp path] (1.320,6.950)--(1.500,6.950);
\draw[gp path] (11.947,6.950)--(11.767,6.950);
\node[gp node right] at (1.136,6.950) {$0.8$};
\gpcolor{color=gp lt color axes}
\gpsetlinetype{gp lt axes}
\gpsetdashtype{gp dt axes}
\gpsetlinewidth{0.50}
\draw[gp path] (1.320,8.441)--(11.947,8.441);
\gpcolor{color=gp lt color border}
\gpsetlinetype{gp lt border}
\gpsetdashtype{gp dt solid}
\gpsetlinewidth{1.00}
\draw[gp path] (1.320,8.441)--(1.500,8.441);
\draw[gp path] (11.947,8.441)--(11.767,8.441);
\node[gp node right] at (1.136,8.441) {$1$};
\gpcolor{color=gp lt color axes}
\gpsetlinetype{gp lt axes}
\gpsetdashtype{gp dt axes}
\gpsetlinewidth{0.50}
\draw[gp path] (1.320,0.985)--(1.320,8.441);
\gpcolor{color=gp lt color border}
\gpsetlinetype{gp lt border}
\gpsetdashtype{gp dt solid}
\gpsetlinewidth{1.00}
\draw[gp path] (1.320,0.985)--(1.320,1.165);
\draw[gp path] (1.320,8.441)--(1.320,8.261);
\node[gp node center] at (1.320,0.677) {$0$};
\gpcolor{color=gp lt color axes}
\gpsetlinetype{gp lt axes}
\gpsetdashtype{gp dt axes}
\gpsetlinewidth{0.50}
\draw[gp path] (5.571,0.985)--(5.571,8.441);
\gpcolor{color=gp lt color border}
\gpsetlinetype{gp lt border}
\gpsetdashtype{gp dt solid}
\gpsetlinewidth{1.00}
\draw[gp path] (5.571,0.985)--(5.571,1.165);
\draw[gp path] (5.571,8.441)--(5.571,8.261);
\node[gp node center] at (5.571,0.677) {$14$};
\gpcolor{color=gp lt color axes}
\gpsetlinetype{gp lt axes}
\gpsetdashtype{gp dt axes}
\gpsetlinewidth{0.50}
\draw[gp path] (9.822,0.985)--(9.822,1.165);
\draw[gp path] (9.822,2.705)--(9.822,8.441);
\gpcolor{color=gp lt color border}
\gpsetlinetype{gp lt border}
\gpsetdashtype{gp dt solid}
\gpsetlinewidth{1.00}
\draw[gp path] (9.822,0.985)--(9.822,1.165);
\draw[gp path] (9.822,8.441)--(9.822,8.261);
\node[gp node center] at (9.822,0.677) {$28$};
\draw[gp path] (1.320,8.441)--(1.320,0.985)--(11.947,0.985)--(11.947,8.441)--cycle;
\node[gp node center,rotate=-270] at (0.292,4.713) {Test accuracy};
\node[gp node center] at (6.633,0.215) {Budget used};
\node[gp node right] at (10.479,2.551) {GPN-GCN-entropy-LP};
\gpcolor{rgb color={0.580,0.000,0.827}}
\draw[gp path] (10.663,2.551)--(11.579,2.551);
\draw[gp path] (1.320,2.050)--(3.445,4.336)--(5.571,5.135)--(7.696,5.501)--(9.822,5.608)%
  --(11.947,5.697);
\gpcolor{color=gp lt color border}
\node[gp node right] at (10.479,2.243) {GPN-GCN-medoids-LP};
\gpcolor{rgb color={0.000,0.620,0.451}}
\draw[gp path] (10.663,2.243)--(11.579,2.243);
\draw[gp path] (1.320,2.050)--(3.445,3.854)--(5.571,4.797)--(7.696,5.547)--(9.822,5.725)%
  --(11.947,5.876);
\gpcolor{color=gp lt color border}
\node[gp node right] at (10.479,1.935) {GPN-GCN-own-LP};
\gpcolor{rgb color={0.337,0.706,0.914}}
\draw[gp path] (10.663,1.935)--(11.579,1.935);
\draw[gp path] (1.320,2.050)--(3.445,4.380)--(5.571,4.718)--(7.696,5.312)--(9.822,5.502)%
  --(11.947,5.778);
\gpcolor{color=gp lt color border}
\node[gp node right] at (10.479,1.627) {GPN-GCN-pagerank-LP};
\gpcolor{rgb color={0.902,0.624,0.000}}
\draw[gp path] (10.663,1.627)--(11.579,1.627);
\draw[gp path] (1.320,2.050)--(3.445,4.047)--(5.571,4.929)--(7.696,5.296)--(9.822,5.505)%
  --(11.947,5.610);
\gpcolor{color=gp lt color border}
\node[gp node right] at (10.479,1.319) {GPN-GCN-random-LP};
\gpcolor{rgb color={0.941,0.894,0.259}}
\draw[gp path] (10.663,1.319)--(11.579,1.319);
\draw[gp path] (1.320,2.050)--(3.445,4.336)--(5.571,5.233)--(7.696,5.437)--(9.822,5.581)%
  --(11.947,5.784);
\gpcolor{color=gp lt color border}
\draw[gp path] (1.320,8.441)--(1.320,0.985)--(11.947,0.985)--(11.947,8.441)--cycle;
%% coordinates of the plot area
\gpdefrectangularnode{gp plot 1}{\pgfpoint{1.320cm}{0.985cm}}{\pgfpoint{11.947cm}{8.441cm}}
\end{tikzpicture}
%% gnuplot variables

%% file: plots/stage2/cora_lp.tex
\begin{tikzpicture}[gnuplot]
%% generated with GNUPLOT 5.4p3 (Lua 5.4; terminal rev. Jun 2020, script rev. 115)
%% Sun 21 Jan 2024 04:44:06 PM CET
\path (0.000,0.000) rectangle (12.500,8.750);
\gpcolor{color=gp lt color axes}
\gpsetlinetype{gp lt axes}
\gpsetdashtype{gp dt axes}
\gpsetlinewidth{0.50}
\draw[gp path] (1.320,0.985)--(11.947,0.985);
\gpcolor{color=gp lt color border}
\gpsetlinetype{gp lt border}
\gpsetdashtype{gp dt solid}
\gpsetlinewidth{1.00}
\draw[gp path] (1.320,0.985)--(1.500,0.985);
\draw[gp path] (11.947,0.985)--(11.767,0.985);
\node[gp node right] at (1.136,0.985) {$0$};
\gpcolor{color=gp lt color axes}
\gpsetlinetype{gp lt axes}
\gpsetdashtype{gp dt axes}
\gpsetlinewidth{0.50}
\draw[gp path] (1.320,2.476)--(8.455,2.476);
\draw[gp path] (11.763,2.476)--(11.947,2.476);
\gpcolor{color=gp lt color border}
\gpsetlinetype{gp lt border}
\gpsetdashtype{gp dt solid}
\gpsetlinewidth{1.00}
\draw[gp path] (1.320,2.476)--(1.500,2.476);
\draw[gp path] (11.947,2.476)--(11.767,2.476);
\node[gp node right] at (1.136,2.476) {$0.2$};
\gpcolor{color=gp lt color axes}
\gpsetlinetype{gp lt axes}
\gpsetdashtype{gp dt axes}
\gpsetlinewidth{0.50}
\draw[gp path] (1.320,3.967)--(11.947,3.967);
\gpcolor{color=gp lt color border}
\gpsetlinetype{gp lt border}
\gpsetdashtype{gp dt solid}
\gpsetlinewidth{1.00}
\draw[gp path] (1.320,3.967)--(1.500,3.967);
\draw[gp path] (11.947,3.967)--(11.767,3.967);
\node[gp node right] at (1.136,3.967) {$0.4$};
\gpcolor{color=gp lt color axes}
\gpsetlinetype{gp lt axes}
\gpsetdashtype{gp dt axes}
\gpsetlinewidth{0.50}
\draw[gp path] (1.320,5.459)--(11.947,5.459);
\gpcolor{color=gp lt color border}
\gpsetlinetype{gp lt border}
\gpsetdashtype{gp dt solid}
\gpsetlinewidth{1.00}
\draw[gp path] (1.320,5.459)--(1.500,5.459);
\draw[gp path] (11.947,5.459)--(11.767,5.459);
\node[gp node right] at (1.136,5.459) {$0.6$};
\gpcolor{color=gp lt color axes}
\gpsetlinetype{gp lt axes}
\gpsetdashtype{gp dt axes}
\gpsetlinewidth{0.50}
\draw[gp path] (1.320,6.950)--(11.947,6.950);
\gpcolor{color=gp lt color border}
\gpsetlinetype{gp lt border}
\gpsetdashtype{gp dt solid}
\gpsetlinewidth{1.00}
\draw[gp path] (1.320,6.950)--(1.500,6.950);
\draw[gp path] (11.947,6.950)--(11.767,6.950);
\node[gp node right] at (1.136,6.950) {$0.8$};
\gpcolor{color=gp lt color axes}
\gpsetlinetype{gp lt axes}
\gpsetdashtype{gp dt axes}
\gpsetlinewidth{0.50}
\draw[gp path] (1.320,8.441)--(11.947,8.441);
\gpcolor{color=gp lt color border}
\gpsetlinetype{gp lt border}
\gpsetdashtype{gp dt solid}
\gpsetlinewidth{1.00}
\draw[gp path] (1.320,8.441)--(1.500,8.441);
\draw[gp path] (11.947,8.441)--(11.767,8.441);
\node[gp node right] at (1.136,8.441) {$1$};
\gpcolor{color=gp lt color axes}
\gpsetlinetype{gp lt axes}
\gpsetdashtype{gp dt axes}
\gpsetlinewidth{0.50}
\draw[gp path] (1.320,0.985)--(1.320,8.441);
\gpcolor{color=gp lt color border}
\gpsetlinetype{gp lt border}
\gpsetdashtype{gp dt solid}
\gpsetlinewidth{1.00}
\draw[gp path] (1.320,0.985)--(1.320,1.165);
\draw[gp path] (1.320,8.441)--(1.320,8.261);
\node[gp node center] at (1.320,0.677) {$0$};
\gpcolor{color=gp lt color axes}
\gpsetlinetype{gp lt axes}
\gpsetdashtype{gp dt axes}
\gpsetlinewidth{0.50}
\draw[gp path] (5.571,0.985)--(5.571,8.441);
\gpcolor{color=gp lt color border}
\gpsetlinetype{gp lt border}
\gpsetdashtype{gp dt solid}
\gpsetlinewidth{1.00}
\draw[gp path] (5.571,0.985)--(5.571,1.165);
\draw[gp path] (5.571,8.441)--(5.571,8.261);
\node[gp node center] at (5.571,0.677) {$14$};
\gpcolor{color=gp lt color axes}
\gpsetlinetype{gp lt axes}
\gpsetdashtype{gp dt axes}
\gpsetlinewidth{0.50}
\draw[gp path] (9.822,0.985)--(9.822,1.165);
\draw[gp path] (9.822,2.705)--(9.822,8.441);
\gpcolor{color=gp lt color border}
\gpsetlinetype{gp lt border}
\gpsetdashtype{gp dt solid}
\gpsetlinewidth{1.00}
\draw[gp path] (9.822,0.985)--(9.822,1.165);
\draw[gp path] (9.822,8.441)--(9.822,8.261);
\node[gp node center] at (9.822,0.677) {$28$};
\draw[gp path] (1.320,8.441)--(1.320,0.985)--(11.947,0.985)--(11.947,8.441)--cycle;
\node[gp node center,rotate=-270] at (0.292,4.713) {Test accuracy};
\node[gp node center] at (6.633,0.215) {Budget used};
\node[gp node right] at (10.479,2.551) {LP-entropy};
\gpcolor{rgb color={0.580,0.000,0.827}}
\draw[gp path] (10.663,2.551)--(11.579,2.551);
\draw[gp path] (1.320,2.040)--(3.445,3.363)--(5.571,4.035)--(7.696,4.393)--(9.822,4.639)%
  --(11.947,4.807);
\gpcolor{color=gp lt color border}
\node[gp node right] at (10.479,2.243) {LP-medoids};
\gpcolor{rgb color={0.000,0.620,0.451}}
\draw[gp path] (10.663,2.243)--(11.579,2.243);
\draw[gp path] (1.320,2.040)--(3.445,3.444)--(5.571,4.012)--(7.696,4.644)--(9.822,4.803)%
  --(11.947,4.998);
\gpcolor{color=gp lt color border}
\node[gp node right] at (10.479,1.935) {LP-own};
\gpcolor{rgb color={0.337,0.706,0.914}}
\draw[gp path] (10.663,1.935)--(11.579,1.935);
\draw[gp path] (1.320,2.040)--(3.445,3.586)--(5.571,4.205)--(7.696,4.559)--(9.822,4.857)%
  --(11.947,4.969);
\gpcolor{color=gp lt color border}
\node[gp node right] at (10.479,1.627) {LP-pagerank};
\gpcolor{rgb color={0.902,0.624,0.000}}
\draw[gp path] (10.663,1.627)--(11.579,1.627);
\draw[gp path] (1.320,2.040)--(3.445,3.758)--(5.571,4.449)--(7.696,4.704)--(9.822,4.834)%
  --(11.947,4.997);
\gpcolor{color=gp lt color border}
\node[gp node right] at (10.479,1.319) {LP-random};
\gpcolor{rgb color={0.941,0.894,0.259}}
\draw[gp path] (10.663,1.319)--(11.579,1.319);
\draw[gp path] (1.320,2.040)--(3.445,3.363)--(5.571,4.035)--(7.696,4.393)--(9.822,4.639)%
  --(11.947,4.807);
\gpcolor{color=gp lt color border}
\draw[gp path] (1.320,8.441)--(1.320,0.985)--(11.947,0.985)--(11.947,8.441)--cycle;
%% coordinates of the plot area
\gpdefrectangularnode{gp plot 1}{\pgfpoint{1.320cm}{0.985cm}}{\pgfpoint{11.947cm}{8.441cm}}
\end{tikzpicture}
%% gnuplot variables

%% file: plots/stage2/cora_lp_lp.tex
\begin{tikzpicture}[gnuplot]
%% generated with GNUPLOT 5.4p3 (Lua 5.4; terminal rev. Jun 2020, script rev. 115)
%% Sun 21 Jan 2024 04:43:45 PM CET
\path (0.000,0.000) rectangle (12.500,8.750);
\gpcolor{color=gp lt color axes}
\gpsetlinetype{gp lt axes}
\gpsetdashtype{gp dt axes}
\gpsetlinewidth{0.50}
\draw[gp path] (1.320,0.985)--(11.947,0.985);
\gpcolor{color=gp lt color border}
\gpsetlinetype{gp lt border}
\gpsetdashtype{gp dt solid}
\gpsetlinewidth{1.00}
\draw[gp path] (1.320,0.985)--(1.500,0.985);
\draw[gp path] (11.947,0.985)--(11.767,0.985);
\node[gp node right] at (1.136,0.985) {$0$};
\gpcolor{color=gp lt color axes}
\gpsetlinetype{gp lt axes}
\gpsetdashtype{gp dt axes}
\gpsetlinewidth{0.50}
\draw[gp path] (1.320,2.476)--(7.903,2.476);
\draw[gp path] (11.763,2.476)--(11.947,2.476);
\gpcolor{color=gp lt color border}
\gpsetlinetype{gp lt border}
\gpsetdashtype{gp dt solid}
\gpsetlinewidth{1.00}
\draw[gp path] (1.320,2.476)--(1.500,2.476);
\draw[gp path] (11.947,2.476)--(11.767,2.476);
\node[gp node right] at (1.136,2.476) {$0.2$};
\gpcolor{color=gp lt color axes}
\gpsetlinetype{gp lt axes}
\gpsetdashtype{gp dt axes}
\gpsetlinewidth{0.50}
\draw[gp path] (1.320,3.967)--(11.947,3.967);
\gpcolor{color=gp lt color border}
\gpsetlinetype{gp lt border}
\gpsetdashtype{gp dt solid}
\gpsetlinewidth{1.00}
\draw[gp path] (1.320,3.967)--(1.500,3.967);
\draw[gp path] (11.947,3.967)--(11.767,3.967);
\node[gp node right] at (1.136,3.967) {$0.4$};
\gpcolor{color=gp lt color axes}
\gpsetlinetype{gp lt axes}
\gpsetdashtype{gp dt axes}
\gpsetlinewidth{0.50}
\draw[gp path] (1.320,5.459)--(11.947,5.459);
\gpcolor{color=gp lt color border}
\gpsetlinetype{gp lt border}
\gpsetdashtype{gp dt solid}
\gpsetlinewidth{1.00}
\draw[gp path] (1.320,5.459)--(1.500,5.459);
\draw[gp path] (11.947,5.459)--(11.767,5.459);
\node[gp node right] at (1.136,5.459) {$0.6$};
\gpcolor{color=gp lt color axes}
\gpsetlinetype{gp lt axes}
\gpsetdashtype{gp dt axes}
\gpsetlinewidth{0.50}
\draw[gp path] (1.320,6.950)--(11.947,6.950);
\gpcolor{color=gp lt color border}
\gpsetlinetype{gp lt border}
\gpsetdashtype{gp dt solid}
\gpsetlinewidth{1.00}
\draw[gp path] (1.320,6.950)--(1.500,6.950);
\draw[gp path] (11.947,6.950)--(11.767,6.950);
\node[gp node right] at (1.136,6.950) {$0.8$};
\gpcolor{color=gp lt color axes}
\gpsetlinetype{gp lt axes}
\gpsetdashtype{gp dt axes}
\gpsetlinewidth{0.50}
\draw[gp path] (1.320,8.441)--(11.947,8.441);
\gpcolor{color=gp lt color border}
\gpsetlinetype{gp lt border}
\gpsetdashtype{gp dt solid}
\gpsetlinewidth{1.00}
\draw[gp path] (1.320,8.441)--(1.500,8.441);
\draw[gp path] (11.947,8.441)--(11.767,8.441);
\node[gp node right] at (1.136,8.441) {$1$};
\gpcolor{color=gp lt color axes}
\gpsetlinetype{gp lt axes}
\gpsetdashtype{gp dt axes}
\gpsetlinewidth{0.50}
\draw[gp path] (1.320,0.985)--(1.320,8.441);
\gpcolor{color=gp lt color border}
\gpsetlinetype{gp lt border}
\gpsetdashtype{gp dt solid}
\gpsetlinewidth{1.00}
\draw[gp path] (1.320,0.985)--(1.320,1.165);
\draw[gp path] (1.320,8.441)--(1.320,8.261);
\node[gp node center] at (1.320,0.677) {$0$};
\gpcolor{color=gp lt color axes}
\gpsetlinetype{gp lt axes}
\gpsetdashtype{gp dt axes}
\gpsetlinewidth{0.50}
\draw[gp path] (5.571,0.985)--(5.571,8.441);
\gpcolor{color=gp lt color border}
\gpsetlinetype{gp lt border}
\gpsetdashtype{gp dt solid}
\gpsetlinewidth{1.00}
\draw[gp path] (5.571,0.985)--(5.571,1.165);
\draw[gp path] (5.571,8.441)--(5.571,8.261);
\node[gp node center] at (5.571,0.677) {$14$};
\gpcolor{color=gp lt color axes}
\gpsetlinetype{gp lt axes}
\gpsetdashtype{gp dt axes}
\gpsetlinewidth{0.50}
\draw[gp path] (9.822,0.985)--(9.822,1.165);
\draw[gp path] (9.822,2.705)--(9.822,8.441);
\gpcolor{color=gp lt color border}
\gpsetlinetype{gp lt border}
\gpsetdashtype{gp dt solid}
\gpsetlinewidth{1.00}
\draw[gp path] (9.822,0.985)--(9.822,1.165);
\draw[gp path] (9.822,8.441)--(9.822,8.261);
\node[gp node center] at (9.822,0.677) {$28$};
\draw[gp path] (1.320,8.441)--(1.320,0.985)--(11.947,0.985)--(11.947,8.441)--cycle;
\node[gp node center,rotate=-270] at (0.292,4.713) {Test accuracy};
\node[gp node center] at (6.633,0.215) {Budget used};
\node[gp node right] at (10.479,2.551) {LP-entropy-LP};
\gpcolor{rgb color={0.580,0.000,0.827}}
\draw[gp path] (10.663,2.551)--(11.579,2.551);
\draw[gp path] (1.320,2.040)--(3.445,3.363)--(5.571,4.035)--(7.696,4.393)--(9.822,4.639)%
  --(11.947,4.807);
\gpcolor{color=gp lt color border}
\node[gp node right] at (10.479,2.243) {LP-medoids-LP};
\gpcolor{rgb color={0.000,0.620,0.451}}
\draw[gp path] (10.663,2.243)--(11.579,2.243);
\draw[gp path] (1.320,2.040)--(3.445,3.444)--(5.571,4.012)--(7.696,4.644)--(9.822,4.803)%
  --(11.947,4.998);
\gpcolor{color=gp lt color border}
\node[gp node right] at (10.479,1.935) {LP-own-LP};
\gpcolor{rgb color={0.337,0.706,0.914}}
\draw[gp path] (10.663,1.935)--(11.579,1.935);
\draw[gp path] (1.320,2.040)--(3.445,3.586)--(5.571,4.205)--(7.696,4.559)--(9.822,4.857)%
  --(11.947,4.969);
\gpcolor{color=gp lt color border}
\node[gp node right] at (10.479,1.627) {LP-pagerank-LP};
\gpcolor{rgb color={0.902,0.624,0.000}}
\draw[gp path] (10.663,1.627)--(11.579,1.627);
\draw[gp path] (1.320,2.040)--(3.445,3.758)--(5.571,4.449)--(7.696,4.704)--(9.822,4.834)%
  --(11.947,4.997);
\gpcolor{color=gp lt color border}
\node[gp node right] at (10.479,1.319) {LP-random-LP};
\gpcolor{rgb color={0.941,0.894,0.259}}
\draw[gp path] (10.663,1.319)--(11.579,1.319);
\draw[gp path] (1.320,2.040)--(3.445,3.363)--(5.571,4.035)--(7.696,4.393)--(9.822,4.639)%
  --(11.947,4.807);
\gpcolor{color=gp lt color border}
\draw[gp path] (1.320,8.441)--(1.320,0.985)--(11.947,0.985)--(11.947,8.441)--cycle;
%% coordinates of the plot area
\gpdefrectangularnode{gp plot 1}{\pgfpoint{1.320cm}{0.985cm}}{\pgfpoint{11.947cm}{8.441cm}}
\end{tikzpicture}
%% gnuplot variables

%% file: plots/stage2/cora_gcn.tex
\begin{tikzpicture}[gnuplot]
%% generated with GNUPLOT 5.4p3 (Lua 5.4; terminal rev. Jun 2020, script rev. 115)
%% Sun 21 Jan 2024 04:44:27 PM CET
\path (0.000,0.000) rectangle (12.500,8.750);
\gpcolor{color=gp lt color axes}
\gpsetlinetype{gp lt axes}
\gpsetdashtype{gp dt axes}
\gpsetlinewidth{0.50}
\draw[gp path] (1.320,0.985)--(11.947,0.985);
\gpcolor{color=gp lt color border}
\gpsetlinetype{gp lt border}
\gpsetdashtype{gp dt solid}
\gpsetlinewidth{1.00}
\draw[gp path] (1.320,0.985)--(1.500,0.985);
\draw[gp path] (11.947,0.985)--(11.767,0.985);
\node[gp node right] at (1.136,0.985) {$0$};
\gpcolor{color=gp lt color axes}
\gpsetlinetype{gp lt axes}
\gpsetdashtype{gp dt axes}
\gpsetlinewidth{0.50}
\draw[gp path] (1.320,2.476)--(8.271,2.476);
\draw[gp path] (11.763,2.476)--(11.947,2.476);
\gpcolor{color=gp lt color border}
\gpsetlinetype{gp lt border}
\gpsetdashtype{gp dt solid}
\gpsetlinewidth{1.00}
\draw[gp path] (1.320,2.476)--(1.500,2.476);
\draw[gp path] (11.947,2.476)--(11.767,2.476);
\node[gp node right] at (1.136,2.476) {$0.2$};
\gpcolor{color=gp lt color axes}
\gpsetlinetype{gp lt axes}
\gpsetdashtype{gp dt axes}
\gpsetlinewidth{0.50}
\draw[gp path] (1.320,3.967)--(11.947,3.967);
\gpcolor{color=gp lt color border}
\gpsetlinetype{gp lt border}
\gpsetdashtype{gp dt solid}
\gpsetlinewidth{1.00}
\draw[gp path] (1.320,3.967)--(1.500,3.967);
\draw[gp path] (11.947,3.967)--(11.767,3.967);
\node[gp node right] at (1.136,3.967) {$0.4$};
\gpcolor{color=gp lt color axes}
\gpsetlinetype{gp lt axes}
\gpsetdashtype{gp dt axes}
\gpsetlinewidth{0.50}
\draw[gp path] (1.320,5.459)--(11.947,5.459);
\gpcolor{color=gp lt color border}
\gpsetlinetype{gp lt border}
\gpsetdashtype{gp dt solid}
\gpsetlinewidth{1.00}
\draw[gp path] (1.320,5.459)--(1.500,5.459);
\draw[gp path] (11.947,5.459)--(11.767,5.459);
\node[gp node right] at (1.136,5.459) {$0.6$};
\gpcolor{color=gp lt color axes}
\gpsetlinetype{gp lt axes}
\gpsetdashtype{gp dt axes}
\gpsetlinewidth{0.50}
\draw[gp path] (1.320,6.950)--(11.947,6.950);
\gpcolor{color=gp lt color border}
\gpsetlinetype{gp lt border}
\gpsetdashtype{gp dt solid}
\gpsetlinewidth{1.00}
\draw[gp path] (1.320,6.950)--(1.500,6.950);
\draw[gp path] (11.947,6.950)--(11.767,6.950);
\node[gp node right] at (1.136,6.950) {$0.8$};
\gpcolor{color=gp lt color axes}
\gpsetlinetype{gp lt axes}
\gpsetdashtype{gp dt axes}
\gpsetlinewidth{0.50}
\draw[gp path] (1.320,8.441)--(11.947,8.441);
\gpcolor{color=gp lt color border}
\gpsetlinetype{gp lt border}
\gpsetdashtype{gp dt solid}
\gpsetlinewidth{1.00}
\draw[gp path] (1.320,8.441)--(1.500,8.441);
\draw[gp path] (11.947,8.441)--(11.767,8.441);
\node[gp node right] at (1.136,8.441) {$1$};
\gpcolor{color=gp lt color axes}
\gpsetlinetype{gp lt axes}
\gpsetdashtype{gp dt axes}
\gpsetlinewidth{0.50}
\draw[gp path] (1.320,0.985)--(1.320,8.441);
\gpcolor{color=gp lt color border}
\gpsetlinetype{gp lt border}
\gpsetdashtype{gp dt solid}
\gpsetlinewidth{1.00}
\draw[gp path] (1.320,0.985)--(1.320,1.165);
\draw[gp path] (1.320,8.441)--(1.320,8.261);
\node[gp node center] at (1.320,0.677) {$0$};
\gpcolor{color=gp lt color axes}
\gpsetlinetype{gp lt axes}
\gpsetdashtype{gp dt axes}
\gpsetlinewidth{0.50}
\draw[gp path] (5.571,0.985)--(5.571,8.441);
\gpcolor{color=gp lt color border}
\gpsetlinetype{gp lt border}
\gpsetdashtype{gp dt solid}
\gpsetlinewidth{1.00}
\draw[gp path] (5.571,0.985)--(5.571,1.165);
\draw[gp path] (5.571,8.441)--(5.571,8.261);
\node[gp node center] at (5.571,0.677) {$14$};
\gpcolor{color=gp lt color axes}
\gpsetlinetype{gp lt axes}
\gpsetdashtype{gp dt axes}
\gpsetlinewidth{0.50}
\draw[gp path] (9.822,0.985)--(9.822,1.165);
\draw[gp path] (9.822,2.705)--(9.822,8.441);
\gpcolor{color=gp lt color border}
\gpsetlinetype{gp lt border}
\gpsetdashtype{gp dt solid}
\gpsetlinewidth{1.00}
\draw[gp path] (9.822,0.985)--(9.822,1.165);
\draw[gp path] (9.822,8.441)--(9.822,8.261);
\node[gp node center] at (9.822,0.677) {$28$};
\draw[gp path] (1.320,8.441)--(1.320,0.985)--(11.947,0.985)--(11.947,8.441)--cycle;
\node[gp node center,rotate=-270] at (0.292,4.713) {Test accuracy};
\node[gp node center] at (6.633,0.215) {Budget used};
\node[gp node right] at (10.479,2.551) {GCN-entropy};
\gpcolor{rgb color={0.580,0.000,0.827}}
\draw[gp path] (10.663,2.551)--(11.579,2.551);
\draw[gp path] (1.320,1.946)--(3.445,2.292)--(5.571,3.045)--(7.696,3.741)--(9.822,4.726)%
  --(11.947,5.037);
\gpcolor{color=gp lt color border}
\node[gp node right] at (10.479,2.243) {GCN-medoids};
\gpcolor{rgb color={0.000,0.620,0.451}}
\draw[gp path] (10.663,2.243)--(11.579,2.243);
\draw[gp path] (1.320,1.946)--(3.445,2.875)--(5.571,3.247)--(7.696,3.742)--(9.822,4.437)%
  --(11.947,4.927);
\gpcolor{color=gp lt color border}
\node[gp node right] at (10.479,1.935) {GCN-own};
\gpcolor{rgb color={0.337,0.706,0.914}}
\draw[gp path] (10.663,1.935)--(11.579,1.935);
\draw[gp path] (1.320,1.946)--(3.445,2.666)--(5.571,3.590)--(7.696,4.315)--(9.822,4.682)%
  --(11.947,4.996);
\gpcolor{color=gp lt color border}
\node[gp node right] at (10.479,1.627) {GCN-pagerank};
\gpcolor{rgb color={0.902,0.624,0.000}}
\draw[gp path] (10.663,1.627)--(11.579,1.627);
\draw[gp path] (1.320,1.946)--(3.445,2.608)--(5.571,3.226)--(7.696,3.999)--(9.822,4.505)%
  --(11.947,4.902);
\gpcolor{color=gp lt color border}
\node[gp node right] at (10.479,1.319) {GCN-random};
\gpcolor{rgb color={0.941,0.894,0.259}}
\draw[gp path] (10.663,1.319)--(11.579,1.319);
\draw[gp path] (1.320,1.946)--(3.445,2.292)--(5.571,3.006)--(7.696,3.739)--(9.822,4.483)%
  --(11.947,5.085);
\gpcolor{color=gp lt color border}
\draw[gp path] (1.320,8.441)--(1.320,0.985)--(11.947,0.985)--(11.947,8.441)--cycle;
%% coordinates of the plot area
\gpdefrectangularnode{gp plot 1}{\pgfpoint{1.320cm}{0.985cm}}{\pgfpoint{11.947cm}{8.441cm}}
\end{tikzpicture}
%% gnuplot variables

%% file: plots/stage2/cora_gcn_lp.tex
\begin{tikzpicture}[gnuplot]
%% generated with GNUPLOT 5.4p3 (Lua 5.4; terminal rev. Jun 2020, script rev. 115)
%% Sun 21 Jan 2024 04:44:46 PM CET
\path (0.000,0.000) rectangle (12.500,8.750);
\gpcolor{color=gp lt color axes}
\gpsetlinetype{gp lt axes}
\gpsetdashtype{gp dt axes}
\gpsetlinewidth{0.50}
\draw[gp path] (1.320,0.985)--(11.947,0.985);
\gpcolor{color=gp lt color border}
\gpsetlinetype{gp lt border}
\gpsetdashtype{gp dt solid}
\gpsetlinewidth{1.00}
\draw[gp path] (1.320,0.985)--(1.500,0.985);
\draw[gp path] (11.947,0.985)--(11.767,0.985);
\node[gp node right] at (1.136,0.985) {$0$};
\gpcolor{color=gp lt color axes}
\gpsetlinetype{gp lt axes}
\gpsetdashtype{gp dt axes}
\gpsetlinewidth{0.50}
\draw[gp path] (1.320,2.476)--(7.719,2.476);
\draw[gp path] (11.763,2.476)--(11.947,2.476);
\gpcolor{color=gp lt color border}
\gpsetlinetype{gp lt border}
\gpsetdashtype{gp dt solid}
\gpsetlinewidth{1.00}
\draw[gp path] (1.320,2.476)--(1.500,2.476);
\draw[gp path] (11.947,2.476)--(11.767,2.476);
\node[gp node right] at (1.136,2.476) {$0.2$};
\gpcolor{color=gp lt color axes}
\gpsetlinetype{gp lt axes}
\gpsetdashtype{gp dt axes}
\gpsetlinewidth{0.50}
\draw[gp path] (1.320,3.967)--(11.947,3.967);
\gpcolor{color=gp lt color border}
\gpsetlinetype{gp lt border}
\gpsetdashtype{gp dt solid}
\gpsetlinewidth{1.00}
\draw[gp path] (1.320,3.967)--(1.500,3.967);
\draw[gp path] (11.947,3.967)--(11.767,3.967);
\node[gp node right] at (1.136,3.967) {$0.4$};
\gpcolor{color=gp lt color axes}
\gpsetlinetype{gp lt axes}
\gpsetdashtype{gp dt axes}
\gpsetlinewidth{0.50}
\draw[gp path] (1.320,5.459)--(11.947,5.459);
\gpcolor{color=gp lt color border}
\gpsetlinetype{gp lt border}
\gpsetdashtype{gp dt solid}
\gpsetlinewidth{1.00}
\draw[gp path] (1.320,5.459)--(1.500,5.459);
\draw[gp path] (11.947,5.459)--(11.767,5.459);
\node[gp node right] at (1.136,5.459) {$0.6$};
\gpcolor{color=gp lt color axes}
\gpsetlinetype{gp lt axes}
\gpsetdashtype{gp dt axes}
\gpsetlinewidth{0.50}
\draw[gp path] (1.320,6.950)--(11.947,6.950);
\gpcolor{color=gp lt color border}
\gpsetlinetype{gp lt border}
\gpsetdashtype{gp dt solid}
\gpsetlinewidth{1.00}
\draw[gp path] (1.320,6.950)--(1.500,6.950);
\draw[gp path] (11.947,6.950)--(11.767,6.950);
\node[gp node right] at (1.136,6.950) {$0.8$};
\gpcolor{color=gp lt color axes}
\gpsetlinetype{gp lt axes}
\gpsetdashtype{gp dt axes}
\gpsetlinewidth{0.50}
\draw[gp path] (1.320,8.441)--(11.947,8.441);
\gpcolor{color=gp lt color border}
\gpsetlinetype{gp lt border}
\gpsetdashtype{gp dt solid}
\gpsetlinewidth{1.00}
\draw[gp path] (1.320,8.441)--(1.500,8.441);
\draw[gp path] (11.947,8.441)--(11.767,8.441);
\node[gp node right] at (1.136,8.441) {$1$};
\gpcolor{color=gp lt color axes}
\gpsetlinetype{gp lt axes}
\gpsetdashtype{gp dt axes}
\gpsetlinewidth{0.50}
\draw[gp path] (1.320,0.985)--(1.320,8.441);
\gpcolor{color=gp lt color border}
\gpsetlinetype{gp lt border}
\gpsetdashtype{gp dt solid}
\gpsetlinewidth{1.00}
\draw[gp path] (1.320,0.985)--(1.320,1.165);
\draw[gp path] (1.320,8.441)--(1.320,8.261);
\node[gp node center] at (1.320,0.677) {$0$};
\gpcolor{color=gp lt color axes}
\gpsetlinetype{gp lt axes}
\gpsetdashtype{gp dt axes}
\gpsetlinewidth{0.50}
\draw[gp path] (5.571,0.985)--(5.571,8.441);
\gpcolor{color=gp lt color border}
\gpsetlinetype{gp lt border}
\gpsetdashtype{gp dt solid}
\gpsetlinewidth{1.00}
\draw[gp path] (5.571,0.985)--(5.571,1.165);
\draw[gp path] (5.571,8.441)--(5.571,8.261);
\node[gp node center] at (5.571,0.677) {$14$};
\gpcolor{color=gp lt color axes}
\gpsetlinetype{gp lt axes}
\gpsetdashtype{gp dt axes}
\gpsetlinewidth{0.50}
\draw[gp path] (9.822,0.985)--(9.822,1.165);
\draw[gp path] (9.822,2.705)--(9.822,8.441);
\gpcolor{color=gp lt color border}
\gpsetlinetype{gp lt border}
\gpsetdashtype{gp dt solid}
\gpsetlinewidth{1.00}
\draw[gp path] (9.822,0.985)--(9.822,1.165);
\draw[gp path] (9.822,8.441)--(9.822,8.261);
\node[gp node center] at (9.822,0.677) {$28$};
\draw[gp path] (1.320,8.441)--(1.320,0.985)--(11.947,0.985)--(11.947,8.441)--cycle;
\node[gp node center,rotate=-270] at (0.292,4.713) {Test accuracy};
\node[gp node center] at (6.633,0.215) {Budget used};
\node[gp node right] at (10.479,2.551) {GCN-entropy-LP};
\gpcolor{rgb color={0.580,0.000,0.827}}
\draw[gp path] (10.663,2.551)--(11.579,2.551);
\draw[gp path] (1.320,1.946)--(3.445,2.776)--(5.571,3.040)--(7.696,4.122)--(9.822,4.457)%
  --(11.947,4.790);
\gpcolor{color=gp lt color border}
\node[gp node right] at (10.479,2.243) {GCN-medoids-LP};
\gpcolor{rgb color={0.000,0.620,0.451}}
\draw[gp path] (10.663,2.243)--(11.579,2.243);
\draw[gp path] (1.320,1.946)--(3.445,2.903)--(5.571,3.638)--(7.696,4.469)--(9.822,4.652)%
  --(11.947,4.977);
\gpcolor{color=gp lt color border}
\node[gp node right] at (10.479,1.935) {GCN-own-LP};
\gpcolor{rgb color={0.337,0.706,0.914}}
\draw[gp path] (10.663,1.935)--(11.579,1.935);
\draw[gp path] (1.320,1.946)--(3.445,3.228)--(5.571,3.486)--(7.696,3.922)--(9.822,4.474)%
  --(11.947,4.724);
\gpcolor{color=gp lt color border}
\node[gp node right] at (10.479,1.627) {GCN-pagerank-LP};
\gpcolor{rgb color={0.902,0.624,0.000}}
\draw[gp path] (10.663,1.627)--(11.579,1.627);
\draw[gp path] (1.320,1.946)--(3.445,2.818)--(5.571,3.324)--(7.696,3.730)--(9.822,4.002)%
  --(11.947,4.179);
\gpcolor{color=gp lt color border}
\node[gp node right] at (10.479,1.319) {GCN-random-LP};
\gpcolor{rgb color={0.941,0.894,0.259}}
\draw[gp path] (10.663,1.319)--(11.579,1.319);
\draw[gp path] (1.320,1.946)--(3.445,2.776)--(5.571,3.142)--(7.696,3.875)--(9.822,4.333)%
  --(11.947,4.364);
\gpcolor{color=gp lt color border}
\draw[gp path] (1.320,8.441)--(1.320,0.985)--(11.947,0.985)--(11.947,8.441)--cycle;
%% coordinates of the plot area
\gpdefrectangularnode{gp plot 1}{\pgfpoint{1.320cm}{0.985cm}}{\pgfpoint{11.947cm}{8.441cm}}
\end{tikzpicture}
%% gnuplot variables

%% file: plots/stage2/cora_gpn-gcn.tex
\begin{tikzpicture}[gnuplot]
%% generated with GNUPLOT 5.4p3 (Lua 5.4; terminal rev. Jun 2020, script rev. 115)
%% Sun 21 Jan 2024 04:45:11 PM CET
\path (0.000,0.000) rectangle (12.500,8.750);
\gpcolor{color=gp lt color axes}
\gpsetlinetype{gp lt axes}
\gpsetdashtype{gp dt axes}
\gpsetlinewidth{0.50}
\draw[gp path] (1.320,0.985)--(11.947,0.985);
\gpcolor{color=gp lt color border}
\gpsetlinetype{gp lt border}
\gpsetdashtype{gp dt solid}
\gpsetlinewidth{1.00}
\draw[gp path] (1.320,0.985)--(1.500,0.985);
\draw[gp path] (11.947,0.985)--(11.767,0.985);
\node[gp node right] at (1.136,0.985) {$0$};
\gpcolor{color=gp lt color axes}
\gpsetlinetype{gp lt axes}
\gpsetdashtype{gp dt axes}
\gpsetlinewidth{0.50}
\draw[gp path] (1.320,2.476)--(7.535,2.476);
\draw[gp path] (11.763,2.476)--(11.947,2.476);
\gpcolor{color=gp lt color border}
\gpsetlinetype{gp lt border}
\gpsetdashtype{gp dt solid}
\gpsetlinewidth{1.00}
\draw[gp path] (1.320,2.476)--(1.500,2.476);
\draw[gp path] (11.947,2.476)--(11.767,2.476);
\node[gp node right] at (1.136,2.476) {$0.2$};
\gpcolor{color=gp lt color axes}
\gpsetlinetype{gp lt axes}
\gpsetdashtype{gp dt axes}
\gpsetlinewidth{0.50}
\draw[gp path] (1.320,3.967)--(11.947,3.967);
\gpcolor{color=gp lt color border}
\gpsetlinetype{gp lt border}
\gpsetdashtype{gp dt solid}
\gpsetlinewidth{1.00}
\draw[gp path] (1.320,3.967)--(1.500,3.967);
\draw[gp path] (11.947,3.967)--(11.767,3.967);
\node[gp node right] at (1.136,3.967) {$0.4$};
\gpcolor{color=gp lt color axes}
\gpsetlinetype{gp lt axes}
\gpsetdashtype{gp dt axes}
\gpsetlinewidth{0.50}
\draw[gp path] (1.320,5.459)--(11.947,5.459);
\gpcolor{color=gp lt color border}
\gpsetlinetype{gp lt border}
\gpsetdashtype{gp dt solid}
\gpsetlinewidth{1.00}
\draw[gp path] (1.320,5.459)--(1.500,5.459);
\draw[gp path] (11.947,5.459)--(11.767,5.459);
\node[gp node right] at (1.136,5.459) {$0.6$};
\gpcolor{color=gp lt color axes}
\gpsetlinetype{gp lt axes}
\gpsetdashtype{gp dt axes}
\gpsetlinewidth{0.50}
\draw[gp path] (1.320,6.950)--(11.947,6.950);
\gpcolor{color=gp lt color border}
\gpsetlinetype{gp lt border}
\gpsetdashtype{gp dt solid}
\gpsetlinewidth{1.00}
\draw[gp path] (1.320,6.950)--(1.500,6.950);
\draw[gp path] (11.947,6.950)--(11.767,6.950);
\node[gp node right] at (1.136,6.950) {$0.8$};
\gpcolor{color=gp lt color axes}
\gpsetlinetype{gp lt axes}
\gpsetdashtype{gp dt axes}
\gpsetlinewidth{0.50}
\draw[gp path] (1.320,8.441)--(11.947,8.441);
\gpcolor{color=gp lt color border}
\gpsetlinetype{gp lt border}
\gpsetdashtype{gp dt solid}
\gpsetlinewidth{1.00}
\draw[gp path] (1.320,8.441)--(1.500,8.441);
\draw[gp path] (11.947,8.441)--(11.767,8.441);
\node[gp node right] at (1.136,8.441) {$1$};
\gpcolor{color=gp lt color axes}
\gpsetlinetype{gp lt axes}
\gpsetdashtype{gp dt axes}
\gpsetlinewidth{0.50}
\draw[gp path] (1.320,0.985)--(1.320,8.441);
\gpcolor{color=gp lt color border}
\gpsetlinetype{gp lt border}
\gpsetdashtype{gp dt solid}
\gpsetlinewidth{1.00}
\draw[gp path] (1.320,0.985)--(1.320,1.165);
\draw[gp path] (1.320,8.441)--(1.320,8.261);
\node[gp node center] at (1.320,0.677) {$0$};
\gpcolor{color=gp lt color axes}
\gpsetlinetype{gp lt axes}
\gpsetdashtype{gp dt axes}
\gpsetlinewidth{0.50}
\draw[gp path] (5.571,0.985)--(5.571,8.441);
\gpcolor{color=gp lt color border}
\gpsetlinetype{gp lt border}
\gpsetdashtype{gp dt solid}
\gpsetlinewidth{1.00}
\draw[gp path] (5.571,0.985)--(5.571,1.165);
\draw[gp path] (5.571,8.441)--(5.571,8.261);
\node[gp node center] at (5.571,0.677) {$14$};
\gpcolor{color=gp lt color axes}
\gpsetlinetype{gp lt axes}
\gpsetdashtype{gp dt axes}
\gpsetlinewidth{0.50}
\draw[gp path] (9.822,0.985)--(9.822,1.165);
\draw[gp path] (9.822,2.705)--(9.822,8.441);
\gpcolor{color=gp lt color border}
\gpsetlinetype{gp lt border}
\gpsetdashtype{gp dt solid}
\gpsetlinewidth{1.00}
\draw[gp path] (9.822,0.985)--(9.822,1.165);
\draw[gp path] (9.822,8.441)--(9.822,8.261);
\node[gp node center] at (9.822,0.677) {$28$};
\draw[gp path] (1.320,8.441)--(1.320,0.985)--(11.947,0.985)--(11.947,8.441)--cycle;
\node[gp node center,rotate=-270] at (0.292,4.713) {Test accuracy};
\node[gp node center] at (6.633,0.215) {Budget used};
\node[gp node right] at (10.479,2.551) {GPN-GCN-entropy};
\gpcolor{rgb color={0.580,0.000,0.827}}
\draw[gp path] (10.663,2.551)--(11.579,2.551);
\draw[gp path] (1.320,1.974)--(3.445,3.157)--(5.571,4.104)--(7.696,4.764)--(9.822,5.442)%
  --(11.947,5.833);
\gpcolor{color=gp lt color border}
\node[gp node right] at (10.479,2.243) {GPN-GCN-medoids};
\gpcolor{rgb color={0.000,0.620,0.451}}
\draw[gp path] (10.663,2.243)--(11.579,2.243);
\draw[gp path] (1.320,1.974)--(3.445,3.160)--(5.571,4.240)--(7.696,4.921)--(9.822,5.509)%
  --(11.947,5.823);
\gpcolor{color=gp lt color border}
\node[gp node right] at (10.479,1.935) {GPN-GCN-own};
\gpcolor{rgb color={0.337,0.706,0.914}}
\draw[gp path] (10.663,1.935)--(11.579,1.935);
\draw[gp path] (1.320,1.974)--(3.445,3.288)--(5.571,4.398)--(7.696,5.007)--(9.822,5.649)%
  --(11.947,5.725);
\gpcolor{color=gp lt color border}
\node[gp node right] at (10.479,1.627) {GPN-GCN-pagerank};
\gpcolor{rgb color={0.902,0.624,0.000}}
\draw[gp path] (10.663,1.627)--(11.579,1.627);
\draw[gp path] (1.320,1.974)--(3.445,3.520)--(5.571,4.204)--(7.696,4.870)--(9.822,5.455)%
  --(11.947,5.837);
\gpcolor{color=gp lt color border}
\node[gp node right] at (10.479,1.319) {GPN-GCN-random};
\gpcolor{rgb color={0.941,0.894,0.259}}
\draw[gp path] (10.663,1.319)--(11.579,1.319);
\draw[gp path] (1.320,1.974)--(3.445,3.157)--(5.571,4.175)--(7.696,4.903)--(9.822,5.556)%
  --(11.947,5.806);
\gpcolor{color=gp lt color border}
\draw[gp path] (1.320,8.441)--(1.320,0.985)--(11.947,0.985)--(11.947,8.441)--cycle;
%% coordinates of the plot area
\gpdefrectangularnode{gp plot 1}{\pgfpoint{1.320cm}{0.985cm}}{\pgfpoint{11.947cm}{8.441cm}}
\end{tikzpicture}
%% gnuplot variables

%% file: plots/stage2/cora_gpn-gcn_lp.tex
\begin{tikzpicture}[gnuplot]
%% generated with GNUPLOT 5.4p3 (Lua 5.4; terminal rev. Jun 2020, script rev. 115)
%% Sun 21 Jan 2024 04:45:32 PM CET
\path (0.000,0.000) rectangle (12.500,8.750);
\gpcolor{color=gp lt color axes}
\gpsetlinetype{gp lt axes}
\gpsetdashtype{gp dt axes}
\gpsetlinewidth{0.50}
\draw[gp path] (1.320,0.985)--(11.947,0.985);
\gpcolor{color=gp lt color border}
\gpsetlinetype{gp lt border}
\gpsetdashtype{gp dt solid}
\gpsetlinewidth{1.00}
\draw[gp path] (1.320,0.985)--(1.500,0.985);
\draw[gp path] (11.947,0.985)--(11.767,0.985);
\node[gp node right] at (1.136,0.985) {$0$};
\gpcolor{color=gp lt color axes}
\gpsetlinetype{gp lt axes}
\gpsetdashtype{gp dt axes}
\gpsetlinewidth{0.50}
\draw[gp path] (1.320,2.476)--(6.983,2.476);
\draw[gp path] (11.763,2.476)--(11.947,2.476);
\gpcolor{color=gp lt color border}
\gpsetlinetype{gp lt border}
\gpsetdashtype{gp dt solid}
\gpsetlinewidth{1.00}
\draw[gp path] (1.320,2.476)--(1.500,2.476);
\draw[gp path] (11.947,2.476)--(11.767,2.476);
\node[gp node right] at (1.136,2.476) {$0.2$};
\gpcolor{color=gp lt color axes}
\gpsetlinetype{gp lt axes}
\gpsetdashtype{gp dt axes}
\gpsetlinewidth{0.50}
\draw[gp path] (1.320,3.967)--(11.947,3.967);
\gpcolor{color=gp lt color border}
\gpsetlinetype{gp lt border}
\gpsetdashtype{gp dt solid}
\gpsetlinewidth{1.00}
\draw[gp path] (1.320,3.967)--(1.500,3.967);
\draw[gp path] (11.947,3.967)--(11.767,3.967);
\node[gp node right] at (1.136,3.967) {$0.4$};
\gpcolor{color=gp lt color axes}
\gpsetlinetype{gp lt axes}
\gpsetdashtype{gp dt axes}
\gpsetlinewidth{0.50}
\draw[gp path] (1.320,5.459)--(11.947,5.459);
\gpcolor{color=gp lt color border}
\gpsetlinetype{gp lt border}
\gpsetdashtype{gp dt solid}
\gpsetlinewidth{1.00}
\draw[gp path] (1.320,5.459)--(1.500,5.459);
\draw[gp path] (11.947,5.459)--(11.767,5.459);
\node[gp node right] at (1.136,5.459) {$0.6$};
\gpcolor{color=gp lt color axes}
\gpsetlinetype{gp lt axes}
\gpsetdashtype{gp dt axes}
\gpsetlinewidth{0.50}
\draw[gp path] (1.320,6.950)--(11.947,6.950);
\gpcolor{color=gp lt color border}
\gpsetlinetype{gp lt border}
\gpsetdashtype{gp dt solid}
\gpsetlinewidth{1.00}
\draw[gp path] (1.320,6.950)--(1.500,6.950);
\draw[gp path] (11.947,6.950)--(11.767,6.950);
\node[gp node right] at (1.136,6.950) {$0.8$};
\gpcolor{color=gp lt color axes}
\gpsetlinetype{gp lt axes}
\gpsetdashtype{gp dt axes}
\gpsetlinewidth{0.50}
\draw[gp path] (1.320,8.441)--(11.947,8.441);
\gpcolor{color=gp lt color border}
\gpsetlinetype{gp lt border}
\gpsetdashtype{gp dt solid}
\gpsetlinewidth{1.00}
\draw[gp path] (1.320,8.441)--(1.500,8.441);
\draw[gp path] (11.947,8.441)--(11.767,8.441);
\node[gp node right] at (1.136,8.441) {$1$};
\gpcolor{color=gp lt color axes}
\gpsetlinetype{gp lt axes}
\gpsetdashtype{gp dt axes}
\gpsetlinewidth{0.50}
\draw[gp path] (1.320,0.985)--(1.320,8.441);
\gpcolor{color=gp lt color border}
\gpsetlinetype{gp lt border}
\gpsetdashtype{gp dt solid}
\gpsetlinewidth{1.00}
\draw[gp path] (1.320,0.985)--(1.320,1.165);
\draw[gp path] (1.320,8.441)--(1.320,8.261);
\node[gp node center] at (1.320,0.677) {$0$};
\gpcolor{color=gp lt color axes}
\gpsetlinetype{gp lt axes}
\gpsetdashtype{gp dt axes}
\gpsetlinewidth{0.50}
\draw[gp path] (5.571,0.985)--(5.571,8.441);
\gpcolor{color=gp lt color border}
\gpsetlinetype{gp lt border}
\gpsetdashtype{gp dt solid}
\gpsetlinewidth{1.00}
\draw[gp path] (5.571,0.985)--(5.571,1.165);
\draw[gp path] (5.571,8.441)--(5.571,8.261);
\node[gp node center] at (5.571,0.677) {$14$};
\gpcolor{color=gp lt color axes}
\gpsetlinetype{gp lt axes}
\gpsetdashtype{gp dt axes}
\gpsetlinewidth{0.50}
\draw[gp path] (9.822,0.985)--(9.822,1.165);
\draw[gp path] (9.822,2.705)--(9.822,8.441);
\gpcolor{color=gp lt color border}
\gpsetlinetype{gp lt border}
\gpsetdashtype{gp dt solid}
\gpsetlinewidth{1.00}
\draw[gp path] (9.822,0.985)--(9.822,1.165);
\draw[gp path] (9.822,8.441)--(9.822,8.261);
\node[gp node center] at (9.822,0.677) {$28$};
\draw[gp path] (1.320,8.441)--(1.320,0.985)--(11.947,0.985)--(11.947,8.441)--cycle;
\node[gp node center,rotate=-270] at (0.292,4.713) {Test accuracy};
\node[gp node center] at (6.633,0.215) {Budget used};
\node[gp node right] at (10.479,2.551) {GPN-GCN-entropy-LP};
\gpcolor{rgb color={0.580,0.000,0.827}}
\draw[gp path] (10.663,2.551)--(11.579,2.551);
\draw[gp path] (1.320,1.974)--(3.445,4.085)--(5.571,5.084)--(7.696,5.246)--(9.822,5.460)%
  --(11.947,5.602);
\gpcolor{color=gp lt color border}
\node[gp node right] at (10.479,2.243) {GPN-GCN-medoids-LP};
\gpcolor{rgb color={0.000,0.620,0.451}}
\draw[gp path] (10.663,2.243)--(11.579,2.243);
\draw[gp path] (1.320,1.974)--(3.445,3.402)--(5.571,4.878)--(7.696,5.465)--(9.822,5.796)%
  --(11.947,6.002);
\gpcolor{color=gp lt color border}
\node[gp node right] at (10.479,1.935) {GPN-GCN-own-LP};
\gpcolor{rgb color={0.337,0.706,0.914}}
\draw[gp path] (10.663,1.935)--(11.579,1.935);
\draw[gp path] (1.320,1.974)--(3.445,3.711)--(5.571,4.412)--(7.696,5.102)--(9.822,5.524)%
  --(11.947,5.717);
\gpcolor{color=gp lt color border}
\node[gp node right] at (10.479,1.627) {GPN-GCN-pagerank-LP};
\gpcolor{rgb color={0.902,0.624,0.000}}
\draw[gp path] (10.663,1.627)--(11.579,1.627);
\draw[gp path] (1.320,1.974)--(3.445,3.909)--(5.571,4.844)--(7.696,5.244)--(9.822,5.426)%
  --(11.947,5.802);
\gpcolor{color=gp lt color border}
\node[gp node right] at (10.479,1.319) {GPN-GCN-random-LP};
\gpcolor{rgb color={0.941,0.894,0.259}}
\draw[gp path] (10.663,1.319)--(11.579,1.319);
\draw[gp path] (1.320,1.974)--(3.445,4.085)--(5.571,5.050)--(7.696,5.400)--(9.822,5.564)%
  --(11.947,5.739);
\gpcolor{color=gp lt color border}
\draw[gp path] (1.320,8.441)--(1.320,0.985)--(11.947,0.985)--(11.947,8.441)--cycle;
%% coordinates of the plot area
\gpdefrectangularnode{gp plot 1}{\pgfpoint{1.320cm}{0.985cm}}{\pgfpoint{11.947cm}{8.441cm}}
\end{tikzpicture}
%% gnuplot variables

%% file: plots/stage3/cora_lp.tex
\begin{tikzpicture}[gnuplot]
%% generated with GNUPLOT 5.4p2 (Lua 5.4; terminal rev. Jun 2020, script rev. 114)
%% Mo 29 Jan 2024 10:54:17 CET
\path (0.000,0.000) rectangle (12.500,8.750);
\gpcolor{color=gp lt color axes}
\gpsetlinetype{gp lt axes}
\gpsetdashtype{gp dt axes}
\gpsetlinewidth{0.50}
\draw[gp path] (1.320,0.985)--(11.947,0.985);
\gpcolor{color=gp lt color border}
\gpsetlinetype{gp lt border}
\gpsetdashtype{gp dt solid}
\gpsetlinewidth{1.00}
\draw[gp path] (1.320,0.985)--(1.500,0.985);
\draw[gp path] (11.947,0.985)--(11.767,0.985);
\node[gp node right] at (1.136,0.985) {$0$};
\gpcolor{color=gp lt color axes}
\gpsetlinetype{gp lt axes}
\gpsetdashtype{gp dt axes}
\gpsetlinewidth{0.50}
\draw[gp path] (1.320,2.476)--(8.455,2.476);
\draw[gp path] (11.763,2.476)--(11.947,2.476);
\gpcolor{color=gp lt color border}
\gpsetlinetype{gp lt border}
\gpsetdashtype{gp dt solid}
\gpsetlinewidth{1.00}
\draw[gp path] (1.320,2.476)--(1.500,2.476);
\draw[gp path] (11.947,2.476)--(11.767,2.476);
\node[gp node right] at (1.136,2.476) {$0.2$};
\gpcolor{color=gp lt color axes}
\gpsetlinetype{gp lt axes}
\gpsetdashtype{gp dt axes}
\gpsetlinewidth{0.50}
\draw[gp path] (1.320,3.967)--(11.947,3.967);
\gpcolor{color=gp lt color border}
\gpsetlinetype{gp lt border}
\gpsetdashtype{gp dt solid}
\gpsetlinewidth{1.00}
\draw[gp path] (1.320,3.967)--(1.500,3.967);
\draw[gp path] (11.947,3.967)--(11.767,3.967);
\node[gp node right] at (1.136,3.967) {$0.4$};
\gpcolor{color=gp lt color axes}
\gpsetlinetype{gp lt axes}
\gpsetdashtype{gp dt axes}
\gpsetlinewidth{0.50}
\draw[gp path] (1.320,5.459)--(11.947,5.459);
\gpcolor{color=gp lt color border}
\gpsetlinetype{gp lt border}
\gpsetdashtype{gp dt solid}
\gpsetlinewidth{1.00}
\draw[gp path] (1.320,5.459)--(1.500,5.459);
\draw[gp path] (11.947,5.459)--(11.767,5.459);
\node[gp node right] at (1.136,5.459) {$0.6$};
\gpcolor{color=gp lt color axes}
\gpsetlinetype{gp lt axes}
\gpsetdashtype{gp dt axes}
\gpsetlinewidth{0.50}
\draw[gp path] (1.320,6.950)--(11.947,6.950);
\gpcolor{color=gp lt color border}
\gpsetlinetype{gp lt border}
\gpsetdashtype{gp dt solid}
\gpsetlinewidth{1.00}
\draw[gp path] (1.320,6.950)--(1.500,6.950);
\draw[gp path] (11.947,6.950)--(11.767,6.950);
\node[gp node right] at (1.136,6.950) {$0.8$};
\gpcolor{color=gp lt color axes}
\gpsetlinetype{gp lt axes}
\gpsetdashtype{gp dt axes}
\gpsetlinewidth{0.50}
\draw[gp path] (1.320,8.441)--(11.947,8.441);
\gpcolor{color=gp lt color border}
\gpsetlinetype{gp lt border}
\gpsetdashtype{gp dt solid}
\gpsetlinewidth{1.00}
\draw[gp path] (1.320,8.441)--(1.500,8.441);
\draw[gp path] (11.947,8.441)--(11.767,8.441);
\node[gp node right] at (1.136,8.441) {$1$};
\gpcolor{color=gp lt color axes}
\gpsetlinetype{gp lt axes}
\gpsetdashtype{gp dt axes}
\gpsetlinewidth{0.50}
\draw[gp path] (1.320,0.985)--(1.320,8.441);
\gpcolor{color=gp lt color border}
\gpsetlinetype{gp lt border}
\gpsetdashtype{gp dt solid}
\gpsetlinewidth{1.00}
\draw[gp path] (1.320,0.985)--(1.320,1.165);
\draw[gp path] (1.320,8.441)--(1.320,8.261);
\node[gp node center] at (1.320,0.677) {$0$};
\gpcolor{color=gp lt color axes}
\gpsetlinetype{gp lt axes}
\gpsetdashtype{gp dt axes}
\gpsetlinewidth{0.50}
\draw[gp path] (8.000,0.985)--(8.000,8.441);
\gpcolor{color=gp lt color border}
\gpsetlinetype{gp lt border}
\gpsetdashtype{gp dt solid}
\gpsetlinewidth{1.00}
\draw[gp path] (8.000,0.985)--(8.000,1.165);
\draw[gp path] (8.000,8.441)--(8.000,8.261);
\node[gp node center] at (8.000,0.677) {$22$};
\draw[gp path] (1.320,8.441)--(1.320,0.985)--(11.947,0.985)--(11.947,8.441)--cycle;
\node[gp node center,rotate=-270] at (0.292,4.713) {Test accuracy};
\node[gp node center] at (6.633,0.215) {Budget used};
\node[gp node right] at (10.479,2.551) {LP-entropy};
\gpcolor{rgb color={0.580,0.000,0.827}}
\draw[gp path] (10.663,2.551)--(11.579,2.551);
\draw[gp path] (1.320,2.037)--(4.660,4.215)--(8.000,4.729)--(11.340,4.926)--(11.947,4.970);
\gpcolor{color=gp lt color border}
\node[gp node right] at (10.479,2.243) {LP-medoids};
\gpcolor{rgb color={0.000,0.620,0.451}}
\draw[gp path] (10.663,2.243)--(11.579,2.243);
\draw[gp path] (1.320,2.037)--(4.660,3.909)--(8.000,4.720)--(11.340,5.135)--(11.947,5.162);
\gpcolor{color=gp lt color border}
\node[gp node right] at (10.479,1.935) {LP-own};
\gpcolor{rgb color={0.337,0.706,0.914}}
\draw[gp path] (10.663,1.935)--(11.579,1.935);
\draw[gp path] (1.320,2.037)--(4.660,4.150)--(8.000,4.700)--(11.340,4.964)--(11.947,4.988);
\gpcolor{color=gp lt color border}
\node[gp node right] at (10.479,1.627) {LP-pagerank};
\gpcolor{rgb color={0.902,0.624,0.000}}
\draw[gp path] (10.663,1.627)--(11.579,1.627);
\draw[gp path] (1.320,2.037)--(4.660,4.342)--(8.000,4.838)--(11.340,4.913)--(11.947,4.950);
\gpcolor{color=gp lt color border}
\node[gp node right] at (10.479,1.319) {LP-random};
\gpcolor{rgb color={0.941,0.894,0.259}}
\draw[gp path] (10.663,1.319)--(11.579,1.319);
\draw[gp path] (1.320,2.037)--(4.660,4.215)--(8.000,4.729)--(11.340,4.926)--(11.947,4.970);
\gpcolor{color=gp lt color border}
\draw[gp path] (1.320,8.441)--(1.320,0.985)--(11.947,0.985)--(11.947,8.441)--cycle;
%% coordinates of the plot area
\gpdefrectangularnode{gp plot 1}{\pgfpoint{1.320cm}{0.985cm}}{\pgfpoint{11.947cm}{8.441cm}}
\end{tikzpicture}
%% gnuplot variables

%% file: plots/stage3/cora_lp_lp.tex
\begin{tikzpicture}[gnuplot]
%% generated with GNUPLOT 5.4p2 (Lua 5.4; terminal rev. Jun 2020, script rev. 114)
%% Mo 29 Jan 2024 10:54:41 CET
\path (0.000,0.000) rectangle (12.500,8.750);
\gpcolor{color=gp lt color axes}
\gpsetlinetype{gp lt axes}
\gpsetdashtype{gp dt axes}
\gpsetlinewidth{0.50}
\draw[gp path] (1.320,0.985)--(11.947,0.985);
\gpcolor{color=gp lt color border}
\gpsetlinetype{gp lt border}
\gpsetdashtype{gp dt solid}
\gpsetlinewidth{1.00}
\draw[gp path] (1.320,0.985)--(1.500,0.985);
\draw[gp path] (11.947,0.985)--(11.767,0.985);
\node[gp node right] at (1.136,0.985) {$0$};
\gpcolor{color=gp lt color axes}
\gpsetlinetype{gp lt axes}
\gpsetdashtype{gp dt axes}
\gpsetlinewidth{0.50}
\draw[gp path] (1.320,2.476)--(7.903,2.476);
\draw[gp path] (11.763,2.476)--(11.947,2.476);
\gpcolor{color=gp lt color border}
\gpsetlinetype{gp lt border}
\gpsetdashtype{gp dt solid}
\gpsetlinewidth{1.00}
\draw[gp path] (1.320,2.476)--(1.500,2.476);
\draw[gp path] (11.947,2.476)--(11.767,2.476);
\node[gp node right] at (1.136,2.476) {$0.2$};
\gpcolor{color=gp lt color axes}
\gpsetlinetype{gp lt axes}
\gpsetdashtype{gp dt axes}
\gpsetlinewidth{0.50}
\draw[gp path] (1.320,3.967)--(11.947,3.967);
\gpcolor{color=gp lt color border}
\gpsetlinetype{gp lt border}
\gpsetdashtype{gp dt solid}
\gpsetlinewidth{1.00}
\draw[gp path] (1.320,3.967)--(1.500,3.967);
\draw[gp path] (11.947,3.967)--(11.767,3.967);
\node[gp node right] at (1.136,3.967) {$0.4$};
\gpcolor{color=gp lt color axes}
\gpsetlinetype{gp lt axes}
\gpsetdashtype{gp dt axes}
\gpsetlinewidth{0.50}
\draw[gp path] (1.320,5.459)--(11.947,5.459);
\gpcolor{color=gp lt color border}
\gpsetlinetype{gp lt border}
\gpsetdashtype{gp dt solid}
\gpsetlinewidth{1.00}
\draw[gp path] (1.320,5.459)--(1.500,5.459);
\draw[gp path] (11.947,5.459)--(11.767,5.459);
\node[gp node right] at (1.136,5.459) {$0.6$};
\gpcolor{color=gp lt color axes}
\gpsetlinetype{gp lt axes}
\gpsetdashtype{gp dt axes}
\gpsetlinewidth{0.50}
\draw[gp path] (1.320,6.950)--(11.947,6.950);
\gpcolor{color=gp lt color border}
\gpsetlinetype{gp lt border}
\gpsetdashtype{gp dt solid}
\gpsetlinewidth{1.00}
\draw[gp path] (1.320,6.950)--(1.500,6.950);
\draw[gp path] (11.947,6.950)--(11.767,6.950);
\node[gp node right] at (1.136,6.950) {$0.8$};
\gpcolor{color=gp lt color axes}
\gpsetlinetype{gp lt axes}
\gpsetdashtype{gp dt axes}
\gpsetlinewidth{0.50}
\draw[gp path] (1.320,8.441)--(11.947,8.441);
\gpcolor{color=gp lt color border}
\gpsetlinetype{gp lt border}
\gpsetdashtype{gp dt solid}
\gpsetlinewidth{1.00}
\draw[gp path] (1.320,8.441)--(1.500,8.441);
\draw[gp path] (11.947,8.441)--(11.767,8.441);
\node[gp node right] at (1.136,8.441) {$1$};
\gpcolor{color=gp lt color axes}
\gpsetlinetype{gp lt axes}
\gpsetdashtype{gp dt axes}
\gpsetlinewidth{0.50}
\draw[gp path] (1.320,0.985)--(1.320,8.441);
\gpcolor{color=gp lt color border}
\gpsetlinetype{gp lt border}
\gpsetdashtype{gp dt solid}
\gpsetlinewidth{1.00}
\draw[gp path] (1.320,0.985)--(1.320,1.165);
\draw[gp path] (1.320,8.441)--(1.320,8.261);
\node[gp node center] at (1.320,0.677) {$0$};
\gpcolor{color=gp lt color axes}
\gpsetlinetype{gp lt axes}
\gpsetdashtype{gp dt axes}
\gpsetlinewidth{0.50}
\draw[gp path] (8.000,0.985)--(8.000,1.165);
\draw[gp path] (8.000,2.705)--(8.000,8.441);
\gpcolor{color=gp lt color border}
\gpsetlinetype{gp lt border}
\gpsetdashtype{gp dt solid}
\gpsetlinewidth{1.00}
\draw[gp path] (8.000,0.985)--(8.000,1.165);
\draw[gp path] (8.000,8.441)--(8.000,8.261);
\node[gp node center] at (8.000,0.677) {$22$};
\draw[gp path] (1.320,8.441)--(1.320,0.985)--(11.947,0.985)--(11.947,8.441)--cycle;
\node[gp node center,rotate=-270] at (0.292,4.713) {Test accuracy};
\node[gp node center] at (6.633,0.215) {Budget used};
\node[gp node right] at (10.479,2.551) {LP-entropy-LP};
\gpcolor{rgb color={0.580,0.000,0.827}}
\draw[gp path] (10.663,2.551)--(11.579,2.551);
\draw[gp path] (1.320,2.037)--(4.660,4.215)--(8.000,4.729)--(11.340,4.926)--(11.947,4.970);
\gpcolor{color=gp lt color border}
\node[gp node right] at (10.479,2.243) {LP-medoids-LP};
\gpcolor{rgb color={0.000,0.620,0.451}}
\draw[gp path] (10.663,2.243)--(11.579,2.243);
\draw[gp path] (1.320,2.037)--(4.660,3.909)--(8.000,4.720)--(11.340,5.135)--(11.947,5.162);
\gpcolor{color=gp lt color border}
\node[gp node right] at (10.479,1.935) {LP-own-LP};
\gpcolor{rgb color={0.337,0.706,0.914}}
\draw[gp path] (10.663,1.935)--(11.579,1.935);
\draw[gp path] (1.320,2.037)--(4.660,4.150)--(8.000,4.700)--(11.340,4.964)--(11.947,4.988);
\gpcolor{color=gp lt color border}
\node[gp node right] at (10.479,1.627) {LP-pagerank-LP};
\gpcolor{rgb color={0.902,0.624,0.000}}
\draw[gp path] (10.663,1.627)--(11.579,1.627);
\draw[gp path] (1.320,2.037)--(4.660,4.342)--(8.000,4.838)--(11.340,4.913)--(11.947,4.950);
\gpcolor{color=gp lt color border}
\node[gp node right] at (10.479,1.319) {LP-random-LP};
\gpcolor{rgb color={0.941,0.894,0.259}}
\draw[gp path] (10.663,1.319)--(11.579,1.319);
\draw[gp path] (1.320,2.037)--(4.660,4.215)--(8.000,4.729)--(11.340,4.926)--(11.947,4.970);
\gpcolor{color=gp lt color border}
\draw[gp path] (1.320,8.441)--(1.320,0.985)--(11.947,0.985)--(11.947,8.441)--cycle;
%% coordinates of the plot area
\gpdefrectangularnode{gp plot 1}{\pgfpoint{1.320cm}{0.985cm}}{\pgfpoint{11.947cm}{8.441cm}}
\end{tikzpicture}
%% gnuplot variables

%% file: plots/stage3/cora_gcn.tex
\begin{tikzpicture}[gnuplot]
%% generated with GNUPLOT 5.4p2 (Lua 5.4; terminal rev. Jun 2020, script rev. 114)
%% Mo 29 Jan 2024 10:55:58 CET
\path (0.000,0.000) rectangle (12.500,8.750);
\gpcolor{color=gp lt color axes}
\gpsetlinetype{gp lt axes}
\gpsetdashtype{gp dt axes}
\gpsetlinewidth{0.50}
\draw[gp path] (1.320,0.985)--(11.947,0.985);
\gpcolor{color=gp lt color border}
\gpsetlinetype{gp lt border}
\gpsetdashtype{gp dt solid}
\gpsetlinewidth{1.00}
\draw[gp path] (1.320,0.985)--(1.500,0.985);
\draw[gp path] (11.947,0.985)--(11.767,0.985);
\node[gp node right] at (1.136,0.985) {$0$};
\gpcolor{color=gp lt color axes}
\gpsetlinetype{gp lt axes}
\gpsetdashtype{gp dt axes}
\gpsetlinewidth{0.50}
\draw[gp path] (1.320,2.476)--(8.271,2.476);
\draw[gp path] (11.763,2.476)--(11.947,2.476);
\gpcolor{color=gp lt color border}
\gpsetlinetype{gp lt border}
\gpsetdashtype{gp dt solid}
\gpsetlinewidth{1.00}
\draw[gp path] (1.320,2.476)--(1.500,2.476);
\draw[gp path] (11.947,2.476)--(11.767,2.476);
\node[gp node right] at (1.136,2.476) {$0.2$};
\gpcolor{color=gp lt color axes}
\gpsetlinetype{gp lt axes}
\gpsetdashtype{gp dt axes}
\gpsetlinewidth{0.50}
\draw[gp path] (1.320,3.967)--(11.947,3.967);
\gpcolor{color=gp lt color border}
\gpsetlinetype{gp lt border}
\gpsetdashtype{gp dt solid}
\gpsetlinewidth{1.00}
\draw[gp path] (1.320,3.967)--(1.500,3.967);
\draw[gp path] (11.947,3.967)--(11.767,3.967);
\node[gp node right] at (1.136,3.967) {$0.4$};
\gpcolor{color=gp lt color axes}
\gpsetlinetype{gp lt axes}
\gpsetdashtype{gp dt axes}
\gpsetlinewidth{0.50}
\draw[gp path] (1.320,5.459)--(11.947,5.459);
\gpcolor{color=gp lt color border}
\gpsetlinetype{gp lt border}
\gpsetdashtype{gp dt solid}
\gpsetlinewidth{1.00}
\draw[gp path] (1.320,5.459)--(1.500,5.459);
\draw[gp path] (11.947,5.459)--(11.767,5.459);
\node[gp node right] at (1.136,5.459) {$0.6$};
\gpcolor{color=gp lt color axes}
\gpsetlinetype{gp lt axes}
\gpsetdashtype{gp dt axes}
\gpsetlinewidth{0.50}
\draw[gp path] (1.320,6.950)--(11.947,6.950);
\gpcolor{color=gp lt color border}
\gpsetlinetype{gp lt border}
\gpsetdashtype{gp dt solid}
\gpsetlinewidth{1.00}
\draw[gp path] (1.320,6.950)--(1.500,6.950);
\draw[gp path] (11.947,6.950)--(11.767,6.950);
\node[gp node right] at (1.136,6.950) {$0.8$};
\gpcolor{color=gp lt color axes}
\gpsetlinetype{gp lt axes}
\gpsetdashtype{gp dt axes}
\gpsetlinewidth{0.50}
\draw[gp path] (1.320,8.441)--(11.947,8.441);
\gpcolor{color=gp lt color border}
\gpsetlinetype{gp lt border}
\gpsetdashtype{gp dt solid}
\gpsetlinewidth{1.00}
\draw[gp path] (1.320,8.441)--(1.500,8.441);
\draw[gp path] (11.947,8.441)--(11.767,8.441);
\node[gp node right] at (1.136,8.441) {$1$};
\gpcolor{color=gp lt color axes}
\gpsetlinetype{gp lt axes}
\gpsetdashtype{gp dt axes}
\gpsetlinewidth{0.50}
\draw[gp path] (1.320,0.985)--(1.320,8.441);
\gpcolor{color=gp lt color border}
\gpsetlinetype{gp lt border}
\gpsetdashtype{gp dt solid}
\gpsetlinewidth{1.00}
\draw[gp path] (1.320,0.985)--(1.320,1.165);
\draw[gp path] (1.320,8.441)--(1.320,8.261);
\node[gp node center] at (1.320,0.677) {$0$};
\gpcolor{color=gp lt color axes}
\gpsetlinetype{gp lt axes}
\gpsetdashtype{gp dt axes}
\gpsetlinewidth{0.50}
\draw[gp path] (8.000,0.985)--(8.000,8.441);
\gpcolor{color=gp lt color border}
\gpsetlinetype{gp lt border}
\gpsetdashtype{gp dt solid}
\gpsetlinewidth{1.00}
\draw[gp path] (8.000,0.985)--(8.000,1.165);
\draw[gp path] (8.000,8.441)--(8.000,8.261);
\node[gp node center] at (8.000,0.677) {$22$};
\draw[gp path] (1.320,8.441)--(1.320,0.985)--(11.947,0.985)--(11.947,8.441)--cycle;
\node[gp node center,rotate=-270] at (0.292,4.713) {Test accuracy};
\node[gp node center] at (6.633,0.215) {Budget used};
\node[gp node right] at (10.479,2.551) {GCN-entropy};
\gpcolor{rgb color={0.580,0.000,0.827}}
\draw[gp path] (10.663,2.551)--(11.579,2.551);
\draw[gp path] (1.320,1.565)--(4.660,3.125)--(8.000,3.575)--(11.340,4.284)--(11.947,4.727);
\gpcolor{color=gp lt color border}
\node[gp node right] at (10.479,2.243) {GCN-medoids};
\gpcolor{rgb color={0.000,0.620,0.451}}
\draw[gp path] (10.663,2.243)--(11.579,2.243);
\draw[gp path] (1.320,1.565)--(4.660,3.453)--(8.000,3.856)--(11.340,4.591)--(11.947,5.199);
\gpcolor{color=gp lt color border}
\node[gp node right] at (10.479,1.935) {GCN-own};
\gpcolor{rgb color={0.337,0.706,0.914}}
\draw[gp path] (10.663,1.935)--(11.579,1.935);
\draw[gp path] (1.320,1.565)--(4.660,3.039)--(8.000,4.228)--(11.340,4.749)--(11.947,5.028);
\gpcolor{color=gp lt color border}
\node[gp node right] at (10.479,1.627) {GCN-pagerank};
\gpcolor{rgb color={0.902,0.624,0.000}}
\draw[gp path] (10.663,1.627)--(11.579,1.627);
\draw[gp path] (1.320,1.565)--(4.660,2.978)--(8.000,4.053)--(11.340,4.666)--(11.947,5.051);
\gpcolor{color=gp lt color border}
\node[gp node right] at (10.479,1.319) {GCN-random};
\gpcolor{rgb color={0.941,0.894,0.259}}
\draw[gp path] (10.663,1.319)--(11.579,1.319);
\draw[gp path] (1.320,1.565)--(4.660,3.125)--(8.000,3.673)--(11.340,4.282)--(11.947,4.694);
\gpcolor{color=gp lt color border}
\draw[gp path] (1.320,8.441)--(1.320,0.985)--(11.947,0.985)--(11.947,8.441)--cycle;
%% coordinates of the plot area
\gpdefrectangularnode{gp plot 1}{\pgfpoint{1.320cm}{0.985cm}}{\pgfpoint{11.947cm}{8.441cm}}
\end{tikzpicture}
%% gnuplot variables

%% file: plots/stage3/cora_gcn_lp.tex
\begin{tikzpicture}[gnuplot]
%% generated with GNUPLOT 5.4p2 (Lua 5.4; terminal rev. Jun 2020, script rev. 114)
%% Mo 29 Jan 2024 10:56:17 CET
\path (0.000,0.000) rectangle (12.500,8.750);
\gpcolor{color=gp lt color axes}
\gpsetlinetype{gp lt axes}
\gpsetdashtype{gp dt axes}
\gpsetlinewidth{0.50}
\draw[gp path] (1.320,0.985)--(11.947,0.985);
\gpcolor{color=gp lt color border}
\gpsetlinetype{gp lt border}
\gpsetdashtype{gp dt solid}
\gpsetlinewidth{1.00}
\draw[gp path] (1.320,0.985)--(1.500,0.985);
\draw[gp path] (11.947,0.985)--(11.767,0.985);
\node[gp node right] at (1.136,0.985) {$0$};
\gpcolor{color=gp lt color axes}
\gpsetlinetype{gp lt axes}
\gpsetdashtype{gp dt axes}
\gpsetlinewidth{0.50}
\draw[gp path] (1.320,2.476)--(7.719,2.476);
\draw[gp path] (11.763,2.476)--(11.947,2.476);
\gpcolor{color=gp lt color border}
\gpsetlinetype{gp lt border}
\gpsetdashtype{gp dt solid}
\gpsetlinewidth{1.00}
\draw[gp path] (1.320,2.476)--(1.500,2.476);
\draw[gp path] (11.947,2.476)--(11.767,2.476);
\node[gp node right] at (1.136,2.476) {$0.2$};
\gpcolor{color=gp lt color axes}
\gpsetlinetype{gp lt axes}
\gpsetdashtype{gp dt axes}
\gpsetlinewidth{0.50}
\draw[gp path] (1.320,3.967)--(11.947,3.967);
\gpcolor{color=gp lt color border}
\gpsetlinetype{gp lt border}
\gpsetdashtype{gp dt solid}
\gpsetlinewidth{1.00}
\draw[gp path] (1.320,3.967)--(1.500,3.967);
\draw[gp path] (11.947,3.967)--(11.767,3.967);
\node[gp node right] at (1.136,3.967) {$0.4$};
\gpcolor{color=gp lt color axes}
\gpsetlinetype{gp lt axes}
\gpsetdashtype{gp dt axes}
\gpsetlinewidth{0.50}
\draw[gp path] (1.320,5.459)--(11.947,5.459);
\gpcolor{color=gp lt color border}
\gpsetlinetype{gp lt border}
\gpsetdashtype{gp dt solid}
\gpsetlinewidth{1.00}
\draw[gp path] (1.320,5.459)--(1.500,5.459);
\draw[gp path] (11.947,5.459)--(11.767,5.459);
\node[gp node right] at (1.136,5.459) {$0.6$};
\gpcolor{color=gp lt color axes}
\gpsetlinetype{gp lt axes}
\gpsetdashtype{gp dt axes}
\gpsetlinewidth{0.50}
\draw[gp path] (1.320,6.950)--(11.947,6.950);
\gpcolor{color=gp lt color border}
\gpsetlinetype{gp lt border}
\gpsetdashtype{gp dt solid}
\gpsetlinewidth{1.00}
\draw[gp path] (1.320,6.950)--(1.500,6.950);
\draw[gp path] (11.947,6.950)--(11.767,6.950);
\node[gp node right] at (1.136,6.950) {$0.8$};
\gpcolor{color=gp lt color axes}
\gpsetlinetype{gp lt axes}
\gpsetdashtype{gp dt axes}
\gpsetlinewidth{0.50}
\draw[gp path] (1.320,8.441)--(11.947,8.441);
\gpcolor{color=gp lt color border}
\gpsetlinetype{gp lt border}
\gpsetdashtype{gp dt solid}
\gpsetlinewidth{1.00}
\draw[gp path] (1.320,8.441)--(1.500,8.441);
\draw[gp path] (11.947,8.441)--(11.767,8.441);
\node[gp node right] at (1.136,8.441) {$1$};
\gpcolor{color=gp lt color axes}
\gpsetlinetype{gp lt axes}
\gpsetdashtype{gp dt axes}
\gpsetlinewidth{0.50}
\draw[gp path] (1.320,0.985)--(1.320,8.441);
\gpcolor{color=gp lt color border}
\gpsetlinetype{gp lt border}
\gpsetdashtype{gp dt solid}
\gpsetlinewidth{1.00}
\draw[gp path] (1.320,0.985)--(1.320,1.165);
\draw[gp path] (1.320,8.441)--(1.320,8.261);
\node[gp node center] at (1.320,0.677) {$0$};
\gpcolor{color=gp lt color axes}
\gpsetlinetype{gp lt axes}
\gpsetdashtype{gp dt axes}
\gpsetlinewidth{0.50}
\draw[gp path] (8.000,0.985)--(8.000,1.165);
\draw[gp path] (8.000,2.705)--(8.000,8.441);
\gpcolor{color=gp lt color border}
\gpsetlinetype{gp lt border}
\gpsetdashtype{gp dt solid}
\gpsetlinewidth{1.00}
\draw[gp path] (8.000,0.985)--(8.000,1.165);
\draw[gp path] (8.000,8.441)--(8.000,8.261);
\node[gp node center] at (8.000,0.677) {$22$};
\draw[gp path] (1.320,8.441)--(1.320,0.985)--(11.947,0.985)--(11.947,8.441)--cycle;
\node[gp node center,rotate=-270] at (0.292,4.713) {Test accuracy};
\node[gp node center] at (6.633,0.215) {Budget used};
\node[gp node right] at (10.479,2.551) {GCN-entropy-LP};
\gpcolor{rgb color={0.580,0.000,0.827}}
\draw[gp path] (10.663,2.551)--(11.579,2.551);
\draw[gp path] (1.320,1.565)--(4.660,3.056)--(8.000,3.868)--(11.340,4.605)--(11.947,4.669);
\gpcolor{color=gp lt color border}
\node[gp node right] at (10.479,2.243) {GCN-medoids-LP};
\gpcolor{rgb color={0.000,0.620,0.451}}
\draw[gp path] (10.663,2.243)--(11.579,2.243);
\draw[gp path] (1.320,1.565)--(4.660,3.814)--(8.000,4.355)--(11.340,4.890)--(11.947,4.954);
\gpcolor{color=gp lt color border}
\node[gp node right] at (10.479,1.935) {GCN-own-LP};
\gpcolor{rgb color={0.337,0.706,0.914}}
\draw[gp path] (10.663,1.935)--(11.579,1.935);
\draw[gp path] (1.320,1.565)--(4.660,3.388)--(8.000,4.120)--(11.340,4.495)--(11.947,4.743);
\gpcolor{color=gp lt color border}
\node[gp node right] at (10.479,1.627) {GCN-pagerank-LP};
\gpcolor{rgb color={0.902,0.624,0.000}}
\draw[gp path] (10.663,1.627)--(11.579,1.627);
\draw[gp path] (1.320,1.565)--(4.660,3.551)--(8.000,4.366)--(11.340,4.675)--(11.947,4.648);
\gpcolor{color=gp lt color border}
\node[gp node right] at (10.479,1.319) {GCN-random-LP};
\gpcolor{rgb color={0.941,0.894,0.259}}
\draw[gp path] (10.663,1.319)--(11.579,1.319);
\draw[gp path] (1.320,1.565)--(4.660,2.867)--(8.000,3.824)--(11.340,4.488)--(11.947,4.739);
\gpcolor{color=gp lt color border}
\draw[gp path] (1.320,8.441)--(1.320,0.985)--(11.947,0.985)--(11.947,8.441)--cycle;
%% coordinates of the plot area
\gpdefrectangularnode{gp plot 1}{\pgfpoint{1.320cm}{0.985cm}}{\pgfpoint{11.947cm}{8.441cm}}
\end{tikzpicture}
%% gnuplot variables

%% file: plots/stage3/cora_gpn.tex
\begin{tikzpicture}[gnuplot]
%% generated with GNUPLOT 5.4p2 (Lua 5.4; terminal rev. Jun 2020, script rev. 114)
%% Mo 29 Jan 2024 10:55:11 CET
\path (0.000,0.000) rectangle (12.500,8.750);
\gpcolor{color=gp lt color axes}
\gpsetlinetype{gp lt axes}
\gpsetdashtype{gp dt axes}
\gpsetlinewidth{0.50}
\draw[gp path] (1.320,0.985)--(11.947,0.985);
\gpcolor{color=gp lt color border}
\gpsetlinetype{gp lt border}
\gpsetdashtype{gp dt solid}
\gpsetlinewidth{1.00}
\draw[gp path] (1.320,0.985)--(1.500,0.985);
\draw[gp path] (11.947,0.985)--(11.767,0.985);
\node[gp node right] at (1.136,0.985) {$0$};
\gpcolor{color=gp lt color axes}
\gpsetlinetype{gp lt axes}
\gpsetdashtype{gp dt axes}
\gpsetlinewidth{0.50}
\draw[gp path] (1.320,2.476)--(7.535,2.476);
\draw[gp path] (11.763,2.476)--(11.947,2.476);
\gpcolor{color=gp lt color border}
\gpsetlinetype{gp lt border}
\gpsetdashtype{gp dt solid}
\gpsetlinewidth{1.00}
\draw[gp path] (1.320,2.476)--(1.500,2.476);
\draw[gp path] (11.947,2.476)--(11.767,2.476);
\node[gp node right] at (1.136,2.476) {$0.2$};
\gpcolor{color=gp lt color axes}
\gpsetlinetype{gp lt axes}
\gpsetdashtype{gp dt axes}
\gpsetlinewidth{0.50}
\draw[gp path] (1.320,3.967)--(11.947,3.967);
\gpcolor{color=gp lt color border}
\gpsetlinetype{gp lt border}
\gpsetdashtype{gp dt solid}
\gpsetlinewidth{1.00}
\draw[gp path] (1.320,3.967)--(1.500,3.967);
\draw[gp path] (11.947,3.967)--(11.767,3.967);
\node[gp node right] at (1.136,3.967) {$0.4$};
\gpcolor{color=gp lt color axes}
\gpsetlinetype{gp lt axes}
\gpsetdashtype{gp dt axes}
\gpsetlinewidth{0.50}
\draw[gp path] (1.320,5.459)--(11.947,5.459);
\gpcolor{color=gp lt color border}
\gpsetlinetype{gp lt border}
\gpsetdashtype{gp dt solid}
\gpsetlinewidth{1.00}
\draw[gp path] (1.320,5.459)--(1.500,5.459);
\draw[gp path] (11.947,5.459)--(11.767,5.459);
\node[gp node right] at (1.136,5.459) {$0.6$};
\gpcolor{color=gp lt color axes}
\gpsetlinetype{gp lt axes}
\gpsetdashtype{gp dt axes}
\gpsetlinewidth{0.50}
\draw[gp path] (1.320,6.950)--(11.947,6.950);
\gpcolor{color=gp lt color border}
\gpsetlinetype{gp lt border}
\gpsetdashtype{gp dt solid}
\gpsetlinewidth{1.00}
\draw[gp path] (1.320,6.950)--(1.500,6.950);
\draw[gp path] (11.947,6.950)--(11.767,6.950);
\node[gp node right] at (1.136,6.950) {$0.8$};
\gpcolor{color=gp lt color axes}
\gpsetlinetype{gp lt axes}
\gpsetdashtype{gp dt axes}
\gpsetlinewidth{0.50}
\draw[gp path] (1.320,8.441)--(11.947,8.441);
\gpcolor{color=gp lt color border}
\gpsetlinetype{gp lt border}
\gpsetdashtype{gp dt solid}
\gpsetlinewidth{1.00}
\draw[gp path] (1.320,8.441)--(1.500,8.441);
\draw[gp path] (11.947,8.441)--(11.767,8.441);
\node[gp node right] at (1.136,8.441) {$1$};
\gpcolor{color=gp lt color axes}
\gpsetlinetype{gp lt axes}
\gpsetdashtype{gp dt axes}
\gpsetlinewidth{0.50}
\draw[gp path] (1.320,0.985)--(1.320,8.441);
\gpcolor{color=gp lt color border}
\gpsetlinetype{gp lt border}
\gpsetdashtype{gp dt solid}
\gpsetlinewidth{1.00}
\draw[gp path] (1.320,0.985)--(1.320,1.165);
\draw[gp path] (1.320,8.441)--(1.320,8.261);
\node[gp node center] at (1.320,0.677) {$0$};
\gpcolor{color=gp lt color axes}
\gpsetlinetype{gp lt axes}
\gpsetdashtype{gp dt axes}
\gpsetlinewidth{0.50}
\draw[gp path] (8.000,0.985)--(8.000,1.165);
\draw[gp path] (8.000,2.705)--(8.000,8.441);
\gpcolor{color=gp lt color border}
\gpsetlinetype{gp lt border}
\gpsetdashtype{gp dt solid}
\gpsetlinewidth{1.00}
\draw[gp path] (8.000,0.985)--(8.000,1.165);
\draw[gp path] (8.000,8.441)--(8.000,8.261);
\node[gp node center] at (8.000,0.677) {$22$};
\draw[gp path] (1.320,8.441)--(1.320,0.985)--(11.947,0.985)--(11.947,8.441)--cycle;
\node[gp node center,rotate=-270] at (0.292,4.713) {Test accuracy};
\node[gp node center] at (6.633,0.215) {Budget used};
\node[gp node right] at (10.479,2.551) {GPN-GCN-entropy};
\gpcolor{rgb color={0.580,0.000,0.827}}
\draw[gp path] (10.663,2.551)--(11.579,2.551);
\draw[gp path] (1.320,1.733)--(4.660,3.799)--(8.000,4.966)--(11.340,5.568)--(11.947,5.875);
\gpcolor{color=gp lt color border}
\node[gp node right] at (10.479,2.243) {GPN-GCN-medoids};
\gpcolor{rgb color={0.000,0.620,0.451}}
\draw[gp path] (10.663,2.243)--(11.579,2.243);
\draw[gp path] (1.320,1.733)--(4.660,3.822)--(8.000,5.073)--(11.340,5.769)--(11.947,5.885);
\gpcolor{color=gp lt color border}
\node[gp node right] at (10.479,1.935) {GPN-GCN-own};
\gpcolor{rgb color={0.337,0.706,0.914}}
\draw[gp path] (10.663,1.935)--(11.579,1.935);
\draw[gp path] (1.320,1.733)--(4.660,3.701)--(8.000,4.620)--(11.340,5.645)--(11.947,5.776);
\gpcolor{color=gp lt color border}
\node[gp node right] at (10.479,1.627) {GPN-GCN-pagerank};
\gpcolor{rgb color={0.902,0.624,0.000}}
\draw[gp path] (10.663,1.627)--(11.579,1.627);
\draw[gp path] (1.320,1.733)--(4.660,3.537)--(8.000,4.852)--(11.340,5.529)--(11.947,5.836);
\gpcolor{color=gp lt color border}
\node[gp node right] at (10.479,1.319) {GPN-GCN-random};
\gpcolor{rgb color={0.941,0.894,0.259}}
\draw[gp path] (10.663,1.319)--(11.579,1.319);
\draw[gp path] (1.320,1.733)--(4.660,3.721)--(8.000,4.915)--(11.340,5.650)--(11.947,5.994);
\gpcolor{color=gp lt color border}
\draw[gp path] (1.320,8.441)--(1.320,0.985)--(11.947,0.985)--(11.947,8.441)--cycle;
%% coordinates of the plot area
\gpdefrectangularnode{gp plot 1}{\pgfpoint{1.320cm}{0.985cm}}{\pgfpoint{11.947cm}{8.441cm}}
\end{tikzpicture}
%% gnuplot variables

%% file: plots/stage3/cora_gpn_lp.tex
\begin{tikzpicture}[gnuplot]
%% generated with GNUPLOT 5.4p2 (Lua 5.4; terminal rev. Jun 2020, script rev. 114)
%% Mo 29 Jan 2024 10:55:29 CET
\path (0.000,0.000) rectangle (12.500,8.750);
\gpcolor{color=gp lt color axes}
\gpsetlinetype{gp lt axes}
\gpsetdashtype{gp dt axes}
\gpsetlinewidth{0.50}
\draw[gp path] (1.320,0.985)--(11.947,0.985);
\gpcolor{color=gp lt color border}
\gpsetlinetype{gp lt border}
\gpsetdashtype{gp dt solid}
\gpsetlinewidth{1.00}
\draw[gp path] (1.320,0.985)--(1.500,0.985);
\draw[gp path] (11.947,0.985)--(11.767,0.985);
\node[gp node right] at (1.136,0.985) {$0$};
\gpcolor{color=gp lt color axes}
\gpsetlinetype{gp lt axes}
\gpsetdashtype{gp dt axes}
\gpsetlinewidth{0.50}
\draw[gp path] (1.320,2.476)--(6.983,2.476);
\draw[gp path] (11.763,2.476)--(11.947,2.476);
\gpcolor{color=gp lt color border}
\gpsetlinetype{gp lt border}
\gpsetdashtype{gp dt solid}
\gpsetlinewidth{1.00}
\draw[gp path] (1.320,2.476)--(1.500,2.476);
\draw[gp path] (11.947,2.476)--(11.767,2.476);
\node[gp node right] at (1.136,2.476) {$0.2$};
\gpcolor{color=gp lt color axes}
\gpsetlinetype{gp lt axes}
\gpsetdashtype{gp dt axes}
\gpsetlinewidth{0.50}
\draw[gp path] (1.320,3.967)--(11.947,3.967);
\gpcolor{color=gp lt color border}
\gpsetlinetype{gp lt border}
\gpsetdashtype{gp dt solid}
\gpsetlinewidth{1.00}
\draw[gp path] (1.320,3.967)--(1.500,3.967);
\draw[gp path] (11.947,3.967)--(11.767,3.967);
\node[gp node right] at (1.136,3.967) {$0.4$};
\gpcolor{color=gp lt color axes}
\gpsetlinetype{gp lt axes}
\gpsetdashtype{gp dt axes}
\gpsetlinewidth{0.50}
\draw[gp path] (1.320,5.459)--(11.947,5.459);
\gpcolor{color=gp lt color border}
\gpsetlinetype{gp lt border}
\gpsetdashtype{gp dt solid}
\gpsetlinewidth{1.00}
\draw[gp path] (1.320,5.459)--(1.500,5.459);
\draw[gp path] (11.947,5.459)--(11.767,5.459);
\node[gp node right] at (1.136,5.459) {$0.6$};
\gpcolor{color=gp lt color axes}
\gpsetlinetype{gp lt axes}
\gpsetdashtype{gp dt axes}
\gpsetlinewidth{0.50}
\draw[gp path] (1.320,6.950)--(11.947,6.950);
\gpcolor{color=gp lt color border}
\gpsetlinetype{gp lt border}
\gpsetdashtype{gp dt solid}
\gpsetlinewidth{1.00}
\draw[gp path] (1.320,6.950)--(1.500,6.950);
\draw[gp path] (11.947,6.950)--(11.767,6.950);
\node[gp node right] at (1.136,6.950) {$0.8$};
\gpcolor{color=gp lt color axes}
\gpsetlinetype{gp lt axes}
\gpsetdashtype{gp dt axes}
\gpsetlinewidth{0.50}
\draw[gp path] (1.320,8.441)--(11.947,8.441);
\gpcolor{color=gp lt color border}
\gpsetlinetype{gp lt border}
\gpsetdashtype{gp dt solid}
\gpsetlinewidth{1.00}
\draw[gp path] (1.320,8.441)--(1.500,8.441);
\draw[gp path] (11.947,8.441)--(11.767,8.441);
\node[gp node right] at (1.136,8.441) {$1$};
\gpcolor{color=gp lt color axes}
\gpsetlinetype{gp lt axes}
\gpsetdashtype{gp dt axes}
\gpsetlinewidth{0.50}
\draw[gp path] (1.320,0.985)--(1.320,8.441);
\gpcolor{color=gp lt color border}
\gpsetlinetype{gp lt border}
\gpsetdashtype{gp dt solid}
\gpsetlinewidth{1.00}
\draw[gp path] (1.320,0.985)--(1.320,1.165);
\draw[gp path] (1.320,8.441)--(1.320,8.261);
\node[gp node center] at (1.320,0.677) {$0$};
\gpcolor{color=gp lt color axes}
\gpsetlinetype{gp lt axes}
\gpsetdashtype{gp dt axes}
\gpsetlinewidth{0.50}
\draw[gp path] (8.000,0.985)--(8.000,1.165);
\draw[gp path] (8.000,2.705)--(8.000,8.441);
\gpcolor{color=gp lt color border}
\gpsetlinetype{gp lt border}
\gpsetdashtype{gp dt solid}
\gpsetlinewidth{1.00}
\draw[gp path] (8.000,0.985)--(8.000,1.165);
\draw[gp path] (8.000,8.441)--(8.000,8.261);
\node[gp node center] at (8.000,0.677) {$22$};
\draw[gp path] (1.320,8.441)--(1.320,0.985)--(11.947,0.985)--(11.947,8.441)--cycle;
\node[gp node center,rotate=-270] at (0.292,4.713) {Test accuracy};
\node[gp node center] at (6.633,0.215) {Budget used};
\node[gp node right] at (10.479,2.551) {GPN-GCN-entropy-LP};
\gpcolor{rgb color={0.580,0.000,0.827}}
\draw[gp path] (10.663,2.551)--(11.579,2.551);
\draw[gp path] (1.320,1.733)--(4.660,4.808)--(8.000,5.550)--(11.340,5.910)--(11.947,5.882);
\gpcolor{color=gp lt color border}
\node[gp node right] at (10.479,2.243) {GPN-GCN-medoids-LP};
\gpcolor{rgb color={0.000,0.620,0.451}}
\draw[gp path] (10.663,2.243)--(11.579,2.243);
\draw[gp path] (1.320,1.733)--(4.660,4.469)--(8.000,5.439)--(11.340,5.798)--(11.947,5.842);
\gpcolor{color=gp lt color border}
\node[gp node right] at (10.479,1.935) {GPN-GCN-own-LP};
\gpcolor{rgb color={0.337,0.706,0.914}}
\draw[gp path] (10.663,1.935)--(11.579,1.935);
\draw[gp path] (1.320,1.733)--(4.660,4.662)--(8.000,5.523)--(11.340,5.923)--(11.947,5.947);
\gpcolor{color=gp lt color border}
\node[gp node right] at (10.479,1.627) {GPN-GCN-pagerank-LP};
\gpcolor{rgb color={0.902,0.624,0.000}}
\draw[gp path] (10.663,1.627)--(11.579,1.627);
\draw[gp path] (1.320,1.733)--(4.660,4.425)--(8.000,5.219)--(11.340,5.835)--(11.947,5.902);
\gpcolor{color=gp lt color border}
\node[gp node right] at (10.479,1.319) {GPN-GCN-random-LP};
\gpcolor{rgb color={0.941,0.894,0.259}}
\draw[gp path] (10.663,1.319)--(11.579,1.319);
\draw[gp path] (1.320,1.733)--(4.660,4.818)--(8.000,5.589)--(11.340,5.977)--(11.947,5.985);
\gpcolor{color=gp lt color border}
\draw[gp path] (1.320,8.441)--(1.320,0.985)--(11.947,0.985)--(11.947,8.441)--cycle;
%% coordinates of the plot area
\gpdefrectangularnode{gp plot 1}{\pgfpoint{1.320cm}{0.985cm}}{\pgfpoint{11.947cm}{8.441cm}}
\end{tikzpicture}
%% gnuplot variables